%% file: main_arxiv.tex
\documentclass{article}

\usepackage{iclr2027_conference,times}

\input{preamble}

\title{Video World Models Need Predictable Latents}

\author{Mattia Scardecchia \quad Sherry Yang \\
New York University \\
\texttt{mattia.scardecchia@nyu.edu}
}

\iclrfinalcopy

\begin{document}

\maketitle
\lhead{Preprint}

\input{sections/abstract}

\input{sections/introduction}

\input{sections/related_work}

\input{sections/background}

\input{sections/synthetic}

\input{sections/experiments}

\input{sections/discussion}

\input{sections/ai_statement}

\subsection*{Reproducibility statement}
Section~\ref{sec:background} and Appendices~\ref{app:objectives} to~\ref{app:evaluation} specify the objectives, architectures, training and evaluation of the video experiments, Appendix~\ref{app:matched-rec} and Table~\ref{tab:settings} give the settings of the two main comparisons, and Appendix~\ref{app:toy} specifies the synthetic study, including its data.
The source code for fine-tuning the autoencoders, training the world models and evaluating them is available at \url{https://github.com/mattia-scardecchia/predictable-latents}.

\subsection*{Acknowledgments}
This work was supported in part through the NYU IT High Performance Computing resources, services, and staff expertise.
We gratefully acknowledge use of the research computing resources of the Empire AI Consortium, Inc, with support from Empire State Development of the State of New York, the Simons Foundation, and the Secunda Family Foundation.

\bibliography{references}
\bibliographystyle{iclr2027_conference}

\clearpage
\appendix

\input{sections/appendix/guide}

\input{sections/appendix/synthetic}

\input{sections/appendix/objectives}

\input{sections/appendix/architectures}

\input{sections/appendix/training}

\input{sections/appendix/evaluation}

\input{sections/appendix/equal_reconstruction}

\input{sections/appendix/additional_results}

\end{document}

%% file: preamble.tex
\input{math_commands.tex}

\usepackage[utf8]{inputenc}
\usepackage[T1]{fontenc}
\usepackage{hyperref}
\usepackage{url}
\usepackage{booktabs}
\usepackage{array}
\usepackage{nicefrac}
\usepackage{microtype}
\usepackage{xcolor}
\definecolor{accent}{HTML}{D55E00}
\definecolor{pastframe}{HTML}{009E73}
\usepackage{colortbl}
\usepackage{graphicx}
\usepackage{placeins}

%% file: math_commands.tex
\usepackage{amsmath,amsfonts,bm}

\def\eqref#1{equation~\ref{#1}}

\def\1{\bm{1}}

\DeclareMathAlphabet{\mathsfit}{\encodingdefault}{\sfdefault}{m}{sl}
\SetMathAlphabet{\mathsfit}{bold}{\encodingdefault}{\sfdefault}{bx}{n}

\def\gN{{\mathcal{N}}}

\newcommand{\E}{\mathbb{E}}

\newcommand{\R}{\mathbb{R}}

%% file: sections/abstract.tex
\begin{abstract}
Video world models trained on large-scale robot data promise a safe and scalable approach for learning and evaluating policies that act in the physical world.
However, they still struggle with long-horizon rollouts and out-of-distribution generalization, and they remain computationally demanding at inference time.
Typically, video world models make predictions in the latent space of a reconstruction-trained autoencoder, which however is not directly optimized for dynamics modeling.
On the other hand, world models designed for latent planning usually optimize latents to be predictable, but they forgo accurate prediction of observations in pixel space.
In this paper, we show that it is possible to achieve the best of both worlds: a latent geometry optimized to make future prediction as simple as possible without sacrificing accurate prediction of future observations.
More precisely, we find that temporal objectives for representation learning, paired with a past-conditioned decoder architecture, lead to consistent improvements in generative video world models at the same training budget.
The gains include up to 5x cheaper inference, over 7x faster training, and 15--24\% lower pixel and perceptual errors out of distribution and over long horizons.
Our world models also agree more closely with real-world policy evaluation when using a vision-language model to judge task success from their pixel-space predictions.
Overall, our study indicates that temporal predictability is an important design criterion for the latent space of video world models, and provides concrete guidance on how to incorporate it into the training of autoencoders.
\end{abstract}

%% file: sections/introduction.tex
\section{Introduction}

Learning from experience with high-capacity function approximators has led to unprecedented capabilities in domains like games~\citep{silverMasteringChessShogi2017a}, mathematics~\citep{hubertOlympiadlevelFormalMathematical2026}, and programming~\citep{openaiCompetitiveProgrammingLarge2025}.
These successes have been enabled by two key ingredients: a cheap, safe, parallelizable way to interact with the environment, and a verifiable reward signal to drive policy improvement.
Extending these results to domains like robotics, autonomous driving, and scientific research is challenging because interaction with the environment is often costly and unsafe, and a reward function can be difficult to specify.
While these difficulties have been overcome in specific applications, like drone racing~\citep{kaufmannChampionlevelDroneRacing2023a} and robotic table tennis~\citep{durrOutplayingEliteTable2026}, this has required substantial engineering effort to build an accurate simulator for policy training and sim-to-real transfer.

Recently, video world models have emerged as a promising and scalable approach to enable learning from experience in non-verifiable, costly, and unsafe domains, including the physical world.
By learning a photorealistic action-conditioned simulator from video data, these models allow cheap, safe, and parallelizable interaction with imagined environments~\citep{hafnerTrainingAgentsScalable2025, sharmaWorldGymnastTrainingRobots2026}.
Furthermore, their pixel-space predictions can be fed to a general-purpose vision-language model (VLM) acting as a reward model, enabling policy evaluation~\citep{quevedoWorldGymWorldModel2025}, policy training via reinforcement learning (RL)~\citep{sharmaWorldGymnastTrainingRobots2026}, and planning via imagined rollouts~\citep{duVideoLanguagePlanning2023b}.
However, despite their promise, video world models still struggle with long-horizon rollouts~\citep{guoCtrlWorldControllable2025,bardhanPersistentRobotWorld2026,huangLIVELonghorizonInteractive2026} and out-of-distribution generalization~\citep{tsengScalablePolicyEvaluation2025,zhangWorldinWorldWorldModels2025,kangHowFarVideo2025,chenRoboticWorldModels2026}, and they remain computationally demanding at inference time~\citep{quevedoWorldGymWorldModel2025,geminiroboticsteamEvaluatingGeminiRobotics2025,jeonRoboWorldFastReliable2026,luDriftWorldFastWorld2026}.

\begin{figure}[t]
    \centering
    \includegraphics[width=\textwidth]{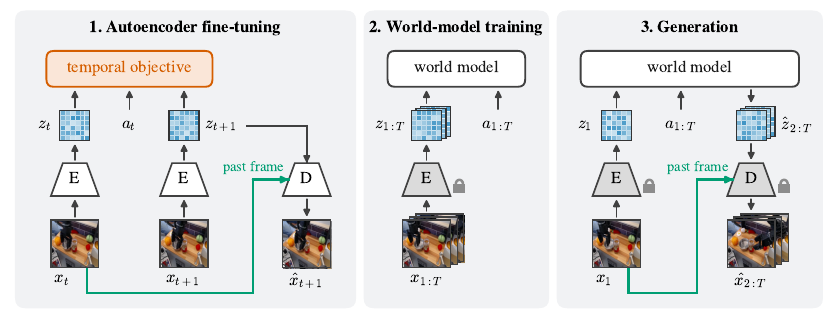}
    \caption{\textbf{Method overview.} (1) An autoencoder is fine-tuned on pairs of nearby video frames $x_t$, $x_{t+1}$, with a reconstruction loss and a temporal objective on the latents $z_t$, $z_{t+1}$ and the action $a_t$; the decoder $D$ conditions on the past frame $x_t$ to reconstruct $x_{t+1}$ from its latent. (2) An action-conditioned diffusion world model is trained on the latents from the frozen encoder $E$. (3) From a context frame $x_1$ and a sequence of actions $a_{1:T}$, the world model generates latents, which $D$ decodes conditioning on $x_1$ as the past frame. We highlight in \textcolor{accent}{orange} and \textcolor{pastframe}{green} the two components that distinguish our recipe from a standard video world model: the temporal objective, and the past-conditioned decoder.}
    \label{fig:setting}
\end{figure}

Following common practice in visual generative modeling~\citep{rombachHighResolutionImageSynthesis2022a,blattmannAlignYourLatents2023,peeblesScalableDiffusionModels2023a}, most existing video world models make predictions in the latent space of an autoencoder learned through reconstruction~\citep{bruceGenieGenerativeInteractive2024b,decartOasisUniverseTransformer2024,nvidiaCosmosWorldFoundation2025a,quevedoWorldGymWorldModel2025,huangVid2WorldCraftingVideo2026}.
However, a reconstruction objective does not necessarily produce latents whose dynamics is easy to model~\citep{yaoReconstructionVsGeneration2025,kouzelisEQVAEEquivarianceRegularized2025,skorokhodovImprovingDiffusabilityAutoencoders2025}.
A different line of work has focused on developing world models that enable efficient latent planning, by learning a compact latent space without any explicit reconstruction pressure~\citep{lecunPathAutonomousMachine2022,hansenTDMPC2ScalableRobust2024,zhouDINOWMWorld2024,sobalLearningRewardFree2025,assranVJEPA2SelfSupervised2025a,maesLeWorldModelStableEnd2026}.
This grants them freedom to discard fine-grained information that would be difficult for the world model to predict, and makes them lightweight and efficient for planning~\citep{vanasselJointEmbeddingVs2025,maesLeWorldModelStableEnd2026}.
However, that same freedom creates the risk of representation collapse, requiring dedicated and sometimes poorly understood mechanisms to mitigate it~\citep{grillBootstrapYourOwn2020,bardesVICRegVarianceInvarianceCovarianceRegularization2022,balestrieroLeJEPAProvableScalable2025,maesLeWorldModelStableEnd2026}.
Furthermore, discarding fine-grained information about observations can hinder their ability to leverage VLMs as reward models for RL, planning, and policy evaluation~\citep{pendharkarPredictiveObjectivesDiscard2026,navasardyanFailBenchHowReliable2026}.

Motivated by the complementary strengths of generative and non-generative approaches, in this work we investigate whether it is possible to get the best of both worlds: a latent space that makes dynamics modeling easy, and yet remains informative enough to support accurate reconstruction of observations.
To answer this question, we consider several families of temporal objectives for representation learning, including latent future prediction, pixel future prediction, contrastive predictive coding, and inverse dynamics~\citep{sermanetTimeContrastiveNetworks2017,oordRepresentationLearningContrastive2019a,nachumProvableRepresentationLearning2021a,hansenTDMPC2ScalableRobust2024}, that use action-labelled videos to encourage useful temporal structure in the latent space.
We carry out controlled experiments where we fine-tune a general-purpose visual autoencoder~\citep{blackforestlabsFLUX2024} with temporal and reconstruction objectives, and then train a video world model making predictions in the learned latent space.
We then measure the quality of the representations by evaluating the downstream performance of the video world model, measuring agreement of its predictions with ground truth observations and its policy evaluation ability when paired with a VLM judge.
This allows us to explore the impact of different temporal objectives, decoder architectures, and other design choices for downstream world model performance, evaluated in and out of distribution, over long prediction horizons, and varying the inference budget.
This also leads to a practical recipe that outperforms the standard practice at equal training budget in training and inference cost, agreement with ground truth and policy evaluation ability. The gains are more pronounced in challenging evaluation conditions, like out of distribution, long horizons, and sampling with few steps.

Our main contributions are the following.
\begin{itemize}
    \item \textbf{Temporal structure is useful for world models.} First, we show in a synthetic environment that learning latents that are both predictable and informative can be achieved with a combination of a reconstruction and a temporal objective, while each individually fails (Section~\ref{sec:toy}). Moving to robot manipulation videos, we find that including temporal objectives in autoencoder fine-tuning leads to improved world model performance, controlling for the reconstruction error of the autoencoder (Section~\ref{sec:matched-rec}). The most effective objectives are those that encourage predictability in latent space directly, like latent prediction and contrastive prediction, while predicting the future in pixel space barely helps. We also show that the most effective objectives induce latents with lower intrinsic dimensionality and weaker high frequency components.
    \item \textbf{Past-conditioned decoding improves the information-predictability frontier.} We find that, with temporal objectives, the training budget to reach a reconstruction target in autoencoder fine-tuning can be much larger than with reconstruction alone. To  characterize the apparent trade-off between predictability of the latents and information they retain, we vary the relative importance of the reconstruction and temporal objectives in autoencoder fine-tuning (Section~\ref{sec:tradeoff}). We then show that a simple architectural change, conditioning the decoder on a past observation, improves this trade-off. At an equal training budget, the resulting latents can carry as much information as those of reconstruction alone with the ordinary decoder, while being more predictable.
    \item \textbf{A practical recipe to improve video world models.} We combine the past-conditioned decoder with temporal objectives like latent prediction or inverse dynamics. Both objectives increase the predictability of the latents, with a 33--80\% reduction in latent MSE. We show that compared to the standard practice, at equal training budget, world models trained on the latents of our recipe obtain over 7x faster training, 5x fewer sampling steps at matched generation quality, a 15--24\% improvement in pixel and perceptual error for out-of-distribution and long-horizon extrapolation, and up to a 0.45 increase in Pearson correlation of policy evaluation between the world model and the real world (Section~\ref{sec:matched-compute}). Both the past-conditioned decoder and the objectives account for part of these gains. Finally, we verify the robustness of these results to changes in the initial autoencoder, world model size, and fine-tuning recipe (Appendix~\ref{app:ablations}).\looseness=-1
\end{itemize}

%% file: sections/related_work.tex
\section{Related work}

\paragraph{Video world models.}
Video world models generate future frames conditioned on past frames and actions, autoregressively over discrete visual tokens~\citep{bruceGenieGenerativeInteractive2024b,wuIVideoGPTInteractiveVideoGPTs2024} or, increasingly, with diffusion~\citep{alonsoDiffusionWorldModeling2024,zhuIRASimFineGrained2024,nvidiaCosmosWorldFoundation2025a,quevedoWorldGymWorldModel2025}.
Apart from a few pixel-space models~\citep{yangLearningInteractiveRealWorld2024b,alonsoDiffusionWorldModeling2024}, they predict in the latent space of an autoencoder trained to reconstruct, with pixel, perceptual and adversarial losses, and sometimes a small KL penalty~\citep{rombachHighResolutionImageSynthesis2022a,yuLanguageModelBeats2023,yangCogVideoXTextVideo2024,kongHunyuanVideoSystematicFramework2024,valevskiDiffusionModelsReal2024,nvidiaCosmosWorldFoundation2025a,teamwanWanOpenAdvanced2025,guoCtrlWorldControllable2025}.
While some recent works have adopted spatiotemporal autoencoders~\citep{yuLanguageModelBeats2023,kongHunyuanVideoSystematicFramework2024,nvidiaCosmosWorldFoundation2025a,teamwanWanOpenAdvanced2025}, time mostly serves as an additional compression axis, and sometimes enters the definition of the reconstruction objective in pixel space, like with clip-level discriminators~\citep{yuLanguageModelBeats2023,yangCogVideoXTextVideo2024,teamwanWanOpenAdvanced2025,russellGAIA2ControllableMultiView2025}, wavelet sub-bands~\citep{hacohenLTXVideoRealtime2024,liWFVAEEnhancing2024}, or an optical-flow loss between input and reconstruction~\citep{nvidiaCosmosWorldFoundation2025a}.
In contrast, we focus on temporal objectives that affect the relationship between past and future representations, acting in pixel space or in latent space directly.\looseness=-1

\paragraph{Reconstruction-free world models.}
Another line of work learns predictable latents without ever reconstructing observations: the encoder is trained jointly with a latent dynamics model to predict future latents, rewards, or values~\citep{hansenTDMPC2ScalableRobust2024,sobalLearningRewardFree2025,maesLeWorldModelStableEnd2026,okadaDreamingModelBased2020,nguyenTemporalPredictiveCoding2021,dengDreamerProReconstructionFree2021}.
This lets the latents discard content that is hard to predict or irrelevant to the task~\citep{nguyenTemporalPredictiveCoding2021,balestrieroLearningReconstructionProduces2024,vanasselJointEmbeddingVs2025}, but also makes these models susceptible to total or partial collapse, whereby latent prediction can favour slow or easily predictable features and discard task-relevant ones~\citep{sobalJointEmbeddingPredictive2022,littwinJEPAAvoidsNoisy2024,pendharkarPredictiveObjectivesDiscard2026,karmannMotionJEPAPreventingTemporal2026}.
Consequently, several of these works report that decoders trained on the latents a posteriori lose content: background texture~\citep{okadaDreamingModelBased2020,nguyenTemporalPredictiveCoding2021}, the orientation of the end effector in long predicted rollouts~\citep{maesLeWorldModelStableEnd2026}, or the foreground configuration when the background is static~\citep{karmannMotionJEPAPreventingTemporal2026}.
This limits their ability to leverage generalist VLMs as a judge for policy training, policy evaluation, and planning with the world model~\citep{tsengScalablePolicyEvaluation2025,jeonRoboWorldFastReliable2026,sharmaWorldGymnastTrainingRobots2026,navasardyanFailBenchHowReliable2026}.
In contrast, we keep a reconstruction objective, which a VLM judge needs, and add temporal objectives for predictability.

\paragraph{Latent space design for generative models.}
In image generation, autoencoders that reconstruct better do not necessarily support better generative models~\citep{yaoReconstructionVsGeneration2025,yaoScalablePretrainingVisual2025}.
Proposed remedies regularize the latent space~\citep{kouzelisEQVAEEquivarianceRegularized2025,skorokhodovImprovingDiffusabilityAutoencoders2025}, align it with the features of a pretrained semantic encoder~\citep{yaoReconstructionVsGeneration2025}, or use those features directly as the latent space~\citep{zhengDiffusionTransformersRepresentation2025}.
Video generation and video world models have also adopted the semantic remedies, with frozen image or video encoders~\citep{guoVRAERethinking2026,xieVideoRAETamingVideo2026,zhangRAENWMNavigation2026,huMultiplayerInteractiveWorld2026} or alignment losses on the video tokenizer~\citep{huGAIA1Generative2023,russellGAIA2ControllableMultiView2025,fsvideoteamFSVideoFastSpeed2026,wangRepWAMWorldAction2026}.
Closest to our setting, \citet{nilakshReconstructionSemanticsMakes2026} compare frozen, off-the-shelf reconstruction and semantic encoders as latent spaces for action-conditioned video world models on Bridge V2.
In the semantic approaches the structure of the latent space is inherited from a pretrained encoder, so the effect of its training objective is not separated from that of its data, architecture, and scale.
We instead keep the autoencoder fixed and vary only its objective. Furthermore, we study objectives that relate past and future latents rather than spatial regularization or alignment with a pretrained encoder; our own experiments fine-tuning a DINOv2 encoder indicate that the two are complementary (Appendix~\ref{app:ablations}).\looseness=-1

\paragraph{Temporal objectives on the autoencoder.}
In model-based reinforcement learning, the encoder is trained jointly with a latent dynamics model, which makes its latents predictable~\citep{hafnerLearningLatentDynamics2018,hafnerMasteringDiverseDomains2024a}.
Video world models, Dreamer~4 included~\citep{hafnerTrainingAgentsScalable2025}, instead train their autoencoder separately, usually for reconstruction alone, and then freeze it.
A few works add a temporal term to the autoencoder objective for video generation: future-frame prediction by the decoder~\citep{zhaoVideoGenerationPredictive2026}, slowness and uniformity losses for video world models in deterministic synthetic environments~\citep{xiaCloningDeterministicWorlds2025}, a motion-consistency loss on the robot-arm latents of a disentangled video autoencoder for robot video world models~\citep{luoEmbodiedVAEDisentangledVideo2026}, and a local spatiotemporal correlation loss for text-to-video generation~\citep{liuDelvingLatentSpectral2025}.
\citet{zhaoVideoGenerationPredictive2026} observe that their objective costs reconstruction and partly recover it by fine-tuning the decoder; however, to our knowledge, none of these works compares families of temporal objectives, none uses actions to shape the representations, and none removes the reconstruction cost: with a past-conditioned decoder, our temporally structured latents can reconstruct better than those of a standard autoencoder trained for reconstruction alone.
Finally, \citet{yeomWhatMakesVideo2026} consider objectives similar to ours, but their goal is finding features from which an inverse-dynamics probe recovers the executed action, in simulation, while we explore latents that can improve a video world model on real-world data.\looseness=-1

%% file: sections/background.tex
\section{Background}
\label{sec:background}

\paragraph{Latent video world models.}
We consider trajectories of frames $x_{1:T}$ and actions $a_{1:T}$. An encoder $E$ maps each frame independently to a latent $z_t = E(x_t)$, and a decoder $D$ maps it back to a frame $\hat{x}_t = D(z_t)$.
We use a reconstruction loss $\mathcal{L}_{\mathrm{rec}}(x_t, \hat{x}_t)$ that combines a pixel and a perceptual term~\citep{zhangUnreasonableEffectivenessDeep2018}, and we also employ a lightly weighted KL penalty~\citep{kingmaAutoEncodingVariationalBayes2014} in some experiments.
A world model $p_\theta(z_{t+1} \mid z_{1:t}, a_{1:t})$ is learned on the frozen latent space, and its generations are decoded by $D$.

\paragraph{Temporal objectives for representation learning.}
\label{sec:objectives}
We will consider, across our experiments, temporal objectives $\mathcal{L}_{\mathrm{temp}}$ based on transitions $(x_t, a_t, x_{t+1})$, using an auxiliary predictor which is discarded for world model training.
\begin{itemize}
    \item latent prediction: predict $z_{t+1}$ from $z_t$ and $a_t$, with a stop-gradient target~\citep{bardesRevisitingFeaturePrediction2024,hansenTDMPC2ScalableRobust2024,sobalLearningRewardFree2025};
    \item pixel prediction: predict $x_{t+1}$ from $z_t$ and $a_t$~\citep{ohActionConditionalVideoPrediction2015,finnUnsupervisedLearningPhysical2016};
    \item contrastive predictive coding: train an energy-based model of $z_{t+1}$ given $z_t$ and $a_t$ with an InfoNCE loss~\citep{oordRepresentationLearningContrastive2019a,nachumProvableRepresentationLearning2021a};
    \item inverse dynamics: predict $a_t$ from $z_t$ and $z_{t+1}$~\citep{agrawalLearningPokePoking2016,pathakCuriositydrivenExplorationSelfsupervised2017}.
\end{itemize}
We will consider a combined objective:
\begin{equation}
    \mathcal{L} = (1 - \lambda)\, \mathcal{L}_{\mathrm{rec}} + \lambda\, \mathcal{L}_{\mathrm{temp}},
    \label{eq:hybrid}
\end{equation}
where $\lambda \in [0, 1]$ sets the weight of the temporal term and $\lambda = 0$ recovers reconstruction alone.

\paragraph{Diffusion-forcing world model.}
We use diffusion forcing~\citep{chenDiffusionForcingNexttoken2024} to learn $p_\theta$: we train a causal video diffusion model with independent noise levels for each frame (Appendix~\ref{app:training}), and we do inference autoregressively with a pyramid schedule (Appendix~\ref{app:evaluation}) using the DDIM sampler~\citep{songDenoisingDiffusionImplicit2021}.
We use the space-time factorized diffusion transformer (DiT)~\citep{peeblesScalableDiffusionModels2023a} of \citet{quevedoWorldGymWorldModel2025}, which alternates bidirectional attention within a frame with causal attention across frames (Appendix~\ref{app:architectures}).\looseness=-1

%% file: sections/synthetic.tex
\section{Latents can be both predictable and informative}
\label{sec:toy}

\begin{figure}[t]
    \centering
    \includegraphics[width=\textwidth]{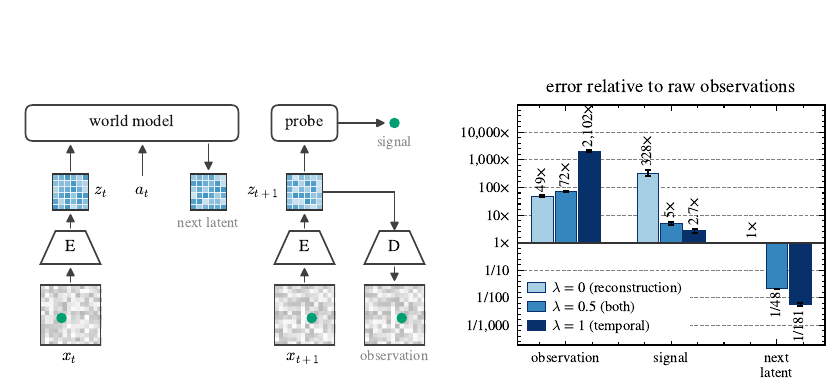}
    \caption{\textbf{Latents can be predictable and informative in a synthetic setting.} \textbf{Left}: Synthetic observations have a predictable component (signal, the dot) and an unpredictable noise component (the static). An encoder $E$ is trained with reconstruction, temporal contrastive learning or both, and three readouts are then trained on its frozen latents: a world model predicts the next latent from the current latent and action, a probe predicts the signal, and a decoder $D$ reconstructs the observation. \textbf{Right}: the error of each readout at latent size 100, divided by the error of the same readout trained on the raw observation; geometric mean and standard error over 3 seeds, lower is better. Reconstruction alone leaves the next latent as hard to predict as the raw observation and reads the signal far less accurately; contrastive learning alone loses the observation; their combination avoids both failures.}
    \label{fig:toy}
\end{figure}

We begin our investigation by considering data from a simple synthetic environment. We show that, by combining a reconstruction and a temporal objective, it is possible to learn latents that remain informative about the observations while being easy to predict. By contrast, using each objective individually either fails to be predictable, or fails to be informative (Figure \ref{fig:toy}).

\paragraph{Synthetic environment.} We consider observations $x_t = (s_t, \xi_t) \in \R^{102}$ that concatenate a two-dimensional \textit{signal} $s_t$ and a 100-dimensional vector of \textit{distractors} $\xi_t$, which are sampled from a standard normal independently at every step. The signal is sampled at time 0 from a standard normal and at each time step, given a random action $a_t \in \mathbb{R}^2$, it evolves deterministically as $s_{t+1} = s_t + a_t$.

\paragraph{Experiment.} An encoder $E$ learns to map the observation to a latent $z_t = E(x_t) \in \R^{d}$ by optimizing Equation~\ref{eq:hybrid}, with the reconstruction objective of a variational autoencoder~\citep{kingmaAutoEncodingVariationalBayes2014} and a temporal contrastive objective~\citep{oordRepresentationLearningContrastive2019a,nachumProvableRepresentationLearning2021a} as the temporal term.
On top of the frozen encoder, we train three modules: a world model that predicts the next latent from the current latent and action, a probe that reads the signal from the latent, and a decoder that reconstructs the observation. On held-out data, we measure the errors of the reconstruction, of the signal, and of the next latent. Each error is divided by the variance of its target, so that predicting the mean gives an error of 1. As a reference, the same modules are trained directly on the raw observation. In Figure~\ref{fig:toy}, we compare $\lambda = 0$ (reconstruction), $\lambda=1$ (temporal), and $\lambda=0.5$ (both) at latent size $d = 100$. Details on the data, networks, training and metrics, as well as the errors at more latent sizes and weights, can be found in Appendix~\ref{app:toy}.\looseness=-1

\paragraph{The combined objective learns predictable and informative latents.}
The reconstruction objective is known to favour the leading principal components of the observation~\citep{balestrieroLearningReconstructionProduces2024}; since every coordinate has approximately unit variance, it recovers the signal only when the latent is nearly as large as the observation, and its latent is no easier to predict than the observation itself.
The temporal contrastive objective can be minimized by capturing nothing more than the signal, since it is the only predictable part in the observation. Empirically, this is almost what happens: the latent reconstructs very poorly, but its signal and next-latent errors are two orders of magnitude below those of reconstruction.
The combined objective avoids both failures: at $d = 100$ it reconstructs nearly as well as reconstruction alone, but its next-latent error is almost 50 times lower, and it recovers the signal at every latent size.
This shows that it is possible, in this simple setting, to learn latents that are both predictable and informative. Next, we explore the use of similar temporal objectives on real-world robot manipulation videos.

%% file: sections/experiments.tex
\section{Experiments with robot manipulation video data}
\label{sec:frontier}

In the previous section we have shown, using a synthetic dataset, that it is largely unnecessary to choose between predictable and informative latents: latents can be both at the same time, making learning world models easy while still enabling reconstruction of predicted future observations. In this section, we explore whether the same is true with real-world video data, and how to leverage this to learn better video world models.

\paragraph{Common experiment setting.} Experiments in this section use the pipeline of Figure~\ref{fig:setting}. We fine-tune the pretrained autoencoder of FLUX.1~\citep{blackforestlabsFLUX2024} on Bridge V2~\citep{walkeBridgeDataV2Dataset2023}, using a combination of a temporal objective and a reconstruction objective (Equation~\ref{eq:hybrid}), and then freeze it.
Unless stated otherwise, the decoder reconstructs each frame from its own latent, $\hat{x}_t = D(z_t)$, as in standard autoencoders; we call it the ordinary decoder, and fine-tuning for reconstruction alone with the ordinary decoder, the standard practice, is our baseline.
For the reconstruction loss, we combine the mean absolute pixel error with a perceptual distance (Appendix~\ref{app:objectives}). We found using a learned discriminator~\citep{esserTamingTransformersHighResolution2021} to be unhelpful, likely because of the relatively low resolution of the data, so we do not use it. We set the KL loss weight to 0, but we show in an ablation that the key conclusions remain the same using a nonzero weight (Table~\ref{tab:kl}).
The world model is the diffusion-forcing transformer of Section~\ref{sec:background}, trained on the frozen latent space. 
We compare its action-conditioned generations with ground truth frames using pixel mean squared error (MSE), LPIPS~\citep{zhangUnreasonableEffectivenessDeep2018}, and SSIM~\citep{wangImageQualityAssessment2004}. We also compute the Fr\'echet inception and video distances, FID~\citep{heuselGANsTrainedTwo2017} and FVD~\citep{unterthinerAccurateGenerativeModels2018}, and in one experiment we assess the ability of the world model to do policy evaluation~\citep{quevedoWorldGymWorldModel2025}. We vary both the inference horizon (up to six times the training horizon) and the inference budget, and we evaluate both in distribution (with held-out Bridge V2 data) and out of distribution on videos from the AutoEval benchmark~\citep{zhouAutoEvalAutonomousEvaluation2025}, recorded on separate robot setups. In the main-text tables, we report the errors in four evaluation conditions: in distribution (ID), out of distribution (OOD), in distribution with two sampling steps instead of ten (2-step), and at the end of the longest rollouts (long-horizon). Full results with more inference budgets and inference horizons are in Appendix~\ref{app:full-tables}. Every pipeline is trained with three seeds unless stated otherwise. Appendices~\ref{app:objectives} to~\ref{app:evaluation} provide all the details.

\subsection{World models can benefit from temporal structure}
\label{sec:matched-rec}

To isolate differences due purely to the latent space geometry from differences in information retention, we first compare the inclusion of temporal objectives with pure reconstruction fine-tuning, at equal reconstruction quality of the autoencoder.

\paragraph{Fine-tuning to equal reconstruction quality.} We fine-tune the autoencoder with four objectives: reconstruction alone (baseline), and its combination with latent, pixel, or contrastive future prediction (Section~\ref{sec:objectives}), the latter with negatives drawn from other videos. We fine-tune each autoencoder until its reconstruction error reaches a common target, so that every latent supports reconstructions of the same quality and the only difference is the geometry of the latent spaces. We then train the same world model on each of them (Appendix~\ref{app:matched-rec}).

\begin{table}[t]
    \caption{\textbf{Temporal structure benefits world models at equal reconstruction quality.} With the pipeline of Figure~\ref{fig:setting}, using the ordinary decoder, we compare different objectives for fine-tuning the autoencoder until a target reconstruction error is reached, training the same world model on the learned latents. We report mean and standard error over 3 seeds for the error of world model predictions against ground truth in the four evaluation conditions of Section~\ref{sec:frontier} (pixel MSE $\times 10^3$; bold: best). Updates in thousands; Components: principal components for 95\% of the latent variance. Temporal objectives, especially latent and contrastive prediction,  improve performance across the board compared to the reconstruction baseline, but they require more autoencoder updates to reach the same reconstruction error.}
    \label{tab:matched-rec}
    \centering
    \smallskip
    \input{figures/tab_matched_reconstruction/table.tex}
\end{table}

\paragraph{Temporal structure improves the world model, but at higher cost.}
At equal reconstruction quality, all three temporal objectives improve the world model compared to the baseline (except pixel prediction in out-of-distribution pixel MSE).
Latent prediction gives the largest gains, on all but one metric of Table~\ref{tab:matched-rec}, followed by contrastive prediction, while pixel prediction stays closer to the baseline.
The temporal objectives improve performance especially in the more challenging evaluation conditions: for example, latent prediction improves MSE by 22.5\% with two sampling steps, 14.0\% out of distribution, and 9.7\% at the end of long rollouts, against 3.0\% in distribution.
FID and FVD are also slightly improved by latent and contrastive prediction (Table~\ref{tab:matched-rec-full}), and the advantage persists throughout world model training (Figure~\ref{fig:benefits-matched-rec}).
Since we are controlling for reconstruction ability, these gains are to be attributed to a more favorable latent space structure, induced by temporal objectives.
However, this comes at a cost: to reach the reconstruction target, the autoencoder needs about 35 times as many updates with latent prediction compared to reconstruction alone, and about 22 times as many with contrastive prediction. In the next section, we will address this limitation.

\subsection{Improving the information-predictability frontier}
\label{sec:tradeoff}

The results of Section~\ref{sec:matched-rec} show that temporal objectives can benefit a world model even with real-world data, but hint at a trade-off between the information a latent retains and its predictability. In this section, we characterize this trade-off and show it can be improved with a simple architectural modification to the decoder.

\paragraph{Past-conditioned decoder.} In Table~\ref{tab:matched-rec} we see that the best performing temporal objectives, which encourage the latents to be easily predictable, tend to induce a mild form of dimensional collapse: their latents need fewer principal components to explain 95\% of their variance compared to reconstruction training. This tendency competes with reconstruction, with a balance set by $\lambda$. To ease the tension, we propose to condition the decoder on a past frame.
Consecutive frames share most of their content, so the latent only has to carry what the past frame does not already provide. We implement this by adding cross-attention layers to the pretrained decoder, attending to the past frame, and initialized so that the decoder starts as the pretrained one. In world-model rollouts, the decoder is conditioned on the initial context frame. During training, several choices are possible; in the reported experiments we condition on the previous frame, $\hat{x}_{t+1} = D(z_{t+1}, x_t)$ (Figure~\ref{fig:setting}; Appendix~\ref{app:architectures}).
Similar ideas have appeared in prior work~\citep{wuPretrainingContextualizedWorld2023,micheliEfficientWorldModels2024,wuIVideoGPTInteractiveVideoGPTs2024,tianREDUCIOGenerating1K2024,fanRefTokReferenceBasedTokenization2025,fsvideoteamFSVideoFastSpeed2026,liDeepContextualVideo2021}, where conditioning serves compression, reconstruction quality, or the dynamics model's focus on what changes; here, we use it to improve the information-predictability frontier, as we will demonstrate shortly.

\begin{figure}[t]
    \centering
    \includegraphics{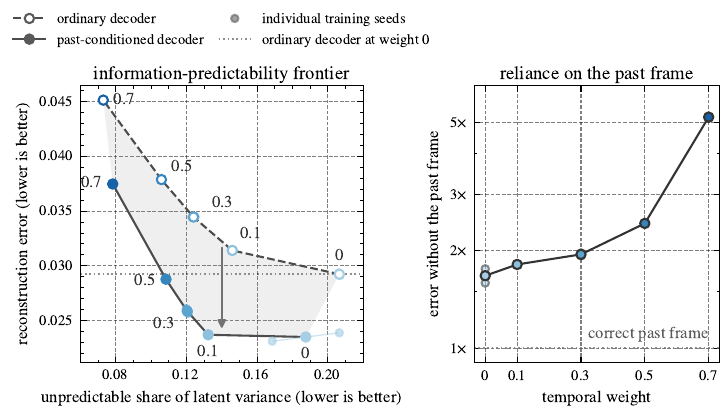}
    \caption{\textbf{The information-predictability frontier and the decoder's reliance on the past frame.} \textbf{Left}: reconstruction error against the unpredictable share of the latent variance, for autoencoders fine-tuned for 80,000 updates with latent prediction at weight $\lambda$, with the ordinary decoder or the past-conditioned decoder. Points are means over two training seeds at $\lambda = 0$ and 0.3 (faint: individual seeds) and one seed elsewhere. Raising $\lambda$ trades reconstruction for predictability, and the past-conditioned decoder improves the frontier at every weight. \textbf{Right}: reconstruction error of each past-conditioned autoencoder with its past-frame pathway disabled (Appendix~\ref{app:interventions}), divided by its error with the correct past frame (log scale); same seeds as on the left. As the pressure from the temporal objective on the latent space increases, the decoder relies more on the past frame for accurate reconstructions.}
    \label{fig:tradeoff}
\end{figure}

\paragraph{Tracing and improving the information-predictability frontier.} We fine-tune the pretrained autoencoder on pairs of consecutive frames with latent prediction, the objective with the largest gains in Section~\ref{sec:matched-rec}, at weights $\lambda \in \{0, 0.1, 0.3, 0.5, 0.7\}$, with the ordinary and with the past-conditioned decoder. Contrary to Section \ref{sec:matched-rec}, we use the same training budget for each, 80,000 autoencoder updates. We measure the information retained by the latents as the reconstruction error on held-out frame pairs. We also measure their predictability in two ways: as the fraction of the latent variance not explained by a linear action-conditioned future predictor fitted on frozen latents (Figure \ref{fig:tradeoff}), and as the error of the nonlinear prediction head used by the temporal objective at the end of training (Appendix~\ref{app:frontier-training}). Both measures agree qualitatively: raising $\lambda$ trades reconstruction for predictability. With the ordinary decoder, from $\lambda = 0$ to 0.7, the linearly unpredictable share falls from 0.21 to 0.07, while the reconstruction error rises to 54\% above that of reconstruction alone (Figure \ref{fig:tradeoff}). Introducing the past-conditioned decoder shifts the frontier, lowering the reconstruction error by 17 to 25\% at every weight: at $\lambda=0.5$, it achieves the same reconstruction error as reconstruction alone with the ordinary decoder, with about half the linearly unpredictable share of the latent variance. Thus, the past-conditioned decoder enables achieving a similar best of both worlds situation, with latents that are both predictable and informative, as in the synthetic setting of Section \ref{sec:toy}. Interestingly, as the weight of the temporal objective increases, and with it the pressure on the latents to be predictable, the decoder relies more and more on the past frame to achieve good reconstructions, as seen in the right panel of Figure~\ref{fig:tradeoff}.

\subsection{A practical recipe to improve world model training}
\label{sec:matched-compute}

In this section, we combine insights from the previous experiments and show that, at equal training budget, the past-conditioned decoder combined with a temporal objective gives better world models than reconstruction alone.

\paragraph{Fine-tuning at equal training budget.} Like in Section \ref{sec:tradeoff}, we fine-tune every autoencoder for 80,000 updates on pairs of consecutive frames, and then train the same world model on frozen latents for 100,000 updates (Table~\ref{tab:settings}). For autoencoder fine-tuning, we consider augmenting the reconstruction objective either with latent prediction, the best objective from Section~\ref{sec:matched-rec}, or with an inverse dynamics objective (Section~\ref{sec:objectives}).
Both use the past-conditioned decoder at $\lambda = 0.7$, the largest weight of Section~\ref{sec:tradeoff}.
We compare against the autoencoder fine-tuned with reconstruction alone and the past-conditioned decoder, to isolate gains from temporal structure at equal training budget, as well as reconstruction fine-tuning with the ordinary decoder (baseline), to measure the total gains from our contributions.
For reference, we also include the pretrained autoencoder without fine-tuning.
In the following, we refer to the world models trained on latents obtained with our recipe (a temporal objective and past-conditioned decoder) as the temporal world models.

\paragraph{Better generations, especially out of distribution and at long horizons.} At equal training budget, the past-conditioned decoder with either temporal objective improves pixel MSE, LPIPS, and SSIM over the baseline, both in and out of distribution, and over long horizons (Table~\ref{tab:matched-compute}, Table~\ref{tab:matched-compute-full}). 
The gains are largest in the more challenging evaluation conditions: out of distribution and at the end of long rollouts, pixel MSE and LPIPS fall by 15 to 24\%. There, fine-tuning for reconstruction alone does not lower the pixel MSE: the pretrained autoencoder without fine-tuning is slightly better than the baseline. The past-conditioned decoder alone already improves on the baseline in every condition; the temporal objectives add further gains, especially in the challenging conditions. The two temporal objectives perform similarly; inverse dynamics has a lead in distribution and on perceptual metrics out of distribution, while latent prediction does better over long horizons (every metric in Table~\ref{tab:matched-compute-full}; rollouts in Figure~\ref{fig:teaser} and Appendix~\ref{app:qualitative}). Distribution metrics (FID, FVD) also improve with both temporal objectives, although in the case of latent prediction this requires a sufficiently small weight $\lambda$ (Figure~\ref{fig:weight-dist}).

\paragraph{Faster world-model training and cheaper inference.} Latents trained with a temporal objective and a past-conditioned decoder support faster world model training. With latent prediction and inverse dynamics, the world model reaches the final validation LPIPS of the baseline 7.6 and 8.8 times sooner, against 1.8 times with the decoder alone; in SSIM, the speed-up is 15 and 13 times, and in pixel MSE 1.2 and 1.8 times (Figure~\ref{fig:benefits}).
Additionally, the temporal world models can do inference with far fewer DDIM steps without sacrificing performance. With 2 sampling steps, both generate better than the baseline does with 10 in LPIPS and SSIM, and inverse dynamics also in pixel MSE (Table~\ref{tab:matched-compute}, Figure~\ref{fig:benefits}). This is not the case with the past-conditioned decoder alone.

\paragraph{Improvements in policy evaluation ability.} Going beyond visual quality, we measure the ability of the world models to evaluate robot policies by following the protocol of WorldGym~\citep{quevedoWorldGymWorldModel2025}. Three generalist robot policies, OpenVLA~\citep{kimOpenVLAOpenSourceVisionLanguageAction2025}, Octo~\citep{ghoshOctoOpenSourceGeneralist2024} and RT-1-X~\citep{oneillOpenXEmbodimentRobotic2024}, act inside each of the five world models of Table~\ref{tab:matched-compute} on 17 Bridge tasks and a vision-language model judges task success.
We then measure the correlation between simulated and real success rates across policies and tasks (Appendix~\ref{app:worldgym}). We test two inference budgets, 10 and 2 DDIM steps, for all world models. In all cases, the temporal world models obtain better agreement with the real world than the baseline, 0.62 and 0.57 Pearson correlation against 0.51. The past-conditioned decoder alone does not improve policy evaluation: the gains come from the temporal objectives. With two steps, the correlation of every reconstruction-only world model falls to 0.15--0.21, while latent prediction keeps 0.67 and inverse dynamics 0.51: at this budget, policy evaluation remains viable only with the temporal objectives (Figure~\ref{fig:worldgym}; per-policy correlations are in Table~\ref{tab:worldgym}).

\paragraph{The gains come with more predictable latents.} The latent MSE, the error of the world model's one-step forecast of the next latent and a direct measure of how predictable each latent space is, falls by 80\% with latent prediction and by 33\% with inverse dynamics, although only the former trains it directly, against 9\% with the decoder alone (Table~\ref{tab:matched-compute}). Their latents are also lower dimensional, spatially smoother and straighter over time than those of reconstruction alone (Appendix~\ref{app:latent}).

\paragraph{Robustness to design choices.} We test the robustness of the gains due to the temporal objectives to various choices and hyperparameters in the pipeline, including training time, KL penalty, world model size, choice of encoder, and the weight of the temporal objective (Appendices~\ref{app:longer} and~\ref{app:ablations}). Each ablation compares a temporal objective with reconstruction alone, both with the past-conditioned decoder. Doubling the training time and adding a KL penalty both preserve the gains in the challenging evaluation conditions. Changing the weight of the latent prediction objective leaves most conclusions intact, but long-rollout performance only improves with a sufficiently large weight, while FID and FVD only improve if the weight is not too large. With a larger world model, the gains out of distribution and with 2 sampling steps remain, but long-rollout gains shrink. Finally, we test replacing the FLUX autoencoder with DINOv2~\citep{oquabDINOv2LearningRobust2024}, a semantic encoder with a high-dimensional latent space, and the decoder of RAE~\citep{zhengDiffusionTransformersRepresentation2025}. This leads to similar gains when the weight of the temporal objective is small enough, except with few sampling steps where the gains shrink compared to the FLUX latent space. 

\begin{table}[t]
    \caption{\textbf{Past-conditioned decoding and temporal objectives improve world models at equal training budget.} With the pipeline of Figure~\ref{fig:setting}, we compare decoders and objectives for fine-tuning the autoencoder for a fixed number of updates, training the same world model on the learned latents. The past-conditioned reconstruction row against the baseline gives the effect of the decoder; the two objective rows against that row give the effect of the temporal objective. We report mean and standard error over 3 seeds (pixel MSE $\times 10^3$; bold: best), except for policy evaluation in WorldGym (one training seed, bootstrap standard error over tasks). ID, OOD, 2-step and long-horizon as in Section~\ref{sec:frontier}; latent MSE: the error of the world model's one-step forecast of the next latent (Section~\ref{sec:matched-compute}). Our recipe improves every metric compared to the baseline; both the past-conditioned decoder and the temporal objectives account for some of the improvements.}
    \label{tab:matched-compute}
    \centering
    \smallskip
    \input{figures/tab_matched_compute/table.tex}
\end{table}

%% file: figures/tab_matched_reconstruction/table.tex
\begingroup\fontsize{8}{9.5}\selectfont\setlength{\tabcolsep}{2.2pt}
\begin{tabular}{@{}l*{10}{c}@{}}
\toprule
 & \multicolumn{2}{c}{ID $\downarrow$} & \multicolumn{2}{c}{OOD $\downarrow$} & \multicolumn{2}{c}{2-step $\downarrow$} & \multicolumn{2}{c}{Long-horizon $\downarrow$} & \multicolumn{2}{c}{Autoencoder} \\
\cmidrule(lr){2-3}\cmidrule(lr){4-5}\cmidrule(lr){6-7}\cmidrule(lr){8-9}\cmidrule(lr){10-11}
 & MSE & LPIPS & MSE & LPIPS & MSE & LPIPS & MSE & LPIPS & Updates & Components \\
\midrule
Reconstruction (baseline) & 11.65 & 0.164 & 33.87 & 0.308 & 15.95 & 0.214 & 33.26 & 0.287 & 4.4 & 388 \\
 & {\fontsize{6.5}{7.5}\selectfont$\pm$0.20} & {\fontsize{6.5}{7.5}\selectfont$\pm$0.001} & {\fontsize{6.5}{7.5}\selectfont$\pm$1.28} & {\fontsize{6.5}{7.5}\selectfont$\pm$0.008} & {\fontsize{6.5}{7.5}\selectfont$\pm$1.03} & {\fontsize{6.5}{7.5}\selectfont$\pm$0.004} & {\fontsize{6.5}{7.5}\selectfont$\pm$1.84} & {\fontsize{6.5}{7.5}\selectfont$\pm$0.008} &  &  \\[1pt]
Pixel prediction & 11.43 & 0.159 & 34.23 & 0.300 & 14.53 & 0.199 & 32.63 & 0.279 & 8.9 & 445 \\
 & {\fontsize{6.5}{7.5}\selectfont$\pm$0.22} & {\fontsize{6.5}{7.5}\selectfont$\pm$0.001} & {\fontsize{6.5}{7.5}\selectfont$\pm$0.67} & {\fontsize{6.5}{7.5}\selectfont$\pm$0.002} & {\fontsize{6.5}{7.5}\selectfont$\pm$0.75} & {\fontsize{6.5}{7.5}\selectfont$\pm$0.002} & {\fontsize{6.5}{7.5}\selectfont$\pm$0.69} & {\fontsize{6.5}{7.5}\selectfont$\pm$0.003} &  &  \\[1pt]
Contrastive prediction & {\boldmath\bfseries 11.17} & 0.160 & 31.23 & 0.303 & 12.38 & 0.185 & 31.50 & 0.277 & 96.0 & 293 \\
 & {\fontsize{6.5}{7.5}\selectfont$\pm$0.05} & {\fontsize{6.5}{7.5}\selectfont$\pm$0.001} & {\fontsize{6.5}{7.5}\selectfont$\pm$0.14} & {\fontsize{6.5}{7.5}\selectfont$\pm$0.005} & {\fontsize{6.5}{7.5}\selectfont$\pm$0.27} & {\fontsize{6.5}{7.5}\selectfont$\pm$0.001} & {\fontsize{6.5}{7.5}\selectfont$\pm$2.62} & {\fontsize{6.5}{7.5}\selectfont$\pm$0.010} &  &  \\[1pt]
Latent prediction & 11.30 & {\boldmath\bfseries 0.153} & {\boldmath\bfseries 29.14} & {\boldmath\bfseries 0.298} & {\boldmath\bfseries 12.36} & {\boldmath\bfseries 0.160} & {\boldmath\bfseries 30.04} & {\boldmath\bfseries 0.262} & 156.5 & 177 \\
 & {\fontsize{6.5}{7.5}\selectfont$\pm$0.10} & {\fontsize{6.5}{7.5}\selectfont$\pm$0.001} & {\fontsize{6.5}{7.5}\selectfont$\pm$0.55} & {\fontsize{6.5}{7.5}\selectfont$\pm$0.005} & {\fontsize{6.5}{7.5}\selectfont$\pm$0.24} & {\fontsize{6.5}{7.5}\selectfont$\pm$0.002} & {\fontsize{6.5}{7.5}\selectfont$\pm$0.43} & {\fontsize{6.5}{7.5}\selectfont$\pm$0.000} &  &  \\[1pt]
\bottomrule
\end{tabular}
\endgroup

%% file: figures/tab_matched_compute/table.tex
\begingroup\fontsize{8}{9.5}\selectfont\setlength{\tabcolsep}{2.1pt}
\begin{tabular}{@{}l*{11}{c}@{}}
\toprule
 & \multicolumn{2}{c}{ID $\downarrow$} & \multicolumn{2}{c}{OOD $\downarrow$} & \multicolumn{2}{c}{2-step $\downarrow$} & \multicolumn{2}{c}{Long-horizon $\downarrow$} & \multicolumn{2}{c}{Policy eval. $\uparrow$} & Predictability $\downarrow$ \\
\cmidrule(lr){2-3}\cmidrule(lr){4-5}\cmidrule(lr){6-7}\cmidrule(lr){8-9}\cmidrule(lr){10-11}\cmidrule(lr){12-12}
 & MSE & LPIPS & MSE & LPIPS & MSE & LPIPS & MSE & LPIPS & 10 steps & 2 steps & Latent MSE \\
\midrule
\multicolumn{12}{@{}l}{\textit{Ordinary decoder}} \\
\hspace{0.5em}Reconstruction (baseline) & 14.34 & 0.194 & 36.15 & 0.302 & 18.35 & 0.232 & 47.93 & 0.394 & 0.51 & 0.21 & 0.232 \\
 & {\fontsize{6.5}{7.5}\selectfont$\pm$0.16} & {\fontsize{6.5}{7.5}\selectfont$\pm$0.002} & {\fontsize{6.5}{7.5}\selectfont$\pm$0.98} & {\fontsize{6.5}{7.5}\selectfont$\pm$0.003} & {\fontsize{6.5}{7.5}\selectfont$\pm$0.65} & {\fontsize{6.5}{7.5}\selectfont$\pm$0.004} & {\fontsize{6.5}{7.5}\selectfont$\pm$1.51} & {\fontsize{6.5}{7.5}\selectfont$\pm$0.002} & {\fontsize{6.5}{7.5}\selectfont$\pm$0.10} & {\fontsize{6.5}{7.5}\selectfont$\pm$0.07} & {\fontsize{6.5}{7.5}\selectfont$\pm$0.008} \\[1pt]
\hspace{0.5em}Pretrained, no fine-tuning & 14.81 & 0.208 & 34.81 & 0.309 & 19.15 & 0.250 & 46.78 & 0.393 & 0.57 & 0.21 & 0.253 \\
 & {\fontsize{6.5}{7.5}\selectfont$\pm$0.26} & {\fontsize{6.5}{7.5}\selectfont$\pm$0.001} & {\fontsize{6.5}{7.5}\selectfont$\pm$1.97} & {\fontsize{6.5}{7.5}\selectfont$\pm$0.008} & {\fontsize{6.5}{7.5}\selectfont$\pm$0.89} & {\fontsize{6.5}{7.5}\selectfont$\pm$0.003} & {\fontsize{6.5}{7.5}\selectfont$\pm$0.83} & {\fontsize{6.5}{7.5}\selectfont$\pm$0.003} & {\fontsize{6.5}{7.5}\selectfont$\pm$0.07} & {\fontsize{6.5}{7.5}\selectfont$\pm$0.09} & {\fontsize{6.5}{7.5}\selectfont$\pm$0.001} \\[1pt]
\multicolumn{12}{@{}l}{\textit{Past-conditioned decoder}} \\
\hspace{0.5em}Reconstruction & 14.00 & 0.176 & 31.45 & 0.270 & 16.99 & 0.198 & 44.61 & 0.366 & 0.49 & 0.15 & 0.212 \\
 & {\fontsize{6.5}{7.5}\selectfont$\pm$0.14} & {\fontsize{6.5}{7.5}\selectfont$\pm$0.002} & {\fontsize{6.5}{7.5}\selectfont$\pm$0.56} & {\fontsize{6.5}{7.5}\selectfont$\pm$0.006} & {\fontsize{6.5}{7.5}\selectfont$\pm$0.64} & {\fontsize{6.5}{7.5}\selectfont$\pm$0.003} & {\fontsize{6.5}{7.5}\selectfont$\pm$1.13} & {\fontsize{6.5}{7.5}\selectfont$\pm$0.006} & {\fontsize{6.5}{7.5}\selectfont$\pm$0.08} & {\fontsize{6.5}{7.5}\selectfont$\pm$0.07} & {\fontsize{6.5}{7.5}\selectfont$\pm$0.008} \\[1pt]
\hspace{0.5em}Inverse dynamics & {\boldmath\bfseries 13.07} & {\boldmath\bfseries 0.160} & 27.92 & {\boldmath\bfseries 0.236} & {\boldmath\bfseries 13.44} & {\boldmath\bfseries 0.162} & 39.39 & 0.333 & 0.57 & 0.51 & 0.155 \\
 & {\fontsize{6.5}{7.5}\selectfont$\pm$0.24} & {\fontsize{6.5}{7.5}\selectfont$\pm$0.001} & {\fontsize{6.5}{7.5}\selectfont$\pm$1.17} & {\fontsize{6.5}{7.5}\selectfont$\pm$0.005} & {\fontsize{6.5}{7.5}\selectfont$\pm$0.28} & {\fontsize{6.5}{7.5}\selectfont$\pm$0.001} & {\fontsize{6.5}{7.5}\selectfont$\pm$2.01} & {\fontsize{6.5}{7.5}\selectfont$\pm$0.005} & {\fontsize{6.5}{7.5}\selectfont$\pm$0.07} & {\fontsize{6.5}{7.5}\selectfont$\pm$0.09} & {\fontsize{6.5}{7.5}\selectfont$\pm$0.001} \\[1pt]
\hspace{0.5em}Latent prediction & 14.28 & 0.169 & {\boldmath\bfseries 27.54} & 0.257 & 15.64 & 0.177 & {\boldmath\bfseries 39.08} & {\boldmath\bfseries 0.331} & {\boldmath\bfseries 0.62} & {\boldmath\bfseries 0.67} & {\boldmath\bfseries 0.046} \\
 & {\fontsize{6.5}{7.5}\selectfont$\pm$0.18} & {\fontsize{6.5}{7.5}\selectfont$\pm$0.001} & {\fontsize{6.5}{7.5}\selectfont$\pm$0.67} & {\fontsize{6.5}{7.5}\selectfont$\pm$0.002} & {\fontsize{6.5}{7.5}\selectfont$\pm$0.26} & {\fontsize{6.5}{7.5}\selectfont$\pm$0.001} & {\fontsize{6.5}{7.5}\selectfont$\pm$1.08} & {\fontsize{6.5}{7.5}\selectfont$\pm$0.007} & {\fontsize{6.5}{7.5}\selectfont$\pm$0.07} & {\fontsize{6.5}{7.5}\selectfont$\pm$0.07} & {\fontsize{6.5}{7.5}\selectfont$\pm$0.001} \\[1pt]
\bottomrule
\end{tabular}
\endgroup

%% file: sections/discussion.tex
\section{Discussion}
Our results show that temporal structure in the latent space of video world models yields practical benefits for robot manipulation. World models trained on temporally structured latents generalize better out of distribution, extrapolate better to horizons longer than those seen during training, and give policy evaluations that agree better with the real world. They are also cheaper: they match the generation quality of the baseline with five times fewer sampling steps, and its perceptual quality with over seven times less training.
Temporal structure should therefore be treated as an important design criterion for the latent space of video world models.

Furthermore, our results show that two lines of work can enter a fruitful dialogue: video world models meant as simulators, and reconstruction-free world models meant for latent planning.
Indeed, a world model can keep generating realistic rollouts, which powerful VLM judges can score and policies can act in, while its latents gain the temporal structure that makes them easy to predict from the past.
On complex real-world data, a simple hierarchy helped to reconcile these two goals: the past-conditioned decoder takes the details that do not change over time from a past frame, so the latents only need to carry what changes.
In the same spirit, it would be interesting to explore learning deeper hierarchies of dynamical representations.
Such hierarchies could enable flexible inference that trades temporal and state resolution for a longer horizon at a fixed computational budget.

\paragraph{Limitations.}
Our experiments cover synthetic data and videos of robot manipulation.
A natural next step is to extend them to other domains where video world models are used, such as egocentric video, navigation and games.
We focused on 2D autoencoders that only compress along the spatial axes, for simplicity. Spatiotemporal autoencoders also compress along the time axis; this is different from, and complementary to, our temporal objectives, and could be considered in future work.
Finally, we used temporal objectives to fine-tune general purpose autoencoders rather than train them from scratch. Pretraining the autoencoders on large video corpora, potentially using latent actions to leverage raw video~\citep{schmidtLearningActActions2023,bruceGenieGenerativeInteractive2024b,garridoLearningLatentAction2026a}, could enable further benefits.

%% file: sections/ai_statement.tex
\subsection*{AI use statement}
In this work, we used generative AI tools for method implementation: writing and editing the research code, including the code that produces the figures and tables.
We have not used generative AI tools for theoretical or conceptual development, hypotheses, methodology or experiment design, synthetic-data generation, dataset cleaning or reformatting, qualitative analysis, or interpretation of results, and mathematical claims or proofs and translation are not applicable to this work.
Additionally, we used generative AI tools for literature search and citation checking, for drafting parts of the appendix, and for editing the manuscript for readability.
A vision-language model, Claude Opus~5.5, is also a component of one evaluation: it judges task success in the policy evaluation of Section~\ref{sec:matched-compute} (Appendix~\ref{app:worldgym}).
We have reviewed all AI-assisted work: AI-assisted code was verified through tests and experiment-level checks, citations were checked against the cited works, and every edit to the manuscript was reviewed by the authors.
We take responsibility for the final content of this work, including text, claims or artifacts produced with the aid of generative AI.

%% file: sections/appendix/guide.tex
\section*{Guide to the appendix}
Appendix~\ref{app:toy} specifies the synthetic study of Section~\ref{sec:toy} and gives its full results.
For the video experiments, Appendices~\ref{app:objectives} to~\ref{app:evaluation} describe the objectives, architectures, training and evaluation.
Appendix~\ref{app:matched-rec} and Table~\ref{tab:settings} give the settings specific to the experiments of Sections~\ref{sec:matched-rec} and~\ref{sec:matched-compute}, which we call the equal-reconstruction and the equal-budget experiments; the experiment of Section~\ref{sec:tradeoff} is the weight sweep.
Appendix~\ref{app:results} contains the additional results: controls, complete results and learning curves, the autoencoder losses over training, interventions on the past-conditioned decoder, policy evaluation, qualitative examples, longer world-model training, the latent spaces and ablations.

%% file: sections/appendix/synthetic.tex
\section{Synthetic study}
\label{app:toy}

\paragraph{Environment.}
A state is $x_t = (s_t, \xi_t)$, with a signal $s_t \in \R^2$ and distractors $\xi_t \in \R^{100}$, and the observation is the state itself.
At the start of a trajectory, $s_0$ and $\xi_0$ are drawn from a standard normal distribution.
At every step, an action is drawn uniformly among the four vectors $\pm 0.3\, e_1$ and $\pm 0.3\, e_2$, the signal moves by it, $s_{t+1} = s_t + a_t$, and $\xi_{t+1}$ is a new standard normal draw, independent of the past.
A trajectory has two transitions.
The training, validation and test sets hold 10,000, 1,000 and 1,000 trajectories, that is, 20,000 training transitions.

\paragraph{Networks.}
Every network is a multilayer perceptron with one hidden layer of 256 units and ReLU activations, except the probe, whose hidden layer has 64 units.
The encoder maps the observation to the mean and the log-variance of a Gaussian posterior over $z \in \R^d$.
The training loss uses a sample from the posterior. All downstream networks use its mean, as the experiments on real video do.
The decoder of $\mathcal{L}_{\mathrm{rec}}$ maps a latent back to $\R^{102}$, and the key network $g$ of $\mathcal{L}_{\mathrm{temp}}$ maps the next observation and the one-hot action to $\R^d$.
The dynamics model maps a latent and the one-hot action to a latent, the probe maps a latent to the signal, and the readout decoder maps a latent to the observation.
These three networks are the world model, the probe and the decoder $D$ of Figure~\ref{fig:toy}.

\paragraph{Objectives.}
The reconstruction term is the squared error averaged over the 102 coordinates, plus a KL penalty with weight $10^{-4}$: the KL divergence from the posterior to the standard normal distribution, summed over the latent coordinates.
The temporal term is the InfoNCE loss.
Its query is the standardized current latent $q_t = \mathrm{LN}(z_t)$ and its keys are $y = \mathrm{LN}(g(x_{t+1}, a_t))$, where $\mathrm{LN}$ is layer normalization without affine parameters, and a query and a key are scored by $-\lVert q_t - y \rVert^2 / (\tau d)$ with temperature $\tau = 1$.
Standardizing fixes the norms of queries and keys at $\sqrt{d}$, so the score equals $-2(1-\cos\phi)/\tau$, with $\phi$ the angle between them.
The score thus depends only on this angle, and the temperature transfers across latent sizes.
The negatives are the keys of the other 255 transitions of the batch, drawn from the whole training set and thus, with rare exceptions, from other trajectories.
They penalize latents that do not distinguish observations, which rules out a collapsed latent as a minimizer of the temporal term~\citep{wangUnderstandingContrastiveRepresentation2020}.
The temperature and the KL weight were selected on the validation next-signal error and are fixed across the grid.
At $\lambda = 1$ the loss is the contrastive objective alone, and at $\lambda = 0$ the reconstruction and KL terms alone, which keep their relative weight at every $\lambda$.

\paragraph{Relation to the objectives on real video.}
The temporal term is the contrastive objective of Appendix~\ref{app:objectives} in its simplest form.
Its score is a squared distance between standardized vectors, without the bilinear form, the momentum encoder and the queue.
The roles of query and key are also exchanged.
Here the current latent is the query, and a separate network computes the keys from the next observation and the action~\citep{nachumProvableRepresentationLearning2021a}.
On real video, a prediction from the current latent is scored against encoded future frames.
The reconstruction terms differ because the data do: a squared error with a KL penalty here, and the pixel and LPIPS terms of Appendix~\ref{app:objectives} on real video.

\paragraph{Training.}
Training runs in four phases: first the encoder, with the decoder of $\mathcal{L}_{\mathrm{rec}}$ and the key network $g$, then the dynamics model, the probe and the readout decoder, each on the latents of the frozen encoder.
We train a separate readout decoder rather than reuse the decoder of $\mathcal{L}_{\mathrm{rec}}$, which receives no gradient at $\lambda = 1$, so that reconstruction is measured in the same way at every $\lambda$.
Each phase runs 7,500 updates of AdamW at batch size 256 and learning rate $10^{-3}$, without weight decay, and the encoder phase clips the gradient norm at 1.
We use the final weights; apart from the selection of the temperature and the KL weight, the validation set serves for monitoring only.
The dynamics model minimizes the squared error to the latent of the next observation, the probe the squared error to the signal, and the readout decoder the squared error to the observation.

\paragraph{Evaluation.}
All errors are squared errors on the test transitions, averaged over the target coordinates and over transitions; lower is better.
The reconstruction error compares the readout decoder's output on $z_t$ with $x_t$, and the current-signal error compares the probe's output on $z_t$ with $s_t$.
The next-latent error compares the dynamics model's prediction from $z_t$ and $a_t$ with $z_{t+1}$, and the next-signal error compares the probe's output on that prediction with $s_{t+1}$.
Each error is divided by the variance of its target on the test set of the same seed, so a constant prediction of the mean scores 1.
The target of the next-latent error is the latent itself, whose scale is learned and differs across latent sizes and objectives.
We estimate the variance of the latent as half the mean squared distance between the latents of two independently drawn test observations, averaged over coordinates.
Figure~\ref{fig:toy} divides each error of an encoder by the error of the raw observation on the same seed, which shares its data, and shows the geometric mean of this ratio over seeds with the standard error of its logarithm.

\begin{figure}[htbp]
    \centering
    \includegraphics[width=\textwidth]{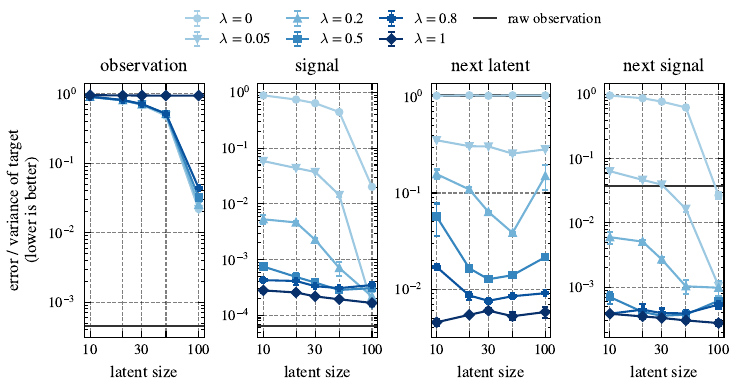}
    \caption{\textbf{The synthetic study at every latent size and weight.} Errors on the observation, the signal, the next latent and the next signal against the latent size, for all six weights $\lambda$ of the contrastive temporal term ($\lambda = 0$: reconstruction alone; $\lambda = 1$: the temporal term alone). Horizontal lines: the raw observation given directly to the same downstream networks (in the next-latent panel, hidden under $\lambda = 0$). Mean and standard error over three seeds. Intermediate weights retain reconstruction information while improving signal recovery and latent predictability over reconstruction alone.}
    \label{fig:toy-full}
\end{figure}

\paragraph{Grid and replication.}
The grid crosses the latent sizes $d \in \{10, 20, 30, 50, 100\}$ with the weights $\lambda \in \{0, 0.05, 0.2, 0.5, 0.8, 1\}$.
Section~\ref{sec:toy} shows $\lambda \in \{0, 0.5, 1\}$, and Figure~\ref{fig:toy-full} shows every weight.
With the raw observation ($d = 102$) as the reference, the grid has 31 configurations.
Each is repeated with three seeds, which set the data draw and every initialization, so the replicates are independent draws of the whole experiment (93 runs).
A run takes a few minutes on 8 CPU cores.

%% file: sections/appendix/objectives.tex
\FloatBarrier
\section{Temporal objectives}
\label{app:objectives}

The autoencoder is trained on pairs of frames $x_t$ and $x_{t+1}$, with latents $z_t = E(x_t)$ and $z_{t+1} = E(x_{t+1})$, and the action $a_t$ executed between them.
In the equal-reconstruction experiment, a third of the pairs are two frames apart, and the predictor then receives both actions between them.
The three forward objectives score the prediction $P(z_t, a_t)$ of the predictor of Appendix~\ref{app:architectures}.

The reconstruction loss $\mathcal{L}_{\mathrm{rec}}$ is the mean absolute pixel error plus VGG-based LPIPS~\citep{zhangUnreasonableEffectivenessDeep2018}, with weight one on each term.
The pixel term is averaged over channels and positions on images in $[0,1]$, and LPIPS receives the images mapped to $[-1,1]$.
All losses are averaged over examples.
We multiply the loss of Equation~\ref{eq:hybrid} by two, which gives both terms unit weight at $\lambda = 1/2$; with AdamW, the factor matters only through gradient clipping.
Unless stated otherwise, no adversarial or KL term is used, and the latent is the mean of the encoder's posterior.

\paragraph{Latent prediction.}
The prediction is regressed onto the future latent,
\begin{equation}
    \mathcal{L}_{\mathrm{latent}} = \frac{1}{d_z}\big\lVert P(z_t, a_t) - \operatorname{sg}[z_{t+1}] \big\rVert_1,
    \label{eq:latent}
\end{equation}
where $d_z$ is the number of latent coordinates (16,384 for FLUX.1; the synthetic study calls its latent size $d$) and $\operatorname{sg}$ is the stop-gradient operator.
The target comes from the same encoder, without a momentum copy.

\paragraph{Pixel prediction.}
The prediction is decoded and compared with the future frame,
\begin{equation}
    \mathcal{L}_{\mathrm{pixel}} = \mathcal{L}_{\mathrm{rec}}\big(x_{t+1},\, D(P(z_t, a_t))\big),
    \label{eq:pixel}
\end{equation}
with the same decoder and reconstruction loss as for the current frame.
The gradient reaches the encoder, the predictor and the decoder.

\paragraph{Contrastive prediction.}
The prediction is trained to identify the future frame among alternatives with the InfoNCE loss.
Queries $q$ and keys $y$ are unit-normalized projections of the prediction and of frames encoded by a momentum copy of the encoder, compared through a learned bilinear form $W$:
\begin{equation}
    \mathcal{L}_{\mathrm{contr}} = -\log \frac{\exp(q^\top W y^{+})}{\exp(q^\top W y^{+}) + \sum_{y^{-} \in \mathcal{Y}^{-}} \exp(q^\top W y^{-})},
    \label{eq:contr}
\end{equation}
where $y^{+}$ is the key of $x_{t+1}$ and $\mathcal{Y}^{-}$ contains the keys of the future frames of the other videos in the batch and a queue of 40,960 keys from recent batches.
A momentum copy of the projection (Appendix~\ref{app:architectures}) is applied to the keys, together with the momentum encoder, with decay 0.9978.
$W$ is initialized at the identity, so the score starts as a cosine similarity; the temperature is one, and the learned $W$ also sets the scale of the scores.

\paragraph{Inverse dynamics.}
A head $H$ predicts the action between two consecutive frames from their latents,
\begin{equation}
    \mathcal{L}_{\mathrm{inv}} = \ell\big(H(z_t, z_{t+1}),\, a_t\big),
    \label{eq:inv}
\end{equation}
where $\ell$ averages over the seven action components: the squared error on the six translation and rotation components, standardized with the mean and standard deviation of the training actions, and the binary cross-entropy on the gripper state.
Both latents receive gradients.

\paragraph{Reconstruction target within a pair.}
With the ordinary decoder, the earlier frame of the pair is reconstructed from its own latent, $D(z_t) \approx x_t$, and the future latent $z_{t+1}$ is computed without gradient.
With the past-conditioned decoder, the later frame is reconstructed from its latent and the earlier frame, $D(z_{t+1}, x_t) \approx x_{t+1}$.
The latent of the earlier frame then receives no reconstruction gradient, and the latent of the later frame receives no gradient from the forward objectives.

%% file: sections/appendix/architectures.tex
\FloatBarrier
\section{Architectures}
\label{app:architectures}

We describe the networks of the FLUX-based video experiments. The DINOv2 ablation changes some of them (Appendix~\ref{app:dino}).

\paragraph{Autoencoder.}
Unless stated otherwise, the video experiments initialize both the encoder and the decoder from the pretrained autoencoder of FLUX.1-schnell~\citep{blackforestlabsFLUX2024}.
We use its variational autoencoder, with 16 latent channels and a spatial downsampling factor of eight: a $256\times256$ frame gives a $16\times32\times32$ latent.
The encoder and the decoder are fine-tuned during representation learning and frozen during world-model training; the pretrained-autoencoder control skips fine-tuning.

\paragraph{Predictor.}
Two $1 \times 1$ convolutions map the 16 channels of the latent to a width of 512 and back, around two residual blocks.
Each block applies a depthwise-separable convolution with kernel 3, self-attention over the $32 \times 32$ positions with two-dimensional rotary position embeddings~\citep{suRoFormerEnhancedTransformer2024} and 8 heads, cross-attention to the action tokens, and a feed-forward layer with expansion factor 4 and GELU activation.
The sublayers use group normalization with 32 groups and residual connections, without dropout.
Each action is embedded by a two-layer perceptron of width 512 with GELU activation, plus a learned position embedding.
Together with a learned summary token, these embeddings pass through a two-layer transformer of width 512 with 8 heads, feed-forward width 2,048, GELU activations and layer normalization before each sublayer and at the output.
Its outputs, for the actions and for the summary token, are the keys and values of the cross-attention.
The predictor has 15.6~M parameters, 6.6~M of them in the action encoder.

\paragraph{Inverse-dynamics head.}
The two latents are concatenated along channels and processed by blocks of the same design, without cross-attention; the output is averaged over positions and mapped to the seven action components by a two-layer perceptron.
The head has 7.1~M parameters.

\paragraph{Contrastive head.}
The projection applied to the prediction and, through its momentum copy, to the keys is a single residual $7 \times 7$ convolution that keeps the shape of the latent.
The bilinear form $W$ of Equation~\ref{eq:contr} acts on the flattened projections.

\paragraph{Past-conditioned decoder.}
The pretrained decoder (49.5~M parameters, all fine-tuned) receives five cross-attention adapters with 8 heads, after the middle block and after each of the four upsampling blocks (3.7~M parameters).
The context is a $32\times32$ grid of 1,024 tokens from the past frame.
At each position, an $8\times8$ pixel patch and the corresponding middle-block features of a frozen copy of the pretrained encoder, the context encoder (34.3~M parameters, not trained) are projected separately to width 512, concatenated, and projected back to width 512 (0.9~M parameters).
Each adapter projects these context tokens to the channel width of its decoder block for keys and values, and uses the decoder features as queries.
The adapters use group normalization with 32 groups and two-dimensional rotary position embeddings on queries and keys, without dropout, and the output projection of every adapter is zero-initialized.
In total, the adapters and the projections add 4.6~M trained parameters.
The context features depend only on the context frame, which stays fixed during a rollout (Appendix~\ref{app:evaluation}), so they need to be computed only once per rollout.
Per generated frame, the decoder adds only the cross-attention adapters to the pretrained decoder.

\paragraph{World-model transformer.}
Each standardized latent frame is patchified with patch size 2 into 256 tokens.
Every block consists of two transformer layers, one with bidirectional self-attention within a frame and one with causal self-attention across frames at each spatial position, each followed by a feed-forward layer.
Both attentions use rotary position embeddings.
Adaptive layer normalization with zero initialization modulates every layer from the sum of the noise-level embedding and a linear embedding of the frame's action, so the modulation is specific to each frame, as diffusion forcing requires.
The action $a_t$ reaches frame $t+1$ through the causal attention across frames.
The final layer predicts the velocity of every token.
Table~\ref{tab:settings} gives the width, depth and number of attention heads in each experiment.

%% file: sections/appendix/training.tex
\FloatBarrier
\section{Training details}
\label{app:training}

\paragraph{Data.}
Bridge V2 is recorded with a WidowX 250 robot arm at 5\,Hz~\citep{walkeBridgeDataV2Dataset2023}.
We use the TensorFlow Datasets release \texttt{bridge\_dataset}, version 1.0.0, with its training and validation splits: 53,192 training videos and 6,872 held-out videos.
Frames are decoded from the stored videos, scaled to $[0, 1]$ and resized to $256 \times 256$ without cropping.
For the FLUX encoder, they are then mapped to $[-1,1]$.
Appendix~\ref{app:dino} gives the preprocessing for DINOv2.

\paragraph{Actions.}
The action $a_t$ has seven components: the translation and the rotation of the end effector, rescaled by fixed factors and clipped, and the gripper state, $-1$ when open and $+1$ when closed.
A further flag marks the null action, which replaces the action when the conditioning is dropped.

\paragraph{Training examples.}
Each epoch draws one pair of frames per training video for autoencoder training, and one clip of ten consecutive frames per training video for world-model training, both with a uniformly distributed start frame.
Videos with fewer than ten frames are left out of world-model training.
Table~\ref{tab:settings} gives the remaining settings of the two main experiments.

\begin{table}[htbp]
    \caption{\textbf{Training settings.} The ablations use the settings of the equal-budget experiment unless Appendix~\ref{app:ablations} states otherwise.}
    \label{tab:settings}
    \centering
    \smallskip
    \begingroup\fontsize{8}{9.5}\selectfont\setlength{\tabcolsep}{4pt}
    \begin{tabular}{@{}>{\raggedright\arraybackslash}p{2.9cm}>{\raggedright\arraybackslash}p{4.9cm}>{\raggedright\arraybackslash}p{5.4cm}@{}}
    \toprule
     & Equal reconstruction (Section~\ref{sec:matched-rec}) & Equal budget and weight sweep (Sections~\ref{sec:tradeoff} and~\ref{sec:matched-compute}) \\
    \midrule
    \multicolumn{3}{@{}l}{\textit{Autoencoder fine-tuning}} \\
    Frame pairs & consecutive, or two frames apart with probability 1/3 & consecutive \\
    Batch & 24 pairs & 48 pairs \\
    Learning rate & $2.4 \times 10^{-6}$ & $3.46 \times 10^{-6}$ \\
    Length & until the reconstruction target of Appendix~\ref{app:matched-rec} & 80,000 updates; final checkpoint \\
    Validation pairs & one per held-out video, redrawn at each check & a fixed set, one per held-out video \\
    Optimizer & \multicolumn{2}{>{\raggedright\arraybackslash}p{10.5cm}@{}}{AdamW~\citep{loshchilovDecoupledWeightDecay2019} with $\beta = (0.9, 0.999)$ and no weight decay; 1,000-update linear warm-up; gradient clipping at norm 0.5} \\
    \midrule
    \multicolumn{3}{@{}l}{\textit{World model}} \\
    Width, blocks, heads & 1,024, 16, 16 & 384, 12, 6 \\
    Parameters & 608~M & 64.4~M \\
    Batch & 4 clips & 16 clips (8 per step, two accumulation steps) \\
    Learning rate & $3 \times 10^{-5}$, constant & $7 \times 10^{-5}$ after a 3,000-update linear warm-up from $10^{-7}$ \\
    Gradient clipping & none & at norm 0.5 \\
    Updates & 238,896 (18 epochs) & 100,000 \\
    Latent standardization & one global standard deviation & per-channel mean and standard deviation \\
    Optimizer & \multicolumn{2}{>{\raggedright\arraybackslash}p{10.5cm}@{}}{AdamW with $\beta = (0.9, 0.99)$ and weight decay 0.02; exponential moving average of the weights with decay 0.999, used for sampling} \\
    \midrule
    Precision & \multicolumn{2}{>{\raggedright\arraybackslash}p{10.5cm}@{}}{bf16 mixed precision for both models} \\
    \bottomrule
    \end{tabular}
    \endgroup
\end{table}

\paragraph{Diffusion process.}
Each latent is corrupted independently as $z_t^{k_t} = \sqrt{\bar\alpha_{k_t}}\, z_t + \sqrt{1 - \bar\alpha_{k_t}}\, \epsilon_t$, with $\epsilon_t \sim \gN(0, I)$, $k_t$ uniform on $\{1, \dots, K\}$, and a variance-preserving schedule $1 > \bar\alpha_1 > \dots > \bar\alpha_K \approx 0$~\citep{hoDenoisingDiffusionProbabilistic2020}.
A network $v_\theta$ receives the noisy latents, their noise levels, and the actions, and regresses the velocity $v_t = \sqrt{\bar\alpha_{k_t}}\, \epsilon_t - \sqrt{1 - \bar\alpha_{k_t}}\, z_t$~\citep{salimansProgressiveDistillationFast2022} of every frame:
\begin{equation}
    \mathcal{L}_{\mathrm{DF}}(\theta) = \E \left[ \sum_{t=1}^{T} \left\| v_\theta\big(z_{1:T}^{k_{1:T}}, k_{1:T}, a_{1:T}\big)_t - v_t \right\|_2^2 \right].
    \label{eq:df}
\end{equation}
The diffusion process has $K = 1{,}000$ noise levels with a sigmoid schedule~\citep{jabriScalableAdaptiveComputation2023}, whose argument runs from $-3$ to $3$ with temperature one; the per-step noise variances are clipped to $[0,0.999]$.
In practice, the squared velocity error is averaged, rather than summed, over latent coordinates and frames, and averaged over examples.
During training, the action is replaced by the null action with probability 0.1, which allows classifier-free guidance~\citep{hoClassifierFreeDiffusionGuidance2021} at sampling time.
The DINOv2 ablation changes the parameterization and the noise-level sampling (Appendix~\ref{app:dino}).

\paragraph{Latent standardization.}
After fine-tuning, the statistics of the encoder output are computed on 100 batches of training frames, one random frame per clip (Table~\ref{tab:settings} gives the standardization of each experiment).
The world model is trained and sampled in the standardized space, and the transformation is inverted before decoding.

\paragraph{Compute.}
Autoencoders and world models are trained on NVIDIA H200 GPUs, one per run except the width-1,024 world models of the size ablation, which use two.
At the equal-budget settings, fine-tuning an autoencoder takes about 31 hours with the ordinary decoder and about 42 hours with the past-conditioned decoder.
In the equal-reconstruction experiment it takes from 1.3 hours (reconstruction alone) to about 30 hours (latent prediction).
Training a world model takes about 17 hours at width 384, and about 37 hours for the larger world models of the equal-reconstruction experiment.
Evaluating a world model in the four conditions of Table~\ref{tab:evaluation-protocol} takes 3 to 5 hours, or about 10 hours in the equal-reconstruction experiment.

%% file: sections/appendix/evaluation.tex
\FloatBarrier
\section{Evaluation details}
\label{app:evaluation}

\paragraph{Evaluation sets.}
In distribution, we evaluate on 2,000 clips of ten frames, one from each of 2,000 randomly chosen held-out videos, with a uniformly distributed start frame; every model is evaluated on the same clips.
The sampling-step sweep uses the first 512 of these clips.
The long-rollout evaluation uses 716 held-out videos: all 224 videos with at least 60 frames, together with 512 videos drawn uniformly among those with at least 20 frames, 20 of which are also among the 224.
Each is rolled out once, for up to 59 predicted frames.
Out of distribution, we use all 479 AutoEval trajectories, recorded on separate robot setups that reproduce four Bridge V2 tasks: closing the drawer (80 trajectories), opening the drawer (94), placing the eggplant in the sink (193) and placing it in the basket (112).
Frames are resized to $256 \times 256$, and the rollout takes the 21st frame of each trajectory as the context frame and predicts the following 29 frames.
The recorded actions are rescaled as in Appendix~\ref{app:training}; the gripper command is binarized and mapped to the sign convention of Bridge V2, which is opposite to that of AutoEval.

\paragraph{Sampling.}
Table~\ref{tab:evaluation-protocol} gives the four evaluation conditions.
DDIM uses the stated number of steps on a uniform grid over the 1,000 noise levels, without added noise, with classifier-free guidance of scale 3 at every step.
The DINOv2 ablation warps the grid (Appendix~\ref{app:dino}).
For a chunk of eight future frames, all eight are appended as noise and denoised jointly at staggered noise levels, each frame starting at a higher level than the previous one: the pyramid schedule of Section~\ref{sec:background}.
The last chunk can be shorter.
The past-conditioned decoder conditions on the initial real context frame throughout the rollout.
Within each comparison, every model uses the same diffusion noise on the same clips, with one rollout per clip and sampling seed.

\begin{table}[!htbp]
    \caption{\textbf{Evaluation conditions.} Each rollout starts from one real context frame, and the scored prediction steps count from the first frame after it. The denoising chunk is the largest number of future frames sampled jointly. The 2-step columns of the result tables come from the sampling-step sweep at 2 steps.}
    \label{tab:evaluation-protocol}
    \centering
    \smallskip
    \begingroup\fontsize{8}{9.5}\selectfont\setlength{\tabcolsep}{4pt}
    \begin{tabular}{@{}lrrrrr@{}}
    \toprule
    Condition & Clips & \shortstack{Scored prediction\\steps} & \shortstack{Denoising\\chunk} & \shortstack{DDIM\\steps} & \shortstack{Sampling\\seeds} \\
    \midrule
    In distribution & 2,000 & 1--9 & 8 & 10 & 3 \\
    Out of distribution & 479 & 1--29 & 1 & 10 & 3 \\
    Sampling-step sweep & 512 & 1--9 & 1 & 2, 4, 6, 8, 10 & 1 \\
    Long rollouts & 716 & 1--59 & 8 & 10 & 1 \\
    \bottomrule
    \end{tabular}
    \endgroup
\end{table}

\paragraph{Metrics.}
Pixel mean squared error (MSE) is computed per frame over channels and pixels in $[0, 1]$.
SSIM uses a Gaussian window of $11 \times 11$ pixels with standard deviation 1.5.
LPIPS uses VGG16 features with the learned linear layers of \citet{zhangUnreasonableEffectivenessDeep2018}, in the implementation of \citet{esserTamingTransformersHighResolution2021}, on inputs in $[-1, 1]$; the same network enters the reconstruction loss of every autoencoder with the same weight.
In and out of distribution, each framewise metric is averaged over the generated frames of a clip, then over clips, then over the three sampling seeds, which gives one value per trained pipeline.
Out-of-distribution values are the average of the four task means, which weights the tasks equally.
The sampling-step sweep and the long rollouts use one sampling seed and the frame windows given below.
FID~\citep{heuselGANsTrainedTwo2017} uses the 2,048-dimensional pool features of Inception-v3 on the nine generated frames of the 2,000 clips, 18,000 frames on each side.
FVD~\citep{unterthinerAccurateGenerativeModels2018} uses the 400-dimensional outputs of an I3D network~\citep{carreiraQuoVadisAction2017} trained on Kinetics-400, with the weights released by \citet{songHistoryGuidedVideoDiffusion2025}, on the 2,000 clips of nine frames.
Both are computed in distribution only, on the same clips for every training and sampling seed.
As for the framewise metrics, we average the three sampling-seed values of each trained pipeline.

\paragraph{Long rollouts and sampling steps.}
In the long-rollout evaluation, the value of a metric at a prediction step is the mean over the videos that reach that step, so every model is scored on the same videos at each step.
Their number falls from 716 at steps 1 to 19 to 224 at steps 57 to 59.
A long rollout is summarized by the mean of this value over prediction steps 51 to 59, which 229 to 224 videos reach.
The sampling-step sweep generates the nine frames of each of the 512 clips one frame at a time, with 2, 4, 6, 8 and 10 DDIM steps under the same noise.

\paragraph{Unpredictable share.}
We fit the same linear predictor to every frozen autoencoder, including those at $\lambda = 0$, on the 2,000 in-distribution clips (18,000 consecutive-frame transitions), with the latents standardized per channel as in Appendix~\ref{app:training}.
We use five-fold cross-fitting over clips, keeping every transition of a clip in one fold.
For each held-out fold, let $\mu$ be the mean latent over the frames of the other four folds, and let $u_t$ be the orthonormal two-dimensional Hartley transform of $z_t - \mu$, applied to each channel.
We predict each frequency coordinate as
\begin{equation}
    \hat{u}_{t+1,j}=b_j u_{t,j}+w_j^\top c_t,
\end{equation}
where $c_t$ concatenates the 256 leading principal coordinates of the centered current latent and the action.
The first term lets each frequency coordinate persist from one frame to the next.
The principal directions are fitted on the four training folds, and so are $b_j$ and $w_j$, jointly by least squares with a ridge penalty on $w_j$ whose strength is selected on an inner split of those folds.
The reported share is
\begin{equation}
    \frac{\sum_{(t,f)}\|z_{t+1}-\hat{z}_{t+1}^{(-f)}\|_2^2}
         {\sum_{(t,f)}\|z_{t+1}-\mu^{(-f)}\|_2^2},
\end{equation}
where the sums run over all held-out transitions, $\hat{z}_{t+1}^{(-f)}$ is the prediction mapped back from the frequency basis, and $(-f)$ denotes a fit that excludes the fold of the transition.
The denominator thus measures the variance of the target latents around the training-fold mean, not the variance of their frame-to-frame differences.

\paragraph{Latent MSE.}
From one context frame and the recorded action, the world model predicts the next latent with 10 DDIM steps and no guidance.
The latent MSE is the mean squared difference to the encoder's latent of the recorded next frame, divided by the pooled variance of the target latents.
It is computed on the first transition of 256 in-distribution clips, and on one transition of each of 256 out-of-distribution trajectories, 64 per task.

\paragraph{Training-curve evaluation.}
The training-speed factors of Section~\ref{sec:matched-compute} use the validation curves of the reconstruction pipelines and the temporal objectives, averaged over their three training seeds.
These runs validate once per epoch (3,318 updates) on 400 fixed held-out clips, generating nine frames from one context frame, one frame per denoising chunk, with 10 DDIM steps and guidance scale 3.
The world models of the equal-reconstruction experiment validate every three epochs (39,816 updates; Figure~\ref{fig:benefits-matched-rec}).
The context frame is excluded from the metrics, except in the equal-reconstruction experiment, whose validation averages over all ten frames.
For each metric, the reference value is the final mean validation value of the baseline, at 99,540 updates (30 epochs).
We find the first crossing of this value by another mean curve, interpolate linearly in log(update) between the adjacent validation points, and divide 99,540 by the crossing update.

\paragraph{Controls.}
The reconstruction ceiling encodes and decodes the recorded frames of every evaluation clip with each autoencoder, in the same way as the rollouts, and applies the same metrics.
Past-conditioned decoders use the initial real context frame, as in the rollouts.
The shuffled-action control repeats every rollout with the action sequence of another clip, assigned by a fixed permutation without fixed points that is shared by every model.
Out of distribution, the donor clip comes from the same task.
The shuffled rollout uses the same diffusion noise as the rollout under the recorded actions.
Both controls use one sampling seed.

\paragraph{Statistics.}
The equal-reconstruction and equal-budget experiments use three training seeds each.
The weight sweep fine-tunes its own autoencoders with one training seed, and with a second one at $\lambda = 0$ and 0.3.
Its world models at $\lambda = 0.1$ to 0.5 use one seed.
At $\lambda = 0$ and 0.7, the sweep reuses the world models of the equal-budget experiment.
The seed also fixes the order of the training data, so the objectives of an experiment see the same data in the same order.
The unit of replication is the trained pipeline, that is, the autoencoder with its world model; sampling seeds and evaluation clips are repeated measurements of one pipeline.
We report the mean and the standard error over training seeds.

%% file: sections/appendix/equal_reconstruction.tex
\FloatBarrier
\section{Equal-reconstruction experiment}
\label{app:matched-rec}

\paragraph{Pipelines.}
The four pipelines differ only in the objective used to fine-tune the autoencoder, its weight, and hence the number of updates needed to reach the target.
The data, the decoder, the world model, its training and the evaluation are shared (Appendices~\ref{app:architectures} to~\ref{app:evaluation} and Table~\ref{tab:settings}).
The decoder is initialized from FLUX.1-schnell and fine-tuned to reconstruct each frame from its latent alone.
The temporal weight $\lambda$ of Equation~\ref{eq:hybrid} is set per objective so that fine-tuning reaches the reconstruction target: 0.3 for latent and contrastive prediction and 0.5 for pixel prediction.

\paragraph{Stopping rule.}
Fine-tuning stops at the first end of an epoch (2,215 updates) at which the reconstruction error on the held-out pairs is below 0.0345.
The update counts of Table~\ref{tab:matched-rec} therefore exceed the number of updates needed to reach the target by less than one epoch.
At the stop, the held-out reconstruction error is 0.0342 for reconstruction alone and 0.0343 to 0.0345 for the temporal objectives.
The principal-component counts of Table~\ref{tab:matched-rec} are computed on the flattened latents of 1,000 held-out frames from 100 videos (Appendix~\ref{app:latent}).

%% file: sections/appendix/additional_results.tex
\FloatBarrier
\section{Additional results}
\label{app:results}

\subsection{Controls}
\label{app:controls}

Two controls test whether the generated trajectories depend on the recorded actions, and how much of the error remains when only the autoencoder is used.
The shuffled-action control and the reconstruction ceiling of Appendix~\ref{app:evaluation} were measured on the four pipelines of the equal-reconstruction experiment and on the three past-conditioned pipelines of the equal-budget experiment.
Under the actions of another clip, the in-distribution pixel MSE rises by 98 to 102\% in the equal-reconstruction experiment and by 62 to 77\% in the equal-budget experiment.
Out of distribution, with a donor clip from the same task, it rises by 21 to 26\% and by 20 to 23\%.
Encoding and decoding the recorded frames gives a pixel MSE ($\times 10^3$) of 0.19 to 0.32 in the equal-reconstruction experiment and 0.14 to 0.43 in the equal-budget experiment, against 11 to 34 for the world models on the same autoencoders.
The world models therefore use the action sequence, and autoencoding alone does not account for their rollout error.

\subsection{Complete results and learning curves}
\label{app:full-tables}

Tables~\ref{tab:matched-rec-full} and~\ref{tab:matched-compute-full} complete Tables~\ref{tab:matched-rec} and~\ref{tab:matched-compute} with SSIM and the distribution metrics, and Figures~\ref{fig:benefits-matched-rec} and~\ref{fig:benefits} show the errors against the number of sampling steps, along long rollouts and during world-model training.

\subsubsection{Equal reconstruction}

\begin{figure}[htbp]
    \centering
    \includegraphics[width=\textwidth]{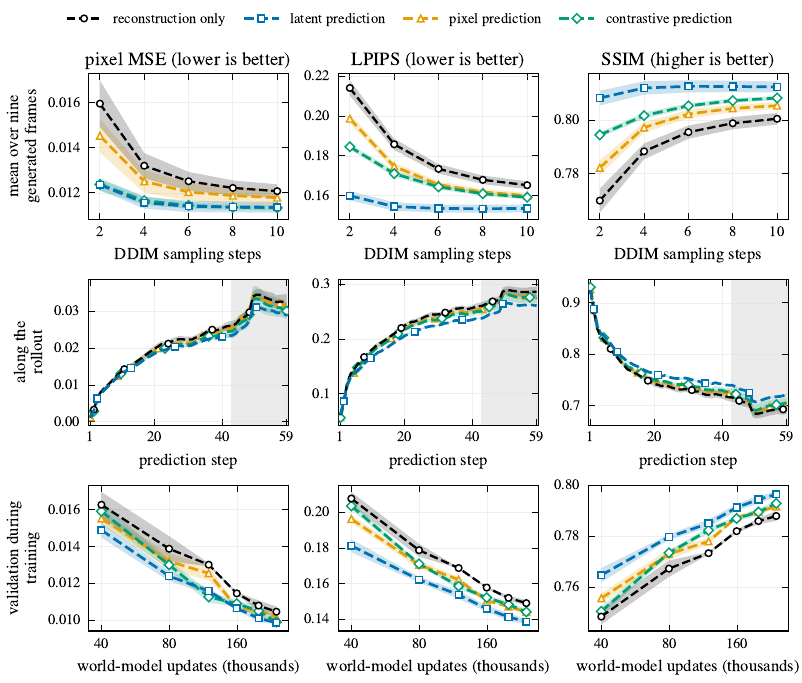}
    \caption{\textbf{World models at equal reconstruction quality, across sampling steps, rollout steps and training.} The four pipelines of Table~\ref{tab:matched-rec}, all with the ordinary decoder, against the number of DDIM sampling steps (top), along the long rollouts (middle) and over world-model training (bottom). Gray shading: prediction steps reached by fewer than half of the videos. Mean and standard error over three training seeds. Latent prediction gives the largest perceptual gains with few sampling steps, and the lowest LPIPS from the fourth prediction step on and throughout training.}
    \label{fig:benefits-matched-rec}
\end{figure}

Latent prediction gives the best perceptual error in every condition, with the largest gains when the sampler has only two steps, and the lowest FID (Table~\ref{tab:matched-rec-full}); contrastive prediction gives a slightly lower FVD.
Figure~\ref{fig:benefits-matched-rec} shows that the gains of latent prediction are largest with few sampling steps.
Its LPIPS is also the lowest from the fourth prediction step to the end of the long rollouts, and at every validation point of world-model training.

\begin{table}[htbp]
    \caption{\textbf{Complete results at equal reconstruction quality.} Table~\ref{tab:matched-rec} without its autoencoder columns, and with SSIM and the distribution metrics added; all pipelines use the ordinary decoder. Mean and standard error over three training seeds (pixel MSE $\times 10^3$); bold: best in each column. ID, OOD, 2-step and long-horizon as in Section~\ref{sec:frontier}. Latent prediction gives the best perceptual metrics, especially with two sampling steps, and the lowest FID; contrastive prediction gives a slightly lower FVD.}
    \label{tab:matched-rec-full}
    \centering
    \smallskip
    \input{figures/tab_matched_reconstruction/table_full.tex}
\end{table}

\FloatBarrier
\subsubsection{Equal training budget}

Both temporal objectives improve on both fine-tuned reconstruction pipelines in LPIPS and SSIM in all four conditions, and in pixel MSE out of distribution, with two sampling steps and on long rollouts (Table~\ref{tab:matched-compute-full}).
For latent prediction, the temporal weight sets a trade-off.
At $\lambda = 0.7$, which favors long rollouts (Appendix~\ref{app:weight}), its FID and FVD are higher than those of reconstruction with the same decoder, and its in-distribution pixel MSE is slightly higher (14.28 against 14.00).
Lower weights avoid this cost in FID and FVD.

\begin{table}[htbp]
    \caption{\textbf{Complete results at equal training budget.} Table~\ref{tab:matched-compute} without its policy-evaluation columns (Table~\ref{tab:worldgym}), and with SSIM, the out-of-distribution latent MSE and the distribution metrics added. Mean and standard error over three training seeds (pixel MSE $\times 10^3$); bold: best in each column. ID, OOD, 2-step and long-horizon as in Section~\ref{sec:frontier}. Both temporal objectives improve LPIPS and SSIM over both fine-tuned reconstruction pipelines in all four conditions; at $\lambda = 0.7$, latent prediction has a higher FID and FVD than reconstruction with the same decoder, which lower weights avoid (Appendix~\ref{app:weight}).}
    \label{tab:matched-compute-full}
    \centering
    \smallskip
    \input{figures/tab_matched_compute/table_full.tex}
\end{table}

\begin{figure}[htbp]
    \centering
    \includegraphics[width=\textwidth]{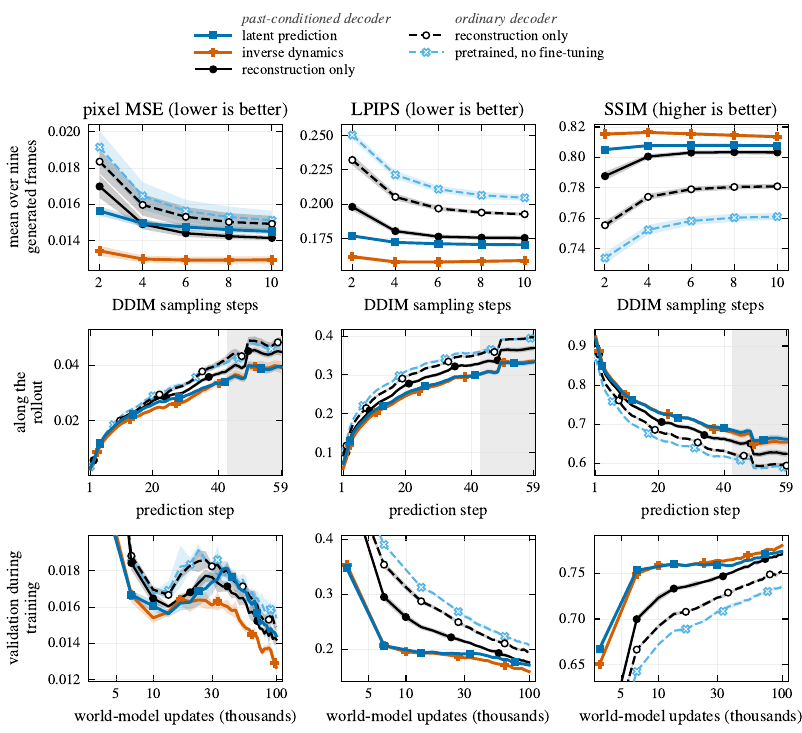}
    \caption{\textbf{World models at equal training budget, across sampling steps, rollout steps and training.} The pipelines of Table~\ref{tab:matched-compute} against the number of DDIM sampling steps (top), along the long rollouts (middle) and over world-model training (bottom). Gray shading: prediction steps reached by fewer than half of the videos. Mean and standard error over three training seeds. The temporal objectives improve LPIPS and SSIM in all three rows, and inverse dynamics also gives a lower pixel MSE than either reconstruction-only pipeline.}
    \label{fig:benefits}
\end{figure}

Figure~\ref{fig:benefits} shows the equal-budget world models against the number of sampling steps, along long rollouts and during training.
The temporal world models lose less quality when the number of sampling steps is reduced, and keep a lower perceptual error along the rollout.
They also reach the baseline's final validation LPIPS earlier (Appendix~\ref{app:evaluation} defines this comparison).
The jump shared by all models near steps 47 to 49, here and in Figure~\ref{fig:benefits-matched-rec}, reflects the number of evaluated videos falling from 322 to 233.

\FloatBarrier
\subsection{Autoencoder losses over training}
\label{app:frontier-training}

\begin{figure}[htbp]
    \centering
    \includegraphics{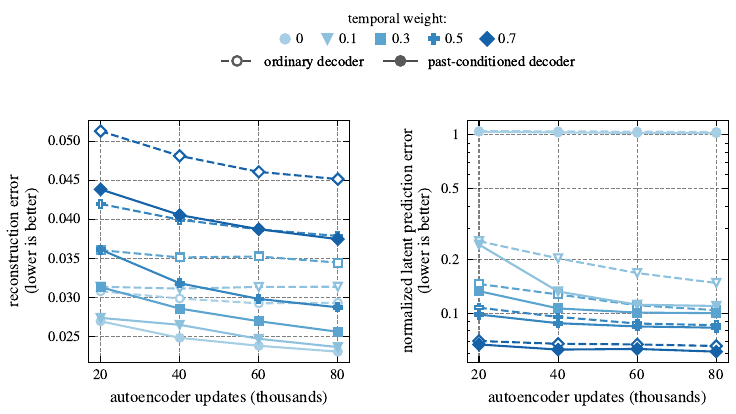}
    \caption{\textbf{Autoencoder losses over fine-tuning in the weight sweep.} Reconstruction error (left) and error of the jointly trained latent predictor (right), for each temporal weight and decoder. One training seed; the predictor receives no gradient at $\lambda = 0$. Past conditioning lowers both errors at every nonzero weight, and larger weights trade reconstruction for a lower prediction error.}
    \label{fig:tradeoff-training}
\end{figure}

Figure~\ref{fig:tradeoff-training} tracks two errors during autoencoder fine-tuning in the weight sweep: the reconstruction error (pixel mean absolute error plus LPIPS) and the error of the jointly trained predictor, relative to the mean magnitude of the latent.
This prediction error differs from the unpredictable share of Figure~\ref{fig:tradeoff}, which comes from a separate linear predictor fitted to each frozen latent space (Appendix~\ref{app:evaluation}).
At every nonzero weight and throughout training, the past-conditioned decoder lowers both the reconstruction error (by 13 to 26\%) and the prediction error (by 3 to 35\%).
Increasing the weight lowers the prediction error and raises the reconstruction error.

\FloatBarrier
\subsection{Interventions on the past-conditioned decoder}
\label{app:interventions}

To see whether the improvement from past conditioning depends on the information in the context frame, we evaluate the autoencoders of the weight sweep after 80,000 updates in three ways (Table~\ref{tab:interventions}): with the correct past frame, with the context pathway disabled, and with past frames permuted within each evaluation batch.
The error is the reconstruction error of fine-tuning (pixel mean absolute error plus LPIPS) on the 6,872 held-out frame pairs.
Disabling the context bypasses the context encoder and the cross-attention adapters, and with the permutation, each target latent is decoded with the past frame of another example.
Disabling or replacing the context raises the reconstruction error at every weight, most at $\lambda = 0.7$.
This is the behavior the design of Section~\ref{sec:tradeoff} aims for: the decoder takes from the past frame what the latent does not carry, and does so more at larger weights.
The same pattern holds for a second training seed at weights 0 and 0.3.

\begin{table}[htbp]
    \caption{\textbf{Dependence of the past-conditioned decoder on its context.} Reconstruction error of the past-conditioned autoencoders of the weight sweep with three contexts; the last column gives the autoencoder trained with the ordinary decoder at the same weight. One training seed. The correct context lowers the error at every weight, and the dependence on the context is largest at $\lambda = 0.7$.}
    \label{tab:interventions}
    \centering
    \smallskip
    \begingroup\fontsize{8}{9.5}\selectfont
    \begin{tabular}{@{}lcccc@{}}
    \toprule
     & \multicolumn{4}{c}{Reconstruction error $\downarrow$} \\
    \cmidrule(lr){2-5}
    $\lambda$ & Past frame & Context disabled & Permuted past frame & Ordinary decoder \\
    \midrule
    0   & 0.0231 & 0.0407 & 0.0394 & 0.0294 \\
    0.1 & 0.0237 & 0.0430 & 0.0424 & 0.0314 \\
    0.3 & 0.0257 & 0.0496 & 0.0481 & 0.0345 \\
    0.5 & 0.0288 & 0.0700 & 0.0565 & 0.0379 \\
    0.7 & 0.0375 & 0.1947 & 0.1072 & 0.0451 \\
    \bottomrule
    \end{tabular}
    \endgroup
\end{table}

\FloatBarrier
\subsection{Policy evaluation in WorldGym}
\label{app:worldgym}

WorldGym measures whether a world model preserves the differences in policy performance measured on the real robot.
The evaluation is closed-loop: each policy chooses its next action from the frames that the world model generates, so different world models can induce different action sequences.

\begin{figure}[h]
    \centering
    \includegraphics[width=\textwidth]{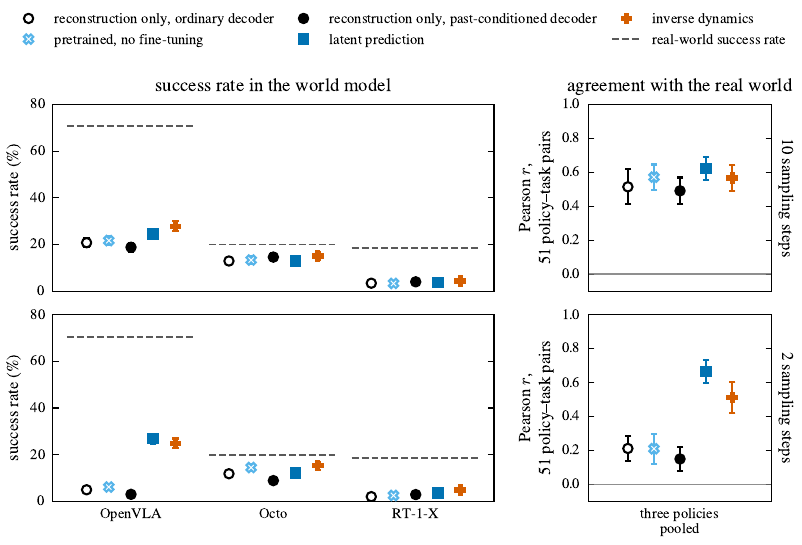}
    \caption{\textbf{Policy evaluation inside the world models.} Three robot policies on the 17 Bridge tasks of WorldGym, inside the five world models of Table~\ref{tab:matched-compute}. One training seed; error bars: standard error over the 170 starting frames (left) and bootstrap standard error over tasks (right). With 2 sampling steps, the temporal world models keep their agreement with the real world and the real ranking of the three policies, while the reconstruction-only world models lose both.}
    \label{fig:worldgym}
\end{figure}

We run the Bridge suite of WorldGym~\citep{quevedoWorldGymWorldModel2025}: 17 tasks with 10 starting frames each.
From each starting frame, a policy acts for 40 frames inside the world model, three times (once in WorldGym).
The world models are the five pipelines of Table~\ref{tab:matched-compute}, with one training seed, at 100,000 updates.
They sample with classifier-free guidance of scale 3 and 10 or 2 sampling steps.
The real-world reference is the per-task success rate of each policy reported by \citet{kimOpenVLAOpenSourceVisionLanguageAction2025}, as used by WorldGym.

\paragraph{Judge.}
The judge is Claude Opus 5.5 at high reasoning effort, in place of WorldGym's GPT-4o, with WorldGym's prompt and, as in WorldGym's released code, frames 0, 20 and 40 of a rollout.
It answers 1 for success, 0 for failure, and 0.5 for partial success on the tasks whose criteria define one.
It is called three times per rollout (five times in WorldGym).
The most frequent answer is the outcome, with ties going to the higher score, and a success rate is the mean outcome, so a partial success counts one half, as in the real-world rates.
We checked the judge on 48 real robot episodes of the AutoEval evaluation set with human success labels, six successes and six failures from each of its four tasks.
With three calls per episode, as in the rollouts, the majority vote agrees with the labels on 42 episodes: all 24 successes and 18 of the 24 failures.
The six errors are failures judged as successes.

\paragraph{Uncertainty.}
The bootstrap resamples the 17 tasks with replacement, 10,000 times, keeping the three policies of a task together.
The standard error of a correlation, Pearson or Spearman, is the standard deviation of its resampled values.
Differences between two world models are computed on the same resampled tasks, with their standard errors and two-sided $p$-values.
These quantities describe the sampling of the tasks for one trained world model; unlike the standard errors of the other evaluations, they do not include the variability between training seeds.

\paragraph{Results.}
Table~\ref{tab:worldgym} gives the agreement per policy and, for the pooled agreement, the difference from the baseline.
With 10 sampling steps, latent prediction has the highest pooled correlation with real-world success.
With 2 steps, both temporal world models keep a much higher pooled agreement than the reconstruction-only world models.
For the reconstruction-only world models, the correlation within each policy changes by at most 0.14 from 10 to 2 steps, while the pooled one falls by 0.30 to 0.36.
The drop therefore comes from the comparison between policies: at 2 steps, these world models rank Octo above OpenVLA, the reverse of the real-world ranking, whereas both temporal world models keep the real ranking of the three policies (Figure~\ref{fig:worldgym}).
Appendix~\ref{app:longer} evaluates the world models continued to 200,000 updates.

\begin{table}[htbp]
    \caption{\textbf{Agreement with real-world policy evaluation at 100,000 updates.} Pearson $r$ and Spearman $\rho$ between simulated and real success rates over the 17 tasks of each policy, or over the 51 policy--task pairs when pooled; $\Delta r$ and $p$: paired difference from the baseline (--) and its two-sided bootstrap $p$-value. One training seed; bootstrap standard errors over tasks under the values. With two sampling steps, both temporal world models agree with real-world performance better than all reconstruction-only world models when the three policies are pooled.}
    \label{tab:worldgym}
    \centering
    \smallskip
    \input{figures/fig_worldgym/table.tex}
\end{table}

\FloatBarrier
\input{sections/appendix/qualitative}

\FloatBarrier
\subsection{Longer world-model training}
\label{app:longer}

The three past-conditioned world models of the equal-budget experiment were continued from 100,000 to 200,000 updates, with the same settings, for two of its training seeds.
Table~\ref{tab:longer} compares them at 200,000 updates, together with the same two seeds at 100,000 updates.
Against reconstruction with the same decoder, inverse dynamics keeps its gains in all four evaluation conditions.
Latent prediction extends its gain out of distribution and keeps a lower LPIPS and a higher SSIM than reconstruction in all four conditions.
On long rollouts its pixel-MSE gain is smaller than at 100,000 updates, and its in-distribution pixel MSE, FID and FVD are higher than those of reconstruction.

\begin{table}[htbp]
    \caption{\textbf{Longer world-model training.} The three past-conditioned pipelines of the equal-budget experiment at 200,000 world-model updates and, for the same two training seeds, at 100,000. Mean and standard error over two training seeds (pixel MSE $\times 10^3$); bold: best in each column at each training length. ID, OOD, 2-step and long-horizon as in Section~\ref{sec:frontier}. Inverse dynamics keeps its gains in every condition; latent prediction keeps a lower LPIPS than reconstruction in every condition and extends its out-of-distribution gain, but its in-distribution pixel MSE, FID and FVD are higher.}
    \label{tab:longer}
    \centering
    \smallskip
    \input{figures/tab_longer/table_full.tex}
\end{table}

\begin{figure}[htbp]
    \centering
    \includegraphics[width=\textwidth]{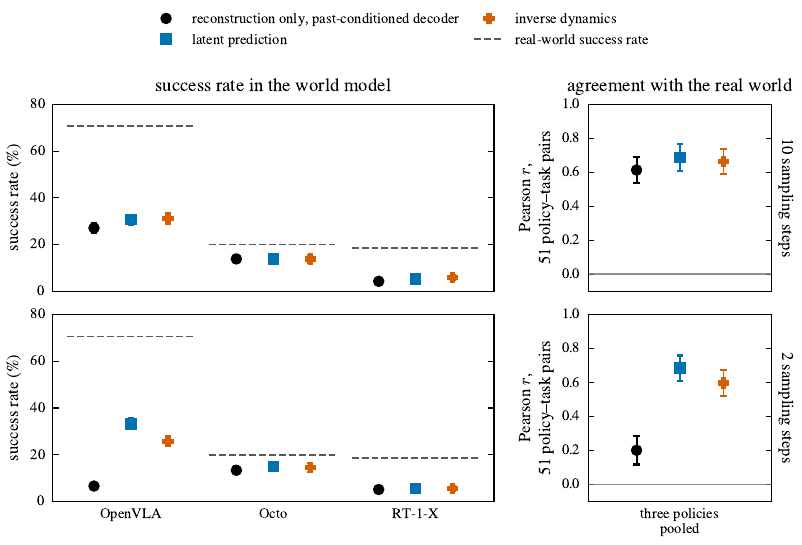}
    \caption{\textbf{Policy evaluation after 200,000 world-model updates.} The three past-conditioned world models of Table~\ref{tab:worldgym-200k}, one training seed; panels, rows and error bars as in Figure~\ref{fig:worldgym}. With 2 steps, the temporal world models keep the real ranking of the policies and their agreement with the real world.}
    \label{fig:worldgym-200k}
\end{figure}

\paragraph{Policy evaluation.}
The world models of one training seed at 200,000 updates were also evaluated in WorldGym, as in Appendix~\ref{app:worldgym} (Table~\ref{tab:worldgym-200k} and Figure~\ref{fig:worldgym-200k}).
The pattern at 100,000 updates holds.
With 10 sampling steps, both temporal world models have a higher pooled correlation with real-world success than reconstruction (0.69 and 0.66 against 0.61).
With 2 steps, the agreement of reconstruction drops (pooled $r$ from 0.61 to 0.20), while that of the temporal world models changes little (from 0.69 to 0.68 and from 0.66 to 0.60), and they keep the real ranking of the three policies.

\begin{table}[htbp]
    \caption{\textbf{Agreement with real-world policy evaluation at 200,000 updates.} The three past-conditioned world models of Table~\ref{tab:longer}, one training seed; columns as in Table~\ref{tab:worldgym}, with $\Delta r$ and $p$ against reconstruction (--). With two sampling steps, both temporal world models agree with real-world performance better than reconstruction when the three policies are pooled; with ten, the differences are smaller.}
    \label{tab:worldgym-200k}
    \centering
    \smallskip
    \input{figures/fig_worldgym/table_200k.tex}
\end{table}

\FloatBarrier
\subsection{Latent spaces}
\label{app:latent}

\begin{figure}[htbp]
    \centering
    \includegraphics[width=\textwidth]{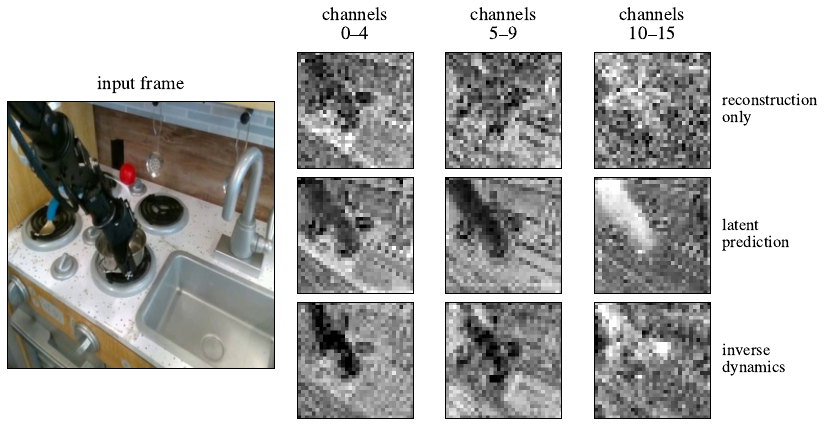}
    \caption{\textbf{Latent maps at equal training budget.} One held-out Bridge frame encoded by the encoders of the three past-conditioned pipelines, one training seed. Columns average channels 0--4, 5--9 and 10--15 of each $16\times32\times32$ latent. Each map is standardized separately, black low and white high, so intensities are not comparable across maps. Latent prediction and inverse dynamics produce smoother regions around the arm and the objects than reconstruction alone.}
    \label{fig:latent-maps}

    \medskip
    \includegraphics[width=\textwidth]{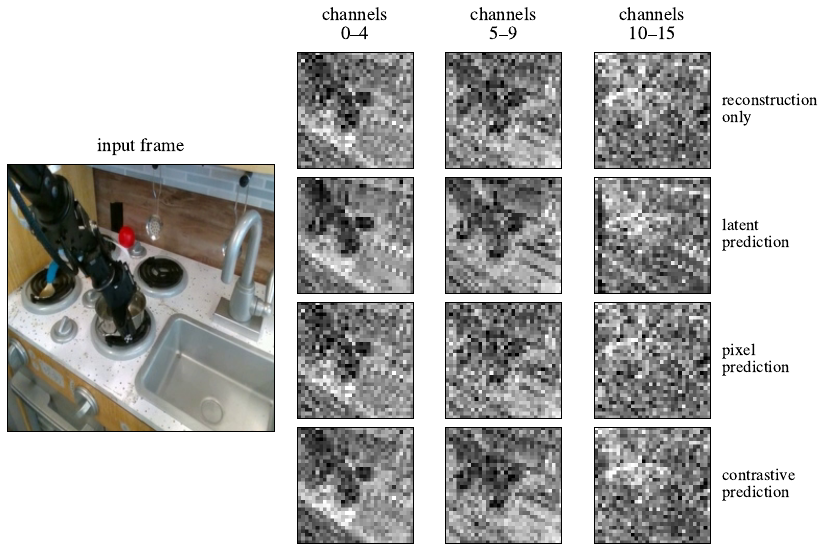}
    \caption{\textbf{Latent maps at equal reconstruction quality.} The same frame encoded by the four encoders of the equal-reconstruction experiment, all with the ordinary decoder, one training seed. Columns and standardization as in Figure~\ref{fig:latent-maps}. Latent and contrastive prediction suppress fine spatial variation compared with reconstruction alone, whereas pixel prediction retains it.}
    \label{fig:latent-maps-equal-rec}
\end{figure}

\begin{figure}[htbp]
    \centering
    \includegraphics[width=\textwidth]{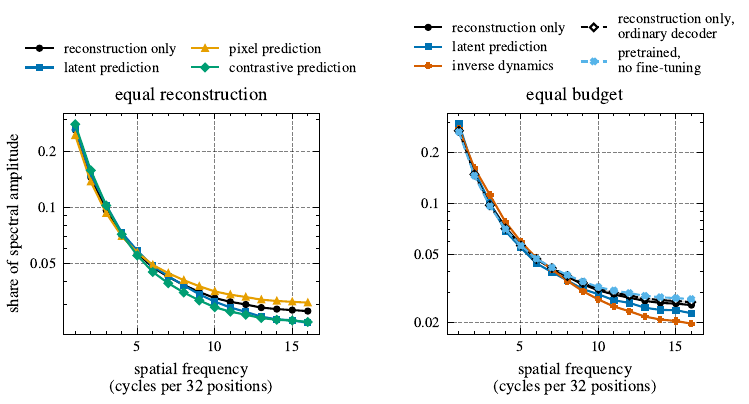}
    \caption{\textbf{Spatial spectra of the frozen latents.} Radial amplitude spectra of the latent channels, as a share of their total amplitude; at equal budget, solid curves use the past-conditioned decoder and dashed curves the ordinary one. Mean and standard error over three training seeds, except the pretrained encoder (one). Latent and contrastive prediction reduce the high-frequency components at equal reconstruction, and latent prediction and inverse dynamics do so at equal budget.}
    \label{fig:spectra}
\end{figure}

To describe what the temporal objectives change in the latent space, we encode 1,000 held-out frames with the frozen encoders of both experiments: ten consecutive frames from each of 100 videos, with a uniformly distributed start frame.
We measure four properties of the flattened latents.
Their dimensionality is the number of principal components for 95\% of the variance, together with the participation ratio of the spectrum.
The straightness of their trajectories is the mean angle between successive latent displacements along a video, divided by the same angle in pixel space.
Clustering by video is the mean distance between frames of one video divided by that between frames of different videos.
For the spatial smoothness of the latent maps, we average the magnitudes of the real two-dimensional Fourier transform of each latent channel over frames and channels.
With $A(\omega)$ the mean amplitude at rounded radial frequency $\omega$, in cycles per 32 latent positions, we report the least-squares slope of $\log A(\omega)$ against $\log \omega$ for $\omega=1,\ldots,16$, and the high-frequency share $\sum_{\omega=8}^{16}A(\omega)^2/\sum_{\omega=1}^{16}A(\omega)^2$.
Figure~\ref{fig:spectra} shows $A(\omega)$ divided by its sum over $\omega=1,\ldots,16$.
The two experiments use different frames, so their values are not compared with each other.
Table~\ref{tab:latent} gives the statistics, Figure~\ref{fig:spectra} the spectra, and Figures~\ref{fig:latent-maps} and~\ref{fig:latent-maps-equal-rec} the latent maps of a single frame.

\begin{table}[htbp]
    \caption{\textbf{Geometry of the frozen latent spaces.} Statistics of the latents in both experiments, defined in the text; lower values mean lower dimension, straighter trajectories, tighter clustering by video and smoother maps. Mean and standard error over three training seeds unless one is stated. Latent prediction reduces dimension and high-frequency content in both experiments, and inverse dynamics also smooths the latents.}
    \label{tab:latent}
    \centering
    \smallskip
    \input{figures/fig_latent_structure/table.tex}
\end{table}

Latent prediction changes the latent space in the same direction in both experiments.
It reduces the number of principal components needed for 95\% of the variance (177 against 388 for the baseline at equal reconstruction; at equal budget, 110 against 316 for reconstruction with the same decoder and 345 for the baseline), the participation ratio, the turning-angle ratio, the high-frequency content and the within-to-across-video distance ratio.
Contrastive prediction shifts every statistic in the same direction, and inverse dynamics every one except clustering by video.
Pixel prediction shifts them in the opposite direction, with 445 principal components.
With reconstruction alone, past conditioning produces smaller changes in the same direction as the temporal objectives (316 against 345 principal components).
The latent maps and spectra show the smoother spatial structure under the temporal objectives (Figures~\ref{fig:latent-maps} to~\ref{fig:spectra}).

\FloatBarrier
\subsection{Ablations}
\label{app:ablations}

The ablations use one training seed unless stated otherwise.

\subsubsection{Temporal weight}
\label{app:weight}
\begin{figure}[htbp]
    \centering
    \includegraphics[width=\textwidth]{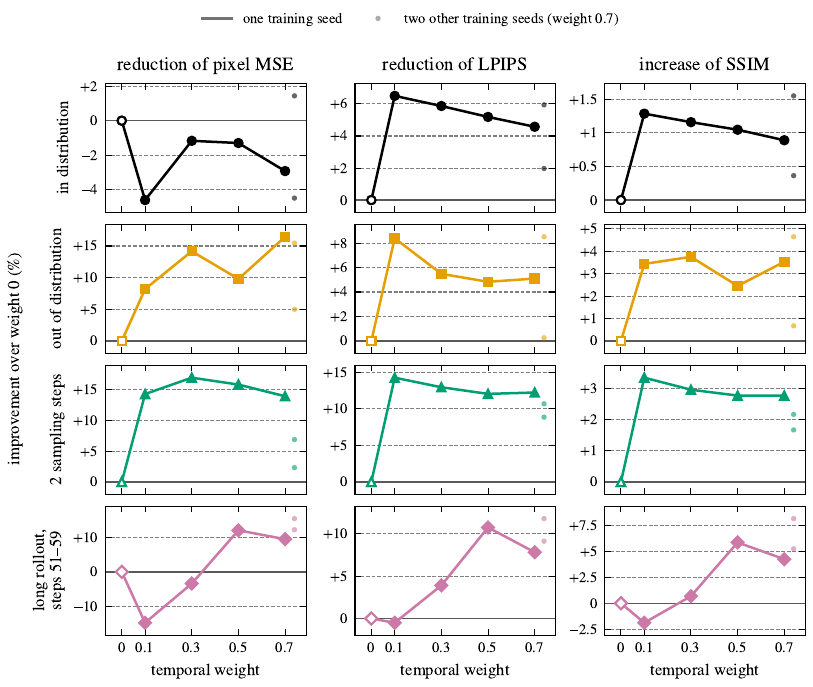}
    \caption{\textbf{World-model improvement over reconstruction as the temporal weight varies.} Latent prediction at each weight against reconstruction ($\lambda = 0$, zero by construction, open marker), both with the past-conditioned decoder and the same training seed; panels have independent scales. One training seed, and two more at $\lambda = 0.7$ (dots). LPIPS improves at every weight in distribution, out of distribution and with two sampling steps; long-rollout gains favor larger weights.}
    \label{fig:weight}
\end{figure}

\begin{figure}[htbp]
    \centering
    \includegraphics{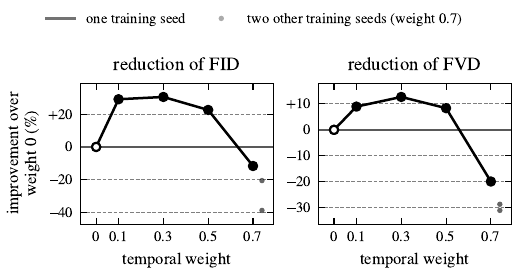}
    \caption{\textbf{Distribution metrics as the temporal weight varies.} FID and FVD of latent prediction at each weight against reconstruction ($\lambda = 0$), both with the past-conditioned decoder and the same training seed, in distribution; panels as in Figure~\ref{fig:weight}. One training seed, and two more at $\lambda = 0.7$ (dots). Weights from 0.1 to 0.5 improve both metrics; at $\lambda = 0.7$, both are worse than at $\lambda = 0$, for all three training seeds.}
    \label{fig:weight-dist}
\end{figure}

Figure~\ref{fig:weight} shows the world models on the past-conditioned autoencoders of the weight sweep at $\lambda = 0.1$ to 0.5, and those of the equal-budget experiment at $\lambda = 0.7$, against reconstruction with the same decoder ($\lambda = 0$).
Lower weights favor the distribution metrics and larger weights long rollouts: at $\lambda = 0.1$ to 0.5, latent prediction gives FID 12.8 to 14.3 and FVD 31.8 to 33.4, against 18.5 and 36.5 for reconstruction with the same decoder and seed (Figure~\ref{fig:weight-dist}).

\FloatBarrier
\subsubsection{A semantic encoder}
\label{app:dino}
We replace the FLUX encoder by DINOv2-Base~\citep{oquabDINOv2LearningRobust2024} with registers~\citep{darcetVisionTransformersNeed2024}, and use the pretrained ViT-XL decoder of RAE~\citep{zhengDiffusionTransformersRepresentation2025} with RAE's released latent statistics.
Each frame is resized to $224\times224$ with bicubic interpolation and normalized with the ImageNet statistics; the encoder produces a $768\times16\times16$ latent, and the decoder reconstructs at $256\times256$.
We make the decoder past-conditioned with seven zero-initialized cross-attention adapters with eight heads, after layers 1, 5, 9, 13, 17, 21 and 25 of its 28.
Their 256 context tokens, of width 768, combine projected image patches with projected features from a context encoder, a frozen copy of the pretrained DINOv2 encoder, as in the FLUX decoder.
The encoder, the decoder and the adapters are fine-tuned with the equal-budget settings (Table~\ref{tab:settings}), at latent-prediction weights 0, 0.1, 0.3 and 0.7.
The latent predictor is that of Appendix~\ref{app:architectures}, with 768 input and output channels.

The world model follows RAE for these latents.
It uses one token per latent position, a backbone whose width equals the token dimension (12 blocks of width 768, 12 heads), RAE's wide output head (two blocks of width 2,048, 16 heads), and standardization of every channel and position.
It also uses RAE's dimension-dependent shift of the noise level, $\kappa\mapsto 7\kappa/(1+6\kappa)$ for the normalized noise index $\kappa\in[0,1]$, in training and on the DDIM grid.
Unlike RAE, it predicts the clean latent, with the loss computed on the velocity~\citep{liBackBasicsLet2025}.
Its optimization, the evaluation sets, the guidance scale and the numbers of sampling steps are those of the equal-budget experiment.
At $\lambda = 0.1$, latent prediction improves every metric (Table~\ref{tab:dino}).
At $\lambda = 0.3$ and 0.7, latent prediction lowers the latent MSE further and still improves out of distribution, in every metric at 0.3 and in pixel MSE at 0.7.
At both weights, it worsens the framewise metrics in distribution, with two sampling steps and on long rollouts, and the distribution metrics.
With a semantic encoder, a small weight is therefore preferable.
This world model differs from those of the FLUX experiments, so the values of Table~\ref{tab:dino} are not compared with those of Table~\ref{tab:matched-compute-full}.

\begin{table}[htbp]
    \caption{\textbf{Latent prediction with a DINOv2 encoder.} DINOv2 encoder and past-conditioned RAE decoder; one training seed (pixel MSE $\times 10^3$); bold: best in each column. ID, OOD, 2-step and long-horizon as in Section~\ref{sec:frontier}. At $\lambda = 0.1$, latent prediction improves every metric; larger weights lower the latent MSE further but worsen most other errors and the distribution metrics.}
    \label{tab:dino}
    \centering
    \smallskip
    \input{figures/tab_dino/table_full.tex}
\end{table}

\FloatBarrier
\subsubsection{World-model size}
\begin{figure}[htbp]
    \centering
    \includegraphics[width=\textwidth]{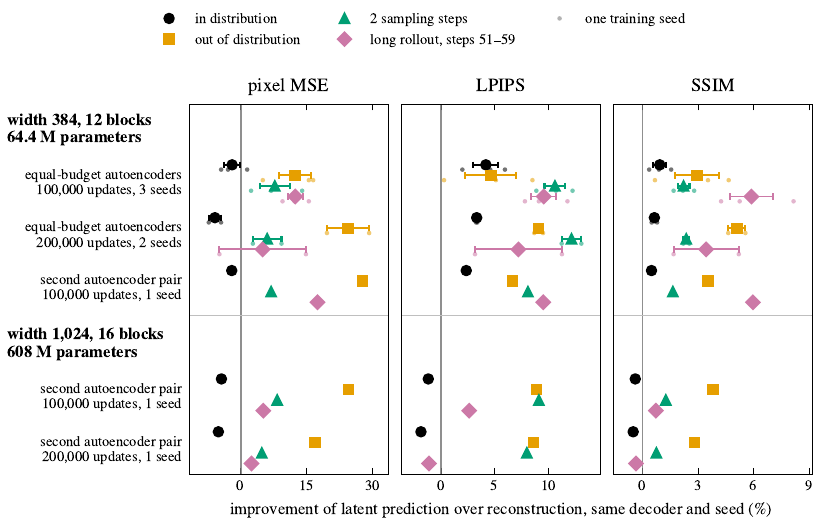}
    \caption{\textbf{Effect of latent prediction across world-model sizes.} Improvement of latent prediction over reconstruction, both with the past-conditioned decoder and the same seed: relative reductions of pixel MSE and LPIPS, and increases of SSIM. Rows of the equal-budget autoencoders show the mean and standard error over training seeds, with dots for single seeds; rows of the second autoencoder pair show one run each. The gains persist out of distribution and with two sampling steps at both widths; in distribution, the effect is mixed at width 384 and negative at width 1,024, where the long-rollout gains also shrink.}
    \label{fig:size}
\end{figure}

On a second pair of autoencoders, trained with the equal-budget settings (reconstruction, and latent prediction at $\lambda = 0.7$; one training run each), we train world models of width 384 with 12 blocks and of width 1,024 with 16 blocks (Figure~\ref{fig:size}).
Out of distribution and with two sampling steps, latent prediction improves the world model at both widths.
In distribution, its effect is mixed at width 384 and negative at width 1,024, where its long-rollout gains also shrink.

\FloatBarrier
\subsubsection{KL penalty}
Our autoencoders are fine-tuned without the KL penalty of latent diffusion (Appendix~\ref{app:objectives}).
We add it with weight $10^{-6}$ to the equal-budget settings, with and without latent prediction (Table~\ref{tab:kl}).
With the penalty, latent prediction still improves on reconstruction in every framewise metric and in FID, most in out-of-distribution pixel MSE.
Its FVD is higher.
The penalty itself improves reconstruction in every metric except the out-of-distribution pixel MSE.

\begin{table}[htbp]
    \caption{\textbf{KL penalty.} Reconstruction and latent prediction at $\lambda = 0.7$, both with the past-conditioned decoder, with and without a KL penalty on the autoencoder; one training seed (pixel MSE $\times 10^3$); bold: best in each column. ID, OOD, 2-step and long-horizon as in Section~\ref{sec:frontier}. With the penalty, latent prediction still improves every framewise metric and FID over reconstruction with the penalty, while its FVD is higher.}
    \label{tab:kl}
    \centering
    \smallskip
    \input{figures/tab_kl/table_full.tex}
\end{table}

%% file: figures/tab_matched_reconstruction/table_full.tex
\begingroup\fontsize{8}{9.5}\selectfont\setlength{\tabcolsep}{3pt}
\begin{tabular}{@{}l*{6}{c}@{}}
\toprule
 & \multicolumn{3}{c}{ID} & \multicolumn{3}{c}{OOD} \\
\cmidrule(lr){2-4}\cmidrule(lr){5-7}
 & MSE $\downarrow$ & LPIPS $\downarrow$ & SSIM $\uparrow$ & MSE $\downarrow$ & LPIPS $\downarrow$ & SSIM $\uparrow$ \\
\midrule
Reconstruction (baseline) & 11.65 & 0.164 & 0.801 & 33.87 & 0.308 & 0.675 \\
 & {\fontsize{6.5}{7.5}\selectfont$\pm$0.20} & {\fontsize{6.5}{7.5}\selectfont$\pm$0.001} & {\fontsize{6.5}{7.5}\selectfont$\pm$0.002} & {\fontsize{6.5}{7.5}\selectfont$\pm$1.28} & {\fontsize{6.5}{7.5}\selectfont$\pm$0.008} & {\fontsize{6.5}{7.5}\selectfont$\pm$0.009} \\[1pt]
Pixel prediction & 11.43 & 0.159 & 0.806 & 34.23 & 0.300 & 0.683 \\
 & {\fontsize{6.5}{7.5}\selectfont$\pm$0.22} & {\fontsize{6.5}{7.5}\selectfont$\pm$0.001} & {\fontsize{6.5}{7.5}\selectfont$\pm$0.001} & {\fontsize{6.5}{7.5}\selectfont$\pm$0.67} & {\fontsize{6.5}{7.5}\selectfont$\pm$0.002} & {\fontsize{6.5}{7.5}\selectfont$\pm$0.002} \\[1pt]
Contrastive prediction & {\boldmath\bfseries 11.17} & 0.160 & 0.807 & 31.23 & 0.303 & 0.681 \\
 & {\fontsize{6.5}{7.5}\selectfont$\pm$0.05} & {\fontsize{6.5}{7.5}\selectfont$\pm$0.001} & {\fontsize{6.5}{7.5}\selectfont$\pm$0.001} & {\fontsize{6.5}{7.5}\selectfont$\pm$0.14} & {\fontsize{6.5}{7.5}\selectfont$\pm$0.005} & {\fontsize{6.5}{7.5}\selectfont$\pm$0.005} \\[1pt]
Latent prediction & 11.30 & {\boldmath\bfseries 0.153} & {\boldmath\bfseries 0.814} & {\boldmath\bfseries 29.14} & {\boldmath\bfseries 0.298} & {\boldmath\bfseries 0.688} \\
 & {\fontsize{6.5}{7.5}\selectfont$\pm$0.10} & {\fontsize{6.5}{7.5}\selectfont$\pm$0.001} & {\fontsize{6.5}{7.5}\selectfont$\pm$0.001} & {\fontsize{6.5}{7.5}\selectfont$\pm$0.55} & {\fontsize{6.5}{7.5}\selectfont$\pm$0.005} & {\fontsize{6.5}{7.5}\selectfont$\pm$0.005} \\[1pt]
\midrule[\heavyrulewidth]
 & \multicolumn{3}{c}{2-step} & \multicolumn{3}{c}{Long-horizon} \\
\cmidrule(lr){2-4}\cmidrule(lr){5-7}
 & MSE $\downarrow$ & LPIPS $\downarrow$ & SSIM $\uparrow$ & MSE $\downarrow$ & LPIPS $\downarrow$ & SSIM $\uparrow$ \\
\midrule
Reconstruction (baseline) & 15.95 & 0.214 & 0.770 & 33.26 & 0.287 & 0.691 \\
 & {\fontsize{6.5}{7.5}\selectfont$\pm$1.03} & {\fontsize{6.5}{7.5}\selectfont$\pm$0.004} & {\fontsize{6.5}{7.5}\selectfont$\pm$0.005} & {\fontsize{6.5}{7.5}\selectfont$\pm$1.84} & {\fontsize{6.5}{7.5}\selectfont$\pm$0.008} & {\fontsize{6.5}{7.5}\selectfont$\pm$0.010} \\[1pt]
Pixel prediction & 14.53 & 0.199 & 0.782 & 32.63 & 0.279 & 0.698 \\
 & {\fontsize{6.5}{7.5}\selectfont$\pm$0.75} & {\fontsize{6.5}{7.5}\selectfont$\pm$0.002} & {\fontsize{6.5}{7.5}\selectfont$\pm$0.003} & {\fontsize{6.5}{7.5}\selectfont$\pm$0.69} & {\fontsize{6.5}{7.5}\selectfont$\pm$0.003} & {\fontsize{6.5}{7.5}\selectfont$\pm$0.004} \\[1pt]
Contrastive prediction & 12.38 & 0.185 & 0.794 & 31.50 & 0.277 & 0.701 \\
 & {\fontsize{6.5}{7.5}\selectfont$\pm$0.27} & {\fontsize{6.5}{7.5}\selectfont$\pm$0.001} & {\fontsize{6.5}{7.5}\selectfont$\pm$0.001} & {\fontsize{6.5}{7.5}\selectfont$\pm$2.62} & {\fontsize{6.5}{7.5}\selectfont$\pm$0.010} & {\fontsize{6.5}{7.5}\selectfont$\pm$0.011} \\[1pt]
Latent prediction & {\boldmath\bfseries 12.36} & {\boldmath\bfseries 0.160} & {\boldmath\bfseries 0.808} & {\boldmath\bfseries 30.04} & {\boldmath\bfseries 0.262} & {\boldmath\bfseries 0.716} \\
 & {\fontsize{6.5}{7.5}\selectfont$\pm$0.24} & {\fontsize{6.5}{7.5}\selectfont$\pm$0.002} & {\fontsize{6.5}{7.5}\selectfont$\pm$0.003} & {\fontsize{6.5}{7.5}\selectfont$\pm$0.43} & {\fontsize{6.5}{7.5}\selectfont$\pm$0.000} & {\fontsize{6.5}{7.5}\selectfont$\pm$0.001} \\[1pt]
\midrule[\heavyrulewidth]
 & \multicolumn{2}{c}{Distribution} \\
\cmidrule(lr){2-3}
 & FID $\downarrow$ & FVD $\downarrow$ \\
\midrule
Reconstruction (baseline) & 9.4 & 22.9 \\
 & {\fontsize{6.5}{7.5}\selectfont$\pm$0.2} & {\fontsize{6.5}{7.5}\selectfont$\pm$0.4} \\[1pt]
Pixel prediction & 9.3 & 23.2 \\
 & {\fontsize{6.5}{7.5}\selectfont$\pm$0.1} & {\fontsize{6.5}{7.5}\selectfont$\pm$0.2} \\[1pt]
Contrastive prediction & 9.1 & {\boldmath\bfseries 22.0} \\
 & {\fontsize{6.5}{7.5}\selectfont$\pm$0.2} & {\fontsize{6.5}{7.5}\selectfont$\pm$0.2} \\[1pt]
Latent prediction & {\boldmath\bfseries 8.3} & 22.7 \\
 & {\fontsize{6.5}{7.5}\selectfont$\pm$0.1} & {\fontsize{6.5}{7.5}\selectfont$\pm$0.8} \\[1pt]
\bottomrule
\end{tabular}
\endgroup

%% file: figures/tab_matched_compute/table_full.tex
\begingroup\fontsize{8}{9.5}\selectfont\setlength{\tabcolsep}{2.2pt}
\begin{tabular}{@{}l*{6}{c}@{}}
\toprule
 & \multicolumn{3}{c}{ID} & \multicolumn{3}{c}{OOD} \\
\cmidrule(lr){2-4}\cmidrule(lr){5-7}
 & MSE $\downarrow$ & LPIPS $\downarrow$ & SSIM $\uparrow$ & MSE $\downarrow$ & LPIPS $\downarrow$ & SSIM $\uparrow$ \\
\midrule
\multicolumn{7}{@{}l}{\textit{Ordinary decoder}} \\
\hspace{0.5em}Reconstruction (baseline) & 14.34 & 0.194 & 0.779 & 36.15 & 0.302 & 0.682 \\
 & {\fontsize{6.5}{7.5}\selectfont$\pm$0.16} & {\fontsize{6.5}{7.5}\selectfont$\pm$0.002} & {\fontsize{6.5}{7.5}\selectfont$\pm$0.001} & {\fontsize{6.5}{7.5}\selectfont$\pm$0.98} & {\fontsize{6.5}{7.5}\selectfont$\pm$0.003} & {\fontsize{6.5}{7.5}\selectfont$\pm$0.004} \\[1pt]
\hspace{0.5em}Pretrained, no fine-tuning & 14.81 & 0.208 & 0.758 & 34.81 & 0.309 & 0.673 \\
 & {\fontsize{6.5}{7.5}\selectfont$\pm$0.26} & {\fontsize{6.5}{7.5}\selectfont$\pm$0.001} & {\fontsize{6.5}{7.5}\selectfont$\pm$0.001} & {\fontsize{6.5}{7.5}\selectfont$\pm$1.97} & {\fontsize{6.5}{7.5}\selectfont$\pm$0.008} & {\fontsize{6.5}{7.5}\selectfont$\pm$0.009} \\[1pt]
\multicolumn{7}{@{}l}{\textit{Past-conditioned decoder}} \\
\hspace{0.5em}Reconstruction & 14.00 & 0.176 & 0.803 & 31.45 & 0.270 & 0.718 \\
 & {\fontsize{6.5}{7.5}\selectfont$\pm$0.14} & {\fontsize{6.5}{7.5}\selectfont$\pm$0.002} & {\fontsize{6.5}{7.5}\selectfont$\pm$0.003} & {\fontsize{6.5}{7.5}\selectfont$\pm$0.56} & {\fontsize{6.5}{7.5}\selectfont$\pm$0.006} & {\fontsize{6.5}{7.5}\selectfont$\pm$0.006} \\[1pt]
\hspace{0.5em}Inverse dynamics & {\boldmath\bfseries 13.07} & {\boldmath\bfseries 0.160} & {\boldmath\bfseries 0.813} & 27.92 & {\boldmath\bfseries 0.236} & {\boldmath\bfseries 0.745} \\
 & {\fontsize{6.5}{7.5}\selectfont$\pm$0.24} & {\fontsize{6.5}{7.5}\selectfont$\pm$0.001} & {\fontsize{6.5}{7.5}\selectfont$\pm$0.001} & {\fontsize{6.5}{7.5}\selectfont$\pm$1.17} & {\fontsize{6.5}{7.5}\selectfont$\pm$0.005} & {\fontsize{6.5}{7.5}\selectfont$\pm$0.006} \\[1pt]
\hspace{0.5em}Latent prediction & 14.28 & 0.169 & 0.810 & {\boldmath\bfseries 27.54} & 0.257 & 0.739 \\
 & {\fontsize{6.5}{7.5}\selectfont$\pm$0.18} & {\fontsize{6.5}{7.5}\selectfont$\pm$0.001} & {\fontsize{6.5}{7.5}\selectfont$\pm$0.001} & {\fontsize{6.5}{7.5}\selectfont$\pm$0.67} & {\fontsize{6.5}{7.5}\selectfont$\pm$0.002} & {\fontsize{6.5}{7.5}\selectfont$\pm$0.003} \\[1pt]
\midrule[\heavyrulewidth]
 & \multicolumn{3}{c}{2-step} & \multicolumn{3}{c}{Long-horizon} \\
\cmidrule(lr){2-4}\cmidrule(lr){5-7}
 & MSE $\downarrow$ & LPIPS $\downarrow$ & SSIM $\uparrow$ & MSE $\downarrow$ & LPIPS $\downarrow$ & SSIM $\uparrow$ \\
\midrule
\multicolumn{7}{@{}l}{\textit{Ordinary decoder}} \\
\hspace{0.5em}Reconstruction (baseline) & 18.35 & 0.232 & 0.756 & 47.93 & 0.394 & 0.595 \\
 & {\fontsize{6.5}{7.5}\selectfont$\pm$0.65} & {\fontsize{6.5}{7.5}\selectfont$\pm$0.004} & {\fontsize{6.5}{7.5}\selectfont$\pm$0.003} & {\fontsize{6.5}{7.5}\selectfont$\pm$1.51} & {\fontsize{6.5}{7.5}\selectfont$\pm$0.002} & {\fontsize{6.5}{7.5}\selectfont$\pm$0.004} \\[1pt]
\hspace{0.5em}Pretrained, no fine-tuning & 19.15 & 0.250 & 0.734 & 46.78 & 0.393 & 0.589 \\
 & {\fontsize{6.5}{7.5}\selectfont$\pm$0.89} & {\fontsize{6.5}{7.5}\selectfont$\pm$0.003} & {\fontsize{6.5}{7.5}\selectfont$\pm$0.004} & {\fontsize{6.5}{7.5}\selectfont$\pm$0.83} & {\fontsize{6.5}{7.5}\selectfont$\pm$0.003} & {\fontsize{6.5}{7.5}\selectfont$\pm$0.002} \\[1pt]
\multicolumn{7}{@{}l}{\textit{Past-conditioned decoder}} \\
\hspace{0.5em}Reconstruction & 16.99 & 0.198 & 0.788 & 44.61 & 0.366 & 0.626 \\
 & {\fontsize{6.5}{7.5}\selectfont$\pm$0.64} & {\fontsize{6.5}{7.5}\selectfont$\pm$0.003} & {\fontsize{6.5}{7.5}\selectfont$\pm$0.003} & {\fontsize{6.5}{7.5}\selectfont$\pm$1.13} & {\fontsize{6.5}{7.5}\selectfont$\pm$0.006} & {\fontsize{6.5}{7.5}\selectfont$\pm$0.007} \\[1pt]
\hspace{0.5em}Inverse dynamics & {\boldmath\bfseries 13.44} & {\boldmath\bfseries 0.162} & {\boldmath\bfseries 0.815} & 39.39 & 0.333 & 0.653 \\
 & {\fontsize{6.5}{7.5}\selectfont$\pm$0.28} & {\fontsize{6.5}{7.5}\selectfont$\pm$0.001} & {\fontsize{6.5}{7.5}\selectfont$\pm$0.001} & {\fontsize{6.5}{7.5}\selectfont$\pm$2.01} & {\fontsize{6.5}{7.5}\selectfont$\pm$0.005} & {\fontsize{6.5}{7.5}\selectfont$\pm$0.008} \\[1pt]
\hspace{0.5em}Latent prediction & 15.64 & 0.177 & 0.805 & {\boldmath\bfseries 39.08} & {\boldmath\bfseries 0.331} & {\boldmath\bfseries 0.662} \\
 & {\fontsize{6.5}{7.5}\selectfont$\pm$0.26} & {\fontsize{6.5}{7.5}\selectfont$\pm$0.001} & {\fontsize{6.5}{7.5}\selectfont$\pm$0.001} & {\fontsize{6.5}{7.5}\selectfont$\pm$1.08} & {\fontsize{6.5}{7.5}\selectfont$\pm$0.007} & {\fontsize{6.5}{7.5}\selectfont$\pm$0.008} \\[1pt]
\midrule[\heavyrulewidth]
 & \multicolumn{2}{c}{Predictability} & \multicolumn{2}{c}{Distribution} \\
\cmidrule(lr){2-3}\cmidrule(lr){4-5}
 & ID latent MSE $\downarrow$ & OOD latent MSE $\downarrow$ & FID $\downarrow$ & FVD $\downarrow$ \\
\midrule
\multicolumn{7}{@{}l}{\textit{Ordinary decoder}} \\
\hspace{0.5em}Reconstruction (baseline) & 0.232 & 0.186 & 22.2 & 39.1 \\
 & {\fontsize{6.5}{7.5}\selectfont$\pm$0.008} & {\fontsize{6.5}{7.5}\selectfont$\pm$0.005} & {\fontsize{6.5}{7.5}\selectfont$\pm$0.6} & {\fontsize{6.5}{7.5}\selectfont$\pm$0.3} \\[1pt]
\hspace{0.5em}Pretrained, no fine-tuning & 0.253 & 0.204 & 28.6 & 42.7 \\
 & {\fontsize{6.5}{7.5}\selectfont$\pm$0.001} & {\fontsize{6.5}{7.5}\selectfont$\pm$0.001} & {\fontsize{6.5}{7.5}\selectfont$\pm$0.3} & {\fontsize{6.5}{7.5}\selectfont$\pm$0.8} \\[1pt]
\multicolumn{7}{@{}l}{\textit{Past-conditioned decoder}} \\
\hspace{0.5em}Reconstruction & 0.212 & 0.173 & 17.5 & 36.0 \\
 & {\fontsize{6.5}{7.5}\selectfont$\pm$0.008} & {\fontsize{6.5}{7.5}\selectfont$\pm$0.011} & {\fontsize{6.5}{7.5}\selectfont$\pm$0.9} & {\fontsize{6.5}{7.5}\selectfont$\pm$0.6} \\[1pt]
\hspace{0.5em}Inverse dynamics & 0.155 & 0.131 & {\boldmath\bfseries 11.1} & {\boldmath\bfseries 33.2} \\
 & {\fontsize{6.5}{7.5}\selectfont$\pm$0.001} & {\fontsize{6.5}{7.5}\selectfont$\pm$0.006} & {\fontsize{6.5}{7.5}\selectfont$\pm$0.4} & {\fontsize{6.5}{7.5}\selectfont$\pm$1.1} \\[1pt]
\hspace{0.5em}Latent prediction & {\boldmath\bfseries 0.046} & {\boldmath\bfseries 0.049} & 21.5 & 45.6 \\
 & {\fontsize{6.5}{7.5}\selectfont$\pm$0.001} & {\fontsize{6.5}{7.5}\selectfont$\pm$0.002} & {\fontsize{6.5}{7.5}\selectfont$\pm$0.4} & {\fontsize{6.5}{7.5}\selectfont$\pm$1.3} \\[1pt]
\bottomrule
\end{tabular}
\endgroup

%% file: figures/fig_worldgym/table.tex
\begingroup\fontsize{8}{9.5}\selectfont\setlength{\tabcolsep}{3pt}
\begin{tabular}{@{}l*{10}{c}@{}}
\toprule
 & \multicolumn{2}{c}{OpenVLA} & \multicolumn{2}{c}{Octo} & \multicolumn{2}{c}{RT-1-X} & \multicolumn{4}{c}{Three policies pooled} \\
\cmidrule(lr){2-3}\cmidrule(lr){4-5}\cmidrule(lr){6-7}\cmidrule(lr){8-11}
 & $r$ $\uparrow$ & $\rho$ $\uparrow$ & $r$ $\uparrow$ & $\rho$ $\uparrow$ & $r$ $\uparrow$ & $\rho$ $\uparrow$ & $r$ $\uparrow$ & $\rho$ $\uparrow$ & $\Delta r$ $\uparrow$ & $p$ \\
\midrule
\multicolumn{11}{@{}l}{10 sampling steps} \\
\multicolumn{11}{@{}l}{\hspace{0.5em}\textit{Ordinary decoder}} \\
\hspace{1em}Reconstruction (baseline) & 0.40 & 0.27 & 0.58 & 0.53 & 0.11 & 0.19 & 0.51 & 0.45 & -- & -- \\
 & {\fontsize{6.5}{7.5}\selectfont$\pm$0.20} & {\fontsize{6.5}{7.5}\selectfont$\pm$0.27} & {\fontsize{6.5}{7.5}\selectfont$\pm$0.21} & {\fontsize{6.5}{7.5}\selectfont$\pm$0.21} & {\fontsize{6.5}{7.5}\selectfont$\pm$0.24} & {\fontsize{6.5}{7.5}\selectfont$\pm$0.24} & {\fontsize{6.5}{7.5}\selectfont$\pm$0.10} & {\fontsize{6.5}{7.5}\selectfont$\pm$0.12} &  &  \\[1pt]
\hspace{1em}Pretrained, no fine-tuning & 0.56 & 0.56 & 0.58 & 0.55 & 0.01 & 0.10 & 0.57 & 0.55 & +0.06 & 0.261 \\
 & {\fontsize{6.5}{7.5}\selectfont$\pm$0.14} & {\fontsize{6.5}{7.5}\selectfont$\pm$0.17} & {\fontsize{6.5}{7.5}\selectfont$\pm$0.20} & {\fontsize{6.5}{7.5}\selectfont$\pm$0.20} & {\fontsize{6.5}{7.5}\selectfont$\pm$0.24} & {\fontsize{6.5}{7.5}\selectfont$\pm$0.26} & {\fontsize{6.5}{7.5}\selectfont$\pm$0.07} & {\fontsize{6.5}{7.5}\selectfont$\pm$0.10} & {\fontsize{6.5}{7.5}\selectfont$\pm$0.06} &  \\[1pt]
\multicolumn{11}{@{}l}{\hspace{0.5em}\textit{Past-conditioned decoder}} \\
\hspace{1em}Reconstruction & 0.53 & 0.58 & 0.50 & 0.46 & 0.24 & 0.49 & 0.49 & 0.55 & $-$0.02 & 0.623 \\
 & {\fontsize{6.5}{7.5}\selectfont$\pm$0.15} & {\fontsize{6.5}{7.5}\selectfont$\pm$0.19} & {\fontsize{6.5}{7.5}\selectfont$\pm$0.22} & {\fontsize{6.5}{7.5}\selectfont$\pm$0.24} & {\fontsize{6.5}{7.5}\selectfont$\pm$0.18} & {\fontsize{6.5}{7.5}\selectfont$\pm$0.18} & {\fontsize{6.5}{7.5}\selectfont$\pm$0.08} & {\fontsize{6.5}{7.5}\selectfont$\pm$0.09} & {\fontsize{6.5}{7.5}\selectfont$\pm$0.05} &  \\[1pt]
\hspace{1em}Inverse dynamics & 0.40 & 0.37 & 0.60 & 0.58 & 0.06 & 0.21 & 0.57 & 0.56 & +0.05 & 0.346 \\
 & {\fontsize{6.5}{7.5}\selectfont$\pm$0.18} & {\fontsize{6.5}{7.5}\selectfont$\pm$0.22} & {\fontsize{6.5}{7.5}\selectfont$\pm$0.19} & {\fontsize{6.5}{7.5}\selectfont$\pm$0.20} & {\fontsize{6.5}{7.5}\selectfont$\pm$0.21} & {\fontsize{6.5}{7.5}\selectfont$\pm$0.23} & {\fontsize{6.5}{7.5}\selectfont$\pm$0.07} & {\fontsize{6.5}{7.5}\selectfont$\pm$0.09} & {\fontsize{6.5}{7.5}\selectfont$\pm$0.06} &  \\[1pt]
\hspace{1em}Latent prediction & 0.54 & 0.53 & 0.68 & 0.66 & 0.08 & 0.15 & 0.62 & 0.64 & +0.11 & 0.076 \\
 & {\fontsize{6.5}{7.5}\selectfont$\pm$0.13} & {\fontsize{6.5}{7.5}\selectfont$\pm$0.18} & {\fontsize{6.5}{7.5}\selectfont$\pm$0.18} & {\fontsize{6.5}{7.5}\selectfont$\pm$0.18} & {\fontsize{6.5}{7.5}\selectfont$\pm$0.23} & {\fontsize{6.5}{7.5}\selectfont$\pm$0.23} & {\fontsize{6.5}{7.5}\selectfont$\pm$0.07} & {\fontsize{6.5}{7.5}\selectfont$\pm$0.09} & {\fontsize{6.5}{7.5}\selectfont$\pm$0.07} &  \\[1pt]
\midrule
\multicolumn{11}{@{}l}{2 sampling steps} \\
\multicolumn{11}{@{}l}{\hspace{0.5em}\textit{Ordinary decoder}} \\
\hspace{1em}Reconstruction (baseline) & 0.43 & 0.36 & 0.57 & 0.54 & 0.21 & 0.28 & 0.21 & 0.30 & -- & -- \\
 & {\fontsize{6.5}{7.5}\selectfont$\pm$0.17} & {\fontsize{6.5}{7.5}\selectfont$\pm$0.23} & {\fontsize{6.5}{7.5}\selectfont$\pm$0.21} & {\fontsize{6.5}{7.5}\selectfont$\pm$0.20} & {\fontsize{6.5}{7.5}\selectfont$\pm$0.24} & {\fontsize{6.5}{7.5}\selectfont$\pm$0.23} & {\fontsize{6.5}{7.5}\selectfont$\pm$0.07} & {\fontsize{6.5}{7.5}\selectfont$\pm$0.09} &  &  \\[1pt]
\hspace{1em}Pretrained, no fine-tuning & 0.47 & 0.43 & 0.57 & 0.58 & 0.13 & 0.36 & 0.21 & 0.30 & $-$0.00 & 0.948 \\
 & {\fontsize{6.5}{7.5}\selectfont$\pm$0.16} & {\fontsize{6.5}{7.5}\selectfont$\pm$0.23} & {\fontsize{6.5}{7.5}\selectfont$\pm$0.21} & {\fontsize{6.5}{7.5}\selectfont$\pm$0.21} & {\fontsize{6.5}{7.5}\selectfont$\pm$0.18} & {\fontsize{6.5}{7.5}\selectfont$\pm$0.21} & {\fontsize{6.5}{7.5}\selectfont$\pm$0.09} & {\fontsize{6.5}{7.5}\selectfont$\pm$0.10} & {\fontsize{6.5}{7.5}\selectfont$\pm$0.05} &  \\[1pt]
\multicolumn{11}{@{}l}{\hspace{0.5em}\textit{Past-conditioned decoder}} \\
\hspace{1em}Reconstruction & 0.40 & 0.42 & 0.54 & 0.54 & 0.30 & 0.45 & 0.15 & 0.28 & $-$0.06 & 0.007 \\
 & {\fontsize{6.5}{7.5}\selectfont$\pm$0.15} & {\fontsize{6.5}{7.5}\selectfont$\pm$0.21} & {\fontsize{6.5}{7.5}\selectfont$\pm$0.20} & {\fontsize{6.5}{7.5}\selectfont$\pm$0.20} & {\fontsize{6.5}{7.5}\selectfont$\pm$0.25} & {\fontsize{6.5}{7.5}\selectfont$\pm$0.20} & {\fontsize{6.5}{7.5}\selectfont$\pm$0.07} & {\fontsize{6.5}{7.5}\selectfont$\pm$0.09} & {\fontsize{6.5}{7.5}\selectfont$\pm$0.02} &  \\[1pt]
\hspace{1em}Inverse dynamics & 0.36 & 0.35 & 0.57 & 0.44 & 0.02 & 0.18 & 0.51 & 0.47 & +0.30 & $<$0.001 \\
 & {\fontsize{6.5}{7.5}\selectfont$\pm$0.20} & {\fontsize{6.5}{7.5}\selectfont$\pm$0.24} & {\fontsize{6.5}{7.5}\selectfont$\pm$0.21} & {\fontsize{6.5}{7.5}\selectfont$\pm$0.23} & {\fontsize{6.5}{7.5}\selectfont$\pm$0.22} & {\fontsize{6.5}{7.5}\selectfont$\pm$0.24} & {\fontsize{6.5}{7.5}\selectfont$\pm$0.09} & {\fontsize{6.5}{7.5}\selectfont$\pm$0.11} & {\fontsize{6.5}{7.5}\selectfont$\pm$0.06} &  \\[1pt]
\hspace{1em}Latent prediction & 0.55 & 0.49 & 0.57 & 0.54 & 0.56 & 0.63 & 0.67 & 0.69 & +0.45 & $<$0.001 \\
 & {\fontsize{6.5}{7.5}\selectfont$\pm$0.14} & {\fontsize{6.5}{7.5}\selectfont$\pm$0.20} & {\fontsize{6.5}{7.5}\selectfont$\pm$0.21} & {\fontsize{6.5}{7.5}\selectfont$\pm$0.21} & {\fontsize{6.5}{7.5}\selectfont$\pm$0.18} & {\fontsize{6.5}{7.5}\selectfont$\pm$0.17} & {\fontsize{6.5}{7.5}\selectfont$\pm$0.07} & {\fontsize{6.5}{7.5}\selectfont$\pm$0.09} & {\fontsize{6.5}{7.5}\selectfont$\pm$0.07} &  \\[1pt]
\bottomrule
\end{tabular}
\endgroup

%% file: sections/appendix/qualitative.tex
\subsection{Qualitative examples}
\label{app:qualitative}
\input{sections/appendix/teaser}

\paragraph{Rollouts.}
Figures~\ref{fig:qual-ood-2-sel} to~\ref{fig:qual-id-10-rand} show rollouts of the five world models of Table~\ref{tab:matched-compute} (one training seed, 100,000 updates) in two evaluation conditions of Appendix~\ref{app:evaluation}, each with 2 and with 10 sampling steps.
Out of distribution, the clips are the 479 AutoEval trajectories, and each world model replays the recorded actions for 29 frames from the context frame, one frame at a time.
In distribution, the clips are the 224 held-out Bridge videos with at least 60 frames of the long-rollout evaluation, and each world model rolls out 59 frames, eight at a time.
All world models roll out the same clip from the same context frame, with the same actions and the same diffusion noise.
The noise differs from that of the quantitative evaluation, so the LPIPS shown beside each row is that of the rollout shown.
For each condition, one figure shows two selected clips and one shows two clips drawn at random.

\paragraph{All clips.}
Table~\ref{tab:qual-stats} places the examples in the distribution over all clips.
Both temporal world models have a lower LPIPS than the baseline on 92.0\% to 100\% of the clips in every condition, with median reductions of 13.1\% to 31.8\%.
The gap is largest with 2 sampling steps: out of distribution, latent prediction has a lower LPIPS than the baseline on all 479 trajectories, and both temporal world models are below all three reconstruction-only world models on 96.2\% of them.
With 10 steps, the reconstruction-only world models degrade more slowly than with 2 steps, and the differences appear mainly late in the long rollouts.

\begin{table}[h]
    \caption{\textbf{Rollout quality over all clips.} Lower than baseline: percentage of clips on which the temporal world model has a lower clip-mean LPIPS than reconstruction with the ordinary decoder. Lower than all reconstruction-only: the same against all three reconstruction-only world models; Both: the two temporal world models at once. Median change: relative change of LPIPS against the baseline. ID: the 224 held-out videos with at least 60 frames, rolled out for 59 frames; OOD: the 479 AutoEval trajectories, rolled out for 29 frames; one rollout per clip. On almost every clip, the temporal world models are closer to the recording than the baseline, most consistently with 2 sampling steps.}
    \label{tab:qual-stats}
    \centering
    \smallskip
    \begingroup\fontsize{8}{9.5}\selectfont\setlength{\tabcolsep}{3pt}
    \input{figures/fig_qualitative_rollouts/table.tex}
    \endgroup
\end{table}

\newcommand{\qualpanel}[2]{\includegraphics[width=\textwidth]{figures/fig_qualitative_rollouts/#1.pdf}\\[1pt]{\small #2}}

\begin{figure}[p]
    \centering
    \qualpanel{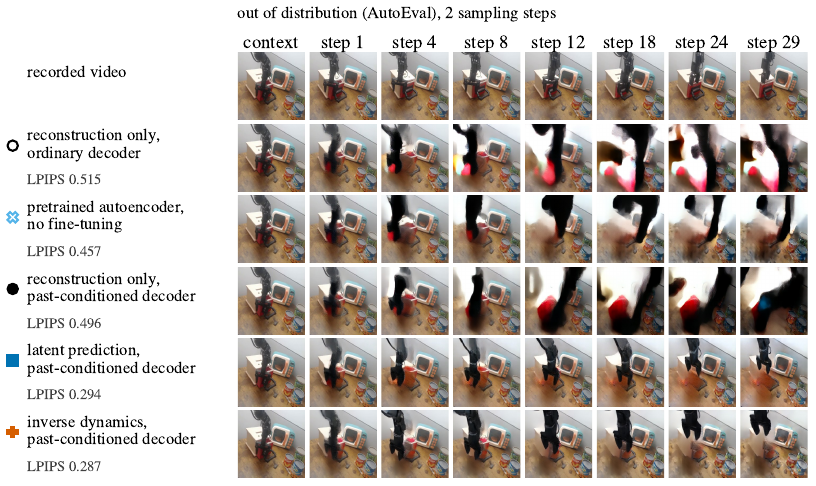}{(a) ``Close the drawer''.}\\[8pt]
    \qualpanel{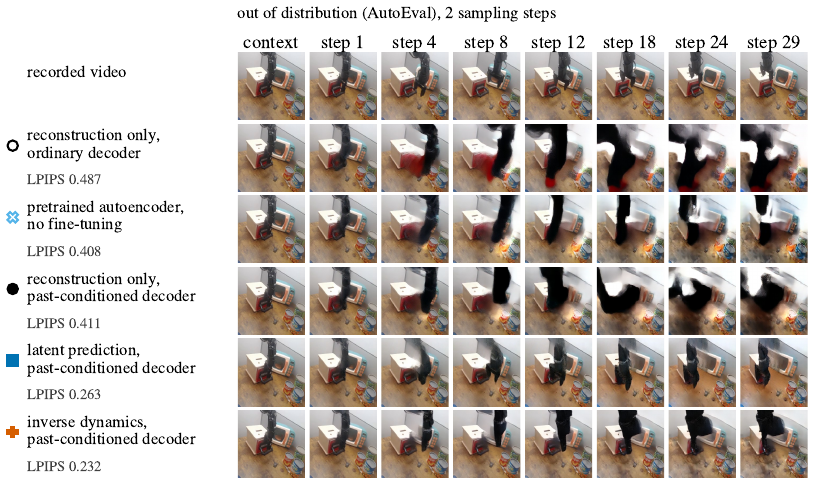}{(b) ``Close the drawer''.}
    \caption{\textbf{Out of distribution, 2 sampling steps, selected trajectories.} AutoEval trajectories with the recorded actions replayed for 29 frames. Top row: the recorded frames; below: the rollout of each world model from the same context frame, with its LPIPS.}
    \label{fig:qual-ood-2-sel}
\end{figure}

\begin{figure}[p]
    \centering
    \qualpanel{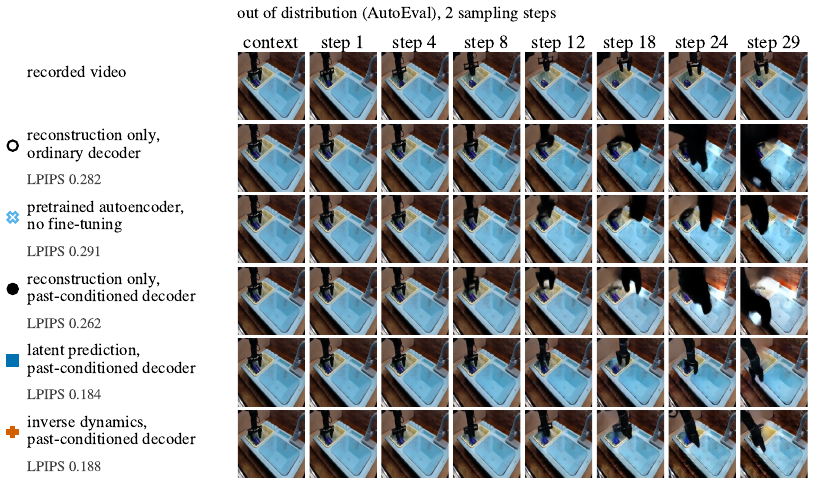}{(a) ``Put the eggplant in the blue sink''.}\\[8pt]
    \qualpanel{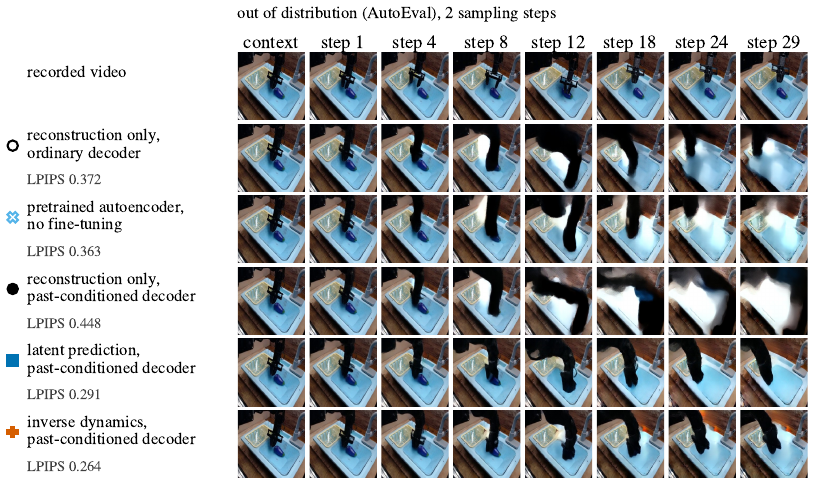}{(b) ``Put the eggplant in the yellow basket''.}
    \caption{\textbf{Out of distribution, 2 sampling steps, random trajectories.} AutoEval trajectories with the recorded actions replayed for 29 frames; rows as in Figure~\ref{fig:qual-ood-2-sel}.}
    \label{fig:qual-ood-2-rand}
\end{figure}

\begin{figure}[p]
    \centering
    \qualpanel{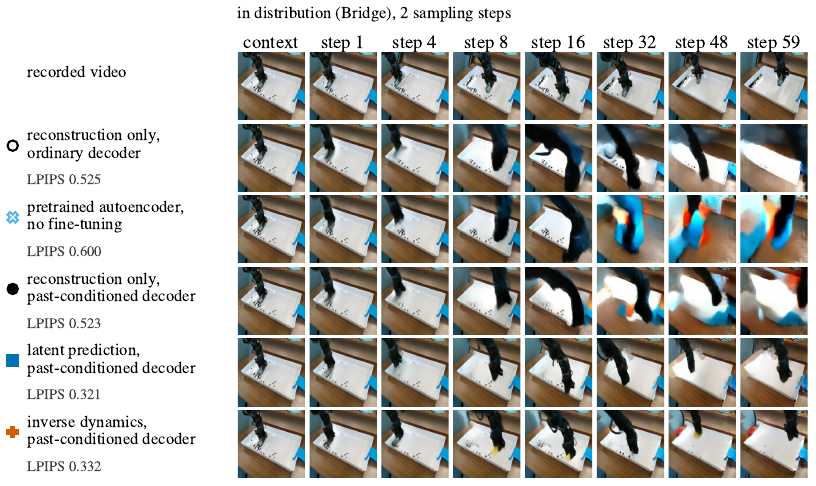}{(a)}\\[8pt]
    \qualpanel{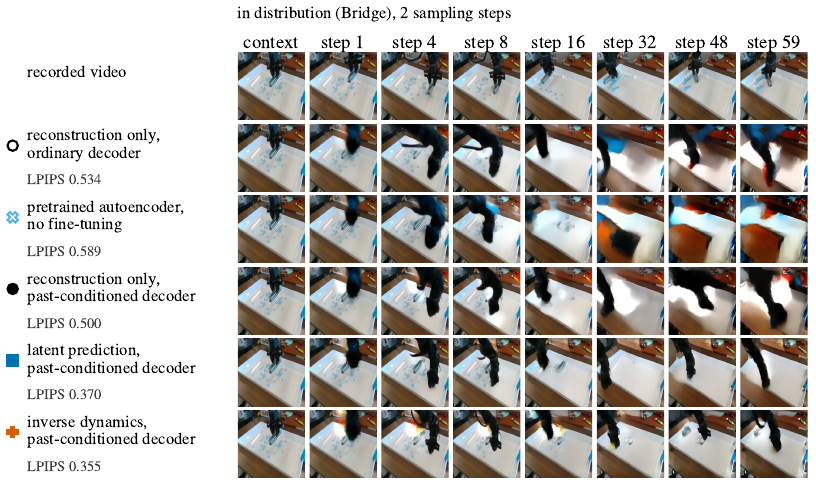}{(b)}
    \caption{\textbf{In distribution, 2 sampling steps, selected videos.} Held-out Bridge videos rolled out for 59 frames; rows as in Figure~\ref{fig:qual-ood-2-sel}.}
    \label{fig:qual-id-2-sel}
\end{figure}

\begin{figure}[p]
    \centering
    \qualpanel{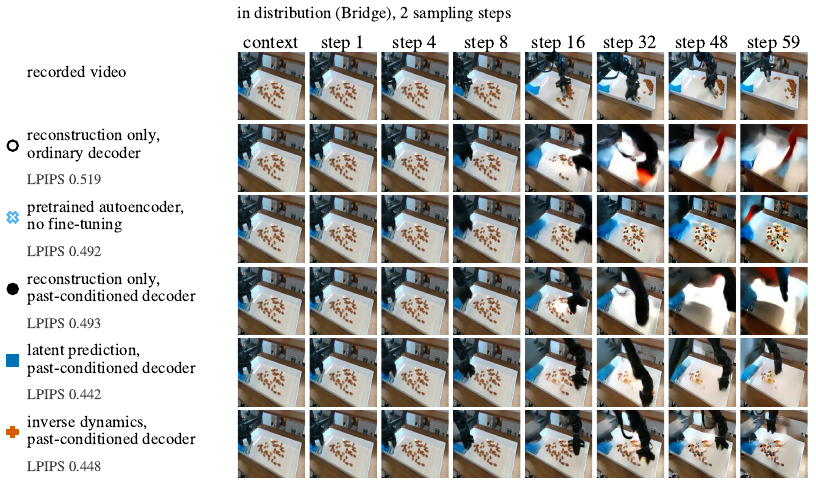}{(a)}\\[8pt]
    \qualpanel{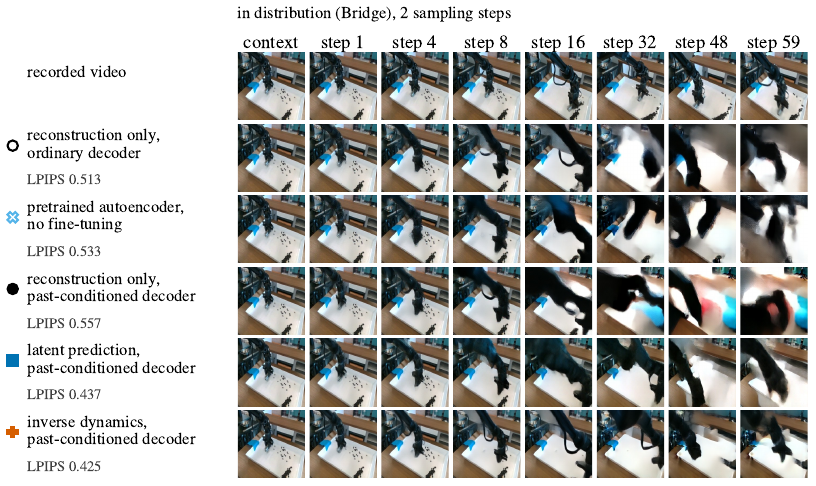}{(b)}
    \caption{\textbf{In distribution, 2 sampling steps, random videos.} Held-out Bridge videos rolled out for 59 frames; rows as in Figure~\ref{fig:qual-ood-2-sel}.}
    \label{fig:qual-id-2-rand}
\end{figure}

\begin{figure}[p]
    \centering
    \qualpanel{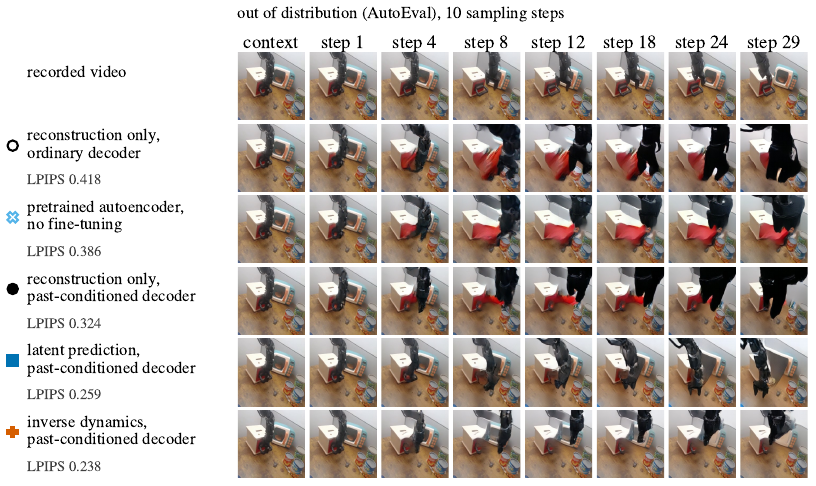}{(a) ``Close the drawer''.}\\[8pt]
    \qualpanel{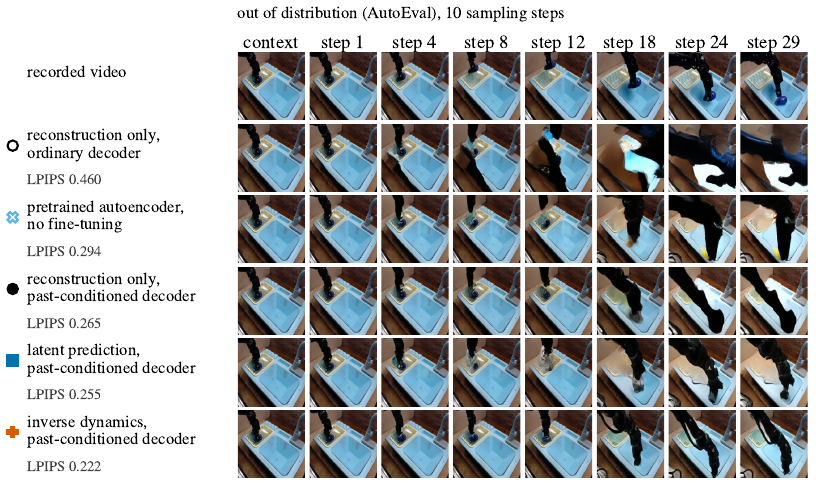}{(b) ``Put the eggplant in the blue sink''.}
    \caption{\textbf{Out of distribution, 10 sampling steps, selected trajectories.} AutoEval trajectories with the recorded actions replayed for 29 frames; rows as in Figure~\ref{fig:qual-ood-2-sel}.}
    \label{fig:qual-ood-10-sel}
\end{figure}

\begin{figure}[p]
    \centering
    \qualpanel{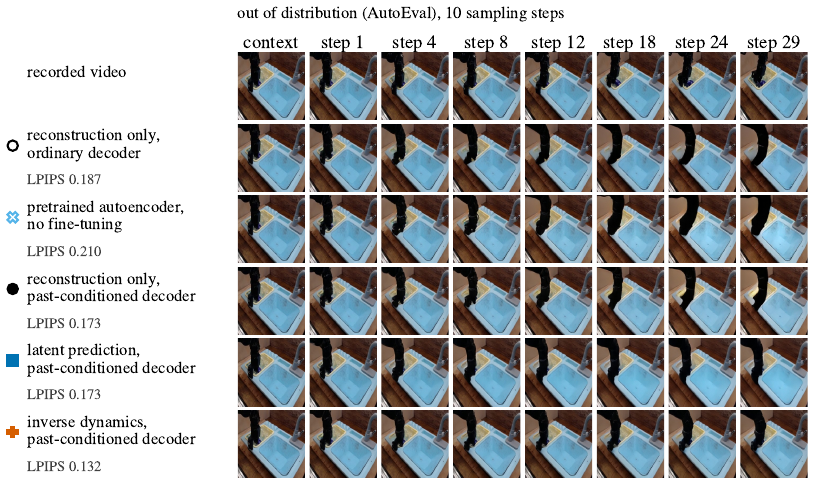}{(a) ``Put the eggplant in the blue sink''.}\\[8pt]
    \qualpanel{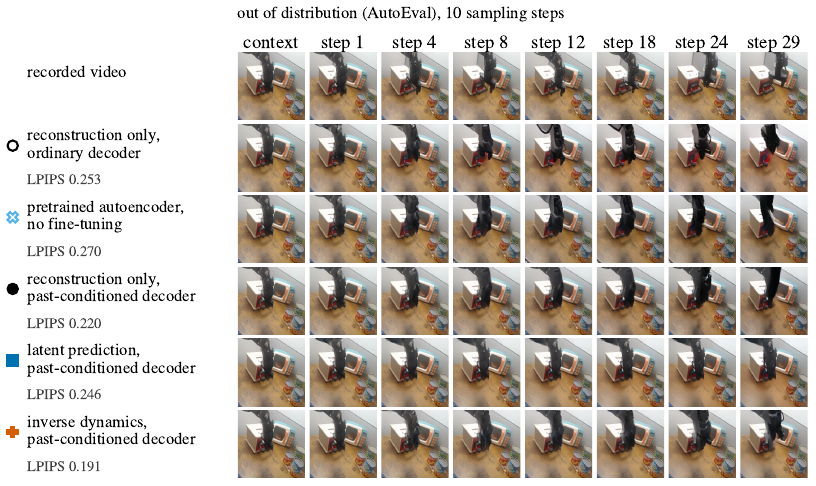}{(b) ``Open the drawer''.}
    \caption{\textbf{Out of distribution, 10 sampling steps, random trajectories.} AutoEval trajectories with the recorded actions replayed for 29 frames; rows as in Figure~\ref{fig:qual-ood-2-sel}.}
    \label{fig:qual-ood-10-rand}
\end{figure}

\begin{figure}[p]
    \centering
    \qualpanel{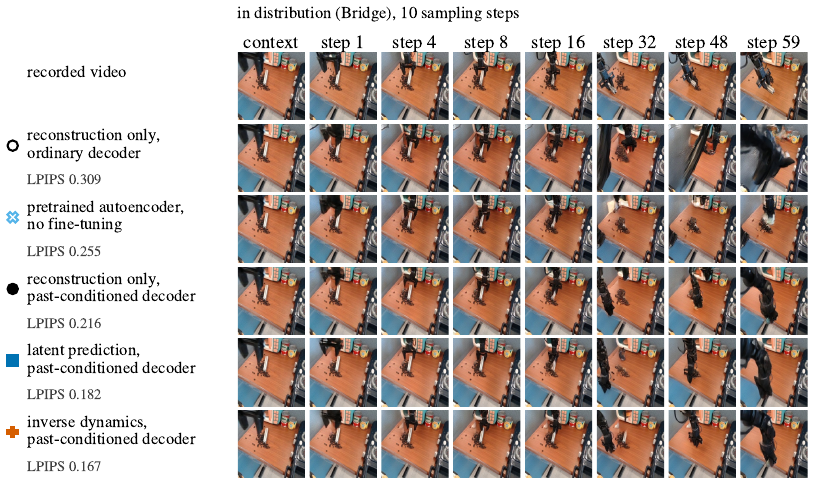}{(a)}\\[8pt]
    \qualpanel{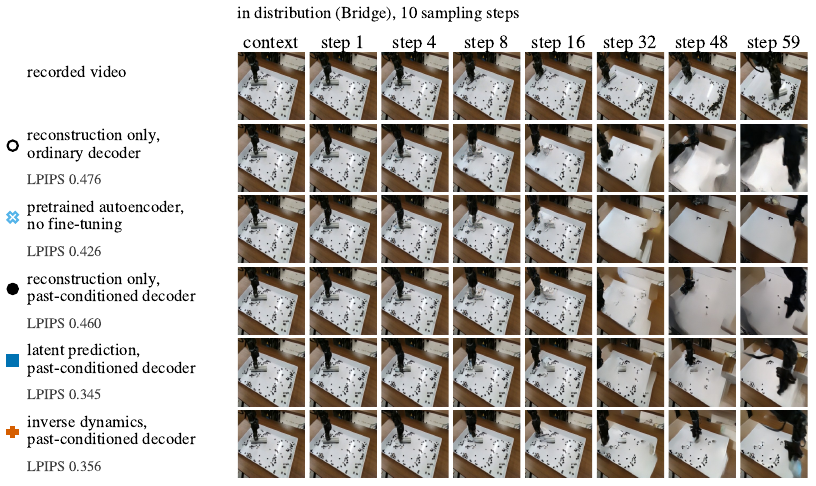}{(b)}
    \caption{\textbf{In distribution, 10 sampling steps, selected videos.} Held-out Bridge videos rolled out for 59 frames; rows as in Figure~\ref{fig:qual-ood-2-sel}.}
    \label{fig:qual-id-10-sel}
\end{figure}

\begin{figure}[p]
    \centering
    \qualpanel{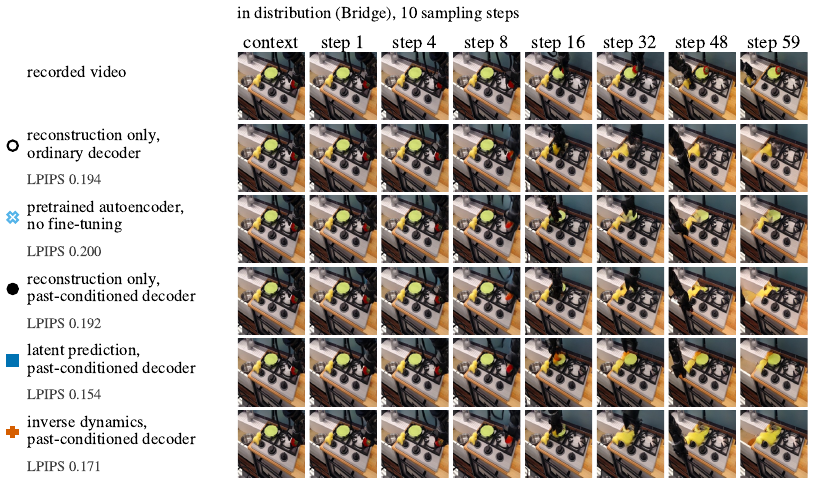}{(a)}\\[8pt]
    \qualpanel{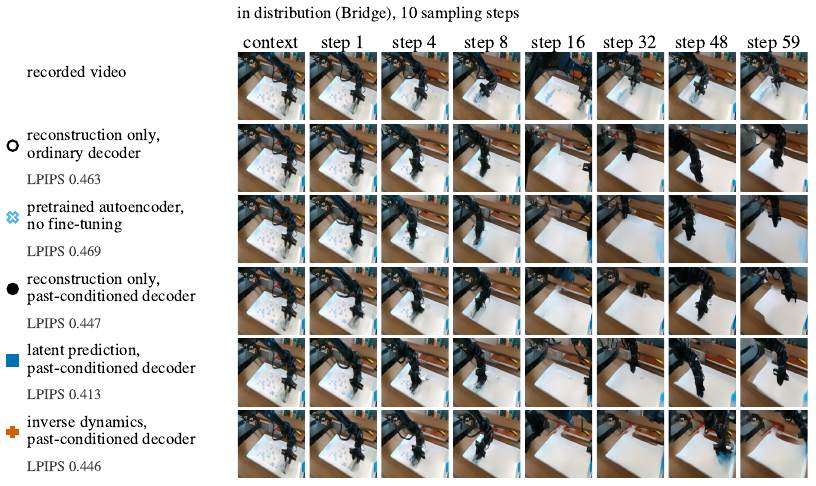}{(b)}
    \caption{\textbf{In distribution, 10 sampling steps, random videos.} Held-out Bridge videos rolled out for 59 frames; rows as in Figure~\ref{fig:qual-ood-2-sel}.}
    \label{fig:qual-id-10-rand}
\end{figure}

\begin{figure}[p]
    \centering
    \includegraphics[width=0.8\textwidth]{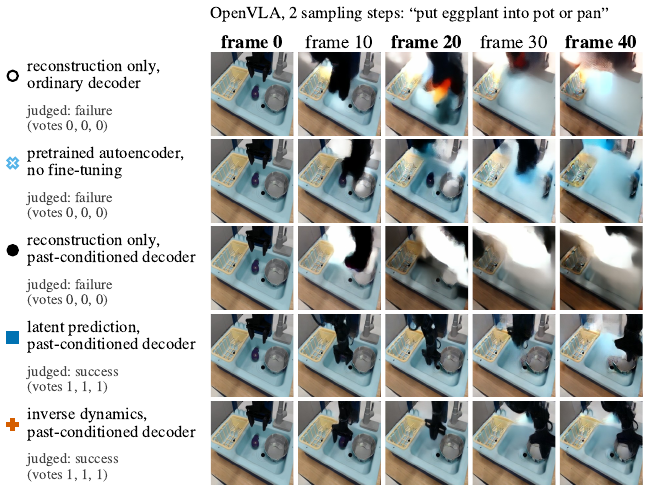}\\[1pt]
    {\small (a) Selected.}\\[8pt]
    \includegraphics[width=0.8\textwidth]{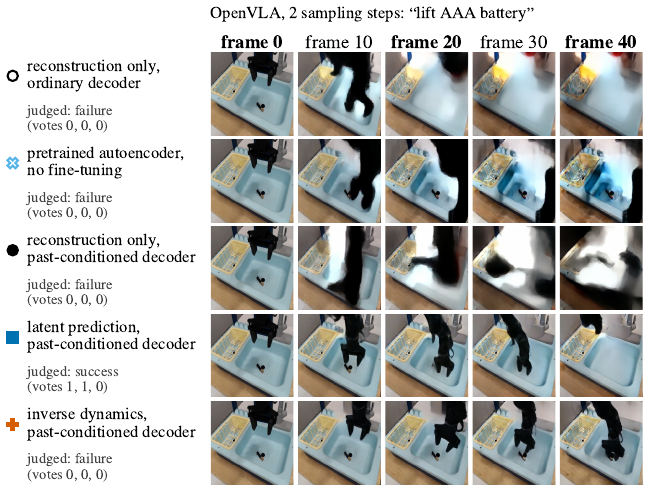}\\[1pt]
    {\small (b) Random starting frame.}
    \caption{\textbf{OpenVLA in WorldGym with 2 sampling steps.} One starting frame per panel, inside the five world models of Table~\ref{tab:matched-compute} (one training seed), at frames 0 to 40; bold: the frames the judge sees. Beside each row: the judged outcome and the three votes.}
    \label{fig:qual-worldgym}
\end{figure}

\paragraph{Policy evaluation.}
Figure~\ref{fig:qual-worldgym} shows OpenVLA acting inside the world models in WorldGym (Appendix~\ref{app:worldgym}) with 2 sampling steps.
Because the loop is closed (Appendix~\ref{app:worldgym}), the rows share the starting frame, the instruction and the policy, but not the actions.

%% file: sections/appendix/teaser.tex
\begin{figure}[t]
    \centering
    \includegraphics[width=\textwidth]{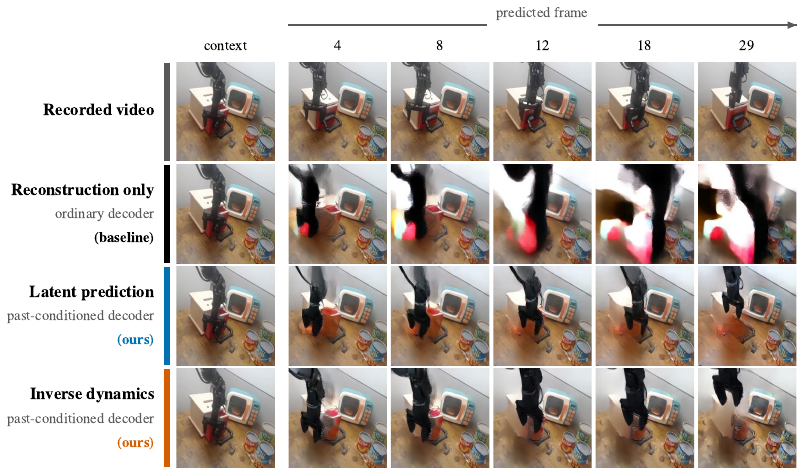}
    \caption{\textbf{AutoEval rollouts with video world models trained on different latent spaces.} The baseline learns its latent space through reconstruction, as is standard; our recipe adds a temporal objective (latent prediction or inverse dynamics) and a past-conditioned decoder. One AutoEval trajectory, out of distribution: the recorded actions are replayed from one context frame for three times the length of the training clips, with 2 sampling steps. Figures~\ref{fig:qual-ood-2-sel} to~\ref{fig:qual-id-10-rand} show more examples, selected and random.}
    \label{fig:teaser}
\end{figure}

%% file: figures/fig_qualitative_rollouts/table.tex
\begin{tabular}{llrccccccc}
\toprule
 & & & \multicolumn{2}{c}{Lower than baseline (\%)} & \multicolumn{3}{c}{Lower than all reconstruction-only (\%)} & \multicolumn{2}{c}{Median change (\%)} \\
\cmidrule(lr){4-5}\cmidrule(lr){6-8}\cmidrule(lr){9-10}
Setting & Steps & Clips & Latent & Inverse & Latent & Inverse & Both & Latent & Inverse \\
\midrule
ID & 10 & 224 & 92.0 & 93.3 & 80.4 & 79.9 & 69.6 & $-13.1$ & $-13.6$ \\
ID & 2 & 224 & 96.0 & 98.7 & 89.3 & 92.9 & 87.5 & $-21.5$ & $-21.1$ \\
OOD & 10 & 479 & 95.0 & 96.7 & 60.5 & 86.2 & 53.2 & $-13.9$ & $-22.8$ \\
OOD & 2 & 479 & 100.0 & 99.2 & 97.9 & 97.9 & 96.2 & $-28.9$ & $-31.8$ \\
\bottomrule
\end{tabular}

%% file: figures/tab_longer/table_full.tex
\begingroup\fontsize{8}{9.5}\selectfont\setlength{\tabcolsep}{2.6pt}
\begin{tabular}{@{}l*{6}{c}@{}}
\toprule
 & \multicolumn{3}{c}{ID} & \multicolumn{3}{c}{OOD} \\
\cmidrule(lr){2-4}\cmidrule(lr){5-7}
 & MSE $\downarrow$ & LPIPS $\downarrow$ & SSIM $\uparrow$ & MSE $\downarrow$ & LPIPS $\downarrow$ & SSIM $\uparrow$ \\
\midrule
\multicolumn{7}{@{}l}{\textit{200,000 world-model updates}} \\
\hspace{0.5em}Reconstruction & 12.14 & 0.163 & 0.813 & 34.53 & 0.280 & 0.707 \\
 & {\fontsize{6.5}{7.5}\selectfont$\pm$0.36} & {\fontsize{6.5}{7.5}\selectfont$\pm$0.001} & {\fontsize{6.5}{7.5}\selectfont$\pm$0.002} & {\fontsize{6.5}{7.5}\selectfont$\pm$2.33} & {\fontsize{6.5}{7.5}\selectfont$\pm$0.006} & {\fontsize{6.5}{7.5}\selectfont$\pm$0.006} \\[1pt]
\hspace{0.5em}Inverse dynamics & {\boldmath\bfseries 11.56} & {\boldmath\bfseries 0.148} & {\boldmath\bfseries 0.823} & 31.19 & {\boldmath\bfseries 0.242} & 0.737 \\
 & {\fontsize{6.5}{7.5}\selectfont$\pm$0.02} & {\fontsize{6.5}{7.5}\selectfont$\pm$0.001} & {\fontsize{6.5}{7.5}\selectfont$\pm$0.001} & {\fontsize{6.5}{7.5}\selectfont$\pm$1.42} & {\fontsize{6.5}{7.5}\selectfont$\pm$0.005} & {\fontsize{6.5}{7.5}\selectfont$\pm$0.005} \\[1pt]
\hspace{0.5em}Latent prediction & 12.85 & 0.157 & 0.818 & {\boldmath\bfseries 26.00} & 0.255 & {\boldmath\bfseries 0.743} \\
 & {\fontsize{6.5}{7.5}\selectfont$\pm$0.21} & {\fontsize{6.5}{7.5}\selectfont$\pm$0.001} & {\fontsize{6.5}{7.5}\selectfont$\pm$0.001} & {\fontsize{6.5}{7.5}\selectfont$\pm$0.10} & {\fontsize{6.5}{7.5}\selectfont$\pm$0.004} & {\fontsize{6.5}{7.5}\selectfont$\pm$0.003} \\[1pt]
\midrule
\multicolumn{7}{@{}l}{\textit{100,000 world-model updates, same training seeds}} \\
\hspace{0.5em}Reconstruction & 14.00 & 0.179 & 0.800 & 32.00 & 0.276 & 0.712 \\
 & {\fontsize{6.5}{7.5}\selectfont$\pm$0.24} & {\fontsize{6.5}{7.5}\selectfont$\pm$0.000} & {\fontsize{6.5}{7.5}\selectfont$\pm$0.001} & {\fontsize{6.5}{7.5}\selectfont$\pm$0.19} & {\fontsize{6.5}{7.5}\selectfont$\pm$0.002} & {\fontsize{6.5}{7.5}\selectfont$\pm$0.001} \\[1pt]
\hspace{0.5em}Inverse dynamics & {\boldmath\bfseries 13.31} & {\boldmath\bfseries 0.161} & {\boldmath\bfseries 0.812} & 28.71 & {\boldmath\bfseries 0.239} & {\boldmath\bfseries 0.743} \\
 & {\fontsize{6.5}{7.5}\selectfont$\pm$0.03} & {\fontsize{6.5}{7.5}\selectfont$\pm$0.001} & {\fontsize{6.5}{7.5}\selectfont$\pm$0.001} & {\fontsize{6.5}{7.5}\selectfont$\pm$1.48} & {\fontsize{6.5}{7.5}\selectfont$\pm$0.008} & {\fontsize{6.5}{7.5}\selectfont$\pm$0.009} \\[1pt]
\hspace{0.5em}Latent prediction & 14.10 & 0.169 & 0.810 & {\boldmath\bfseries 26.89} & 0.257 & 0.741 \\
 & {\fontsize{6.5}{7.5}\selectfont$\pm$0.06} & {\fontsize{6.5}{7.5}\selectfont$\pm$0.001} & {\fontsize{6.5}{7.5}\selectfont$\pm$0.002} & {\fontsize{6.5}{7.5}\selectfont$\pm$0.32} & {\fontsize{6.5}{7.5}\selectfont$\pm$0.003} & {\fontsize{6.5}{7.5}\selectfont$\pm$0.003} \\[1pt]
\midrule[\heavyrulewidth]
 & \multicolumn{3}{c}{2-step} & \multicolumn{3}{c}{Long-horizon} \\
\cmidrule(lr){2-4}\cmidrule(lr){5-7}
 & MSE $\downarrow$ & LPIPS $\downarrow$ & SSIM $\uparrow$ & MSE $\downarrow$ & LPIPS $\downarrow$ & SSIM $\uparrow$ \\
\midrule
\multicolumn{7}{@{}l}{\textit{200,000 world-model updates}} \\
\hspace{0.5em}Reconstruction & 14.89 & 0.187 & 0.795 & 37.27 & 0.321 & 0.670 \\
 & {\fontsize{6.5}{7.5}\selectfont$\pm$0.64} & {\fontsize{6.5}{7.5}\selectfont$\pm$0.001} & {\fontsize{6.5}{7.5}\selectfont$\pm$0.001} & {\fontsize{6.5}{7.5}\selectfont$\pm$0.64} & {\fontsize{6.5}{7.5}\selectfont$\pm$0.000} & {\fontsize{6.5}{7.5}\selectfont$\pm$0.001} \\[1pt]
\hspace{0.5em}Inverse dynamics & {\boldmath\bfseries 12.23} & {\boldmath\bfseries 0.152} & {\boldmath\bfseries 0.822} & {\boldmath\bfseries 33.55} & {\boldmath\bfseries 0.292} & {\boldmath\bfseries 0.694} \\
 & {\fontsize{6.5}{7.5}\selectfont$\pm$0.12} & {\fontsize{6.5}{7.5}\selectfont$\pm$0.000} & {\fontsize{6.5}{7.5}\selectfont$\pm$0.001} & {\fontsize{6.5}{7.5}\selectfont$\pm$0.09} & {\fontsize{6.5}{7.5}\selectfont$\pm$0.003} & {\fontsize{6.5}{7.5}\selectfont$\pm$0.004} \\[1pt]
\hspace{0.5em}Latent prediction & 13.98 & 0.164 & 0.814 & 35.36 & 0.298 & 0.693 \\
 & {\fontsize{6.5}{7.5}\selectfont$\pm$0.12} & {\fontsize{6.5}{7.5}\selectfont$\pm$0.001} & {\fontsize{6.5}{7.5}\selectfont$\pm$0.000} & {\fontsize{6.5}{7.5}\selectfont$\pm$3.06} & {\fontsize{6.5}{7.5}\selectfont$\pm$0.013} & {\fontsize{6.5}{7.5}\selectfont$\pm$0.012} \\[1pt]
\midrule
\multicolumn{7}{@{}l}{\textit{100,000 world-model updates, same training seeds}} \\
\hspace{0.5em}Reconstruction & 16.81 & 0.200 & 0.786 & 45.71 & 0.372 & 0.619 \\
 & {\fontsize{6.5}{7.5}\selectfont$\pm$1.06} & {\fontsize{6.5}{7.5}\selectfont$\pm$0.003} & {\fontsize{6.5}{7.5}\selectfont$\pm$0.004} & {\fontsize{6.5}{7.5}\selectfont$\pm$0.44} & {\fontsize{6.5}{7.5}\selectfont$\pm$0.001} & {\fontsize{6.5}{7.5}\selectfont$\pm$0.002} \\[1pt]
\hspace{0.5em}Inverse dynamics & {\boldmath\bfseries 13.72} & {\boldmath\bfseries 0.163} & {\boldmath\bfseries 0.814} & 40.41 & 0.337 & 0.648 \\
 & {\fontsize{6.5}{7.5}\selectfont$\pm$0.03} & {\fontsize{6.5}{7.5}\selectfont$\pm$0.001} & {\fontsize{6.5}{7.5}\selectfont$\pm$0.000} & {\fontsize{6.5}{7.5}\selectfont$\pm$3.00} & {\fontsize{6.5}{7.5}\selectfont$\pm$0.006} & {\fontsize{6.5}{7.5}\selectfont$\pm$0.011} \\[1pt]
\hspace{0.5em}Latent prediction & 15.38 & 0.177 & 0.805 & {\boldmath\bfseries 39.99} & {\boldmath\bfseries 0.336} & {\boldmath\bfseries 0.657} \\
 & {\fontsize{6.5}{7.5}\selectfont$\pm$0.00} & {\fontsize{6.5}{7.5}\selectfont$\pm$0.001} & {\fontsize{6.5}{7.5}\selectfont$\pm$0.002} & {\fontsize{6.5}{7.5}\selectfont$\pm$0.98} & {\fontsize{6.5}{7.5}\selectfont$\pm$0.009} & {\fontsize{6.5}{7.5}\selectfont$\pm$0.010} \\[1pt]
\midrule[\heavyrulewidth]
 & \multicolumn{2}{c}{Distribution} \\
\cmidrule(lr){2-3}
 & FID $\downarrow$ & FVD $\downarrow$ \\
\midrule
\multicolumn{7}{@{}l}{\textit{200,000 world-model updates}} \\
\hspace{0.5em}Reconstruction & 12.1 & 28.4 \\
 & {\fontsize{6.5}{7.5}\selectfont$\pm$0.0} & {\fontsize{6.5}{7.5}\selectfont$\pm$0.7} \\[1pt]
\hspace{0.5em}Inverse dynamics & {\boldmath\bfseries 8.9} & {\boldmath\bfseries 28.1} \\
 & {\fontsize{6.5}{7.5}\selectfont$\pm$0.3} & {\fontsize{6.5}{7.5}\selectfont$\pm$0.2} \\[1pt]
\hspace{0.5em}Latent prediction & 14.2 & 37.7 \\
 & {\fontsize{6.5}{7.5}\selectfont$\pm$0.7} & {\fontsize{6.5}{7.5}\selectfont$\pm$1.4} \\[1pt]
\midrule
\multicolumn{7}{@{}l}{\textit{100,000 world-model updates, same training seeds}} \\
\hspace{0.5em}Reconstruction & 18.3 & 36.6 \\
 & {\fontsize{6.5}{7.5}\selectfont$\pm$0.2} & {\fontsize{6.5}{7.5}\selectfont$\pm$0.1} \\[1pt]
\hspace{0.5em}Inverse dynamics & {\boldmath\bfseries 11.4} & {\boldmath\bfseries 33.5} \\
 & {\fontsize{6.5}{7.5}\selectfont$\pm$0.3} & {\fontsize{6.5}{7.5}\selectfont$\pm$1.8} \\[1pt]
\hspace{0.5em}Latent prediction & 21.3 & 46.0 \\
 & {\fontsize{6.5}{7.5}\selectfont$\pm$0.6} & {\fontsize{6.5}{7.5}\selectfont$\pm$2.2} \\[1pt]
\bottomrule
\end{tabular}
\endgroup

%% file: figures/fig_worldgym/table_200k.tex
\begingroup\fontsize{8}{9.5}\selectfont\setlength{\tabcolsep}{3pt}
\begin{tabular}{@{}l*{10}{c}@{}}
\toprule
 & \multicolumn{2}{c}{OpenVLA} & \multicolumn{2}{c}{Octo} & \multicolumn{2}{c}{RT-1-X} & \multicolumn{4}{c}{Three policies pooled} \\
\cmidrule(lr){2-3}\cmidrule(lr){4-5}\cmidrule(lr){6-7}\cmidrule(lr){8-11}
 & $r$ $\uparrow$ & $\rho$ $\uparrow$ & $r$ $\uparrow$ & $\rho$ $\uparrow$ & $r$ $\uparrow$ & $\rho$ $\uparrow$ & $r$ $\uparrow$ & $\rho$ $\uparrow$ & $\Delta r$ $\uparrow$ & $p$ \\
\midrule
\multicolumn{11}{@{}l}{10 sampling steps} \\
\hspace{0.5em}Reconstruction & 0.50 & 0.45 & 0.55 & 0.53 & 0.24 & 0.41 & 0.61 & 0.61 & -- & -- \\
 & {\fontsize{6.5}{7.5}\selectfont$\pm$0.16} & {\fontsize{6.5}{7.5}\selectfont$\pm$0.22} & {\fontsize{6.5}{7.5}\selectfont$\pm$0.21} & {\fontsize{6.5}{7.5}\selectfont$\pm$0.21} & {\fontsize{6.5}{7.5}\selectfont$\pm$0.23} & {\fontsize{6.5}{7.5}\selectfont$\pm$0.21} & {\fontsize{6.5}{7.5}\selectfont$\pm$0.08} & {\fontsize{6.5}{7.5}\selectfont$\pm$0.10} &  &  \\[1pt]
\hspace{0.5em}Inverse dynamics & 0.54 & 0.47 & 0.57 & 0.53 & 0.34 & 0.30 & 0.66 & 0.65 & +0.05 & 0.081 \\
 & {\fontsize{6.5}{7.5}\selectfont$\pm$0.15} & {\fontsize{6.5}{7.5}\selectfont$\pm$0.20} & {\fontsize{6.5}{7.5}\selectfont$\pm$0.21} & {\fontsize{6.5}{7.5}\selectfont$\pm$0.21} & {\fontsize{6.5}{7.5}\selectfont$\pm$0.19} & {\fontsize{6.5}{7.5}\selectfont$\pm$0.19} & {\fontsize{6.5}{7.5}\selectfont$\pm$0.07} & {\fontsize{6.5}{7.5}\selectfont$\pm$0.10} & {\fontsize{6.5}{7.5}\selectfont$\pm$0.03} &  \\[1pt]
\hspace{0.5em}Latent prediction & 0.61 & 0.58 & 0.53 & 0.52 & 0.56 & 0.58 & 0.69 & 0.68 & +0.07 & 0.086 \\
 & {\fontsize{6.5}{7.5}\selectfont$\pm$0.14} & {\fontsize{6.5}{7.5}\selectfont$\pm$0.20} & {\fontsize{6.5}{7.5}\selectfont$\pm$0.21} & {\fontsize{6.5}{7.5}\selectfont$\pm$0.21} & {\fontsize{6.5}{7.5}\selectfont$\pm$0.15} & {\fontsize{6.5}{7.5}\selectfont$\pm$0.19} & {\fontsize{6.5}{7.5}\selectfont$\pm$0.08} & {\fontsize{6.5}{7.5}\selectfont$\pm$0.10} & {\fontsize{6.5}{7.5}\selectfont$\pm$0.04} &  \\[1pt]
\midrule
\multicolumn{11}{@{}l}{2 sampling steps} \\
\hspace{0.5em}Reconstruction & 0.44 & 0.40 & 0.53 & 0.53 & 0.21 & 0.51 & 0.20 & 0.31 & -- & -- \\
 & {\fontsize{6.5}{7.5}\selectfont$\pm$0.16} & {\fontsize{6.5}{7.5}\selectfont$\pm$0.23} & {\fontsize{6.5}{7.5}\selectfont$\pm$0.21} & {\fontsize{6.5}{7.5}\selectfont$\pm$0.21} & {\fontsize{6.5}{7.5}\selectfont$\pm$0.18} & {\fontsize{6.5}{7.5}\selectfont$\pm$0.18} & {\fontsize{6.5}{7.5}\selectfont$\pm$0.08} & {\fontsize{6.5}{7.5}\selectfont$\pm$0.10} &  &  \\[1pt]
\hspace{0.5em}Inverse dynamics & 0.50 & 0.46 & 0.58 & 0.55 & 0.55 & 0.70 & 0.60 & 0.61 & +0.40 & $<$0.001 \\
 & {\fontsize{6.5}{7.5}\selectfont$\pm$0.14} & {\fontsize{6.5}{7.5}\selectfont$\pm$0.20} & {\fontsize{6.5}{7.5}\selectfont$\pm$0.21} & {\fontsize{6.5}{7.5}\selectfont$\pm$0.21} & {\fontsize{6.5}{7.5}\selectfont$\pm$0.20} & {\fontsize{6.5}{7.5}\selectfont$\pm$0.14} & {\fontsize{6.5}{7.5}\selectfont$\pm$0.07} & {\fontsize{6.5}{7.5}\selectfont$\pm$0.10} & {\fontsize{6.5}{7.5}\selectfont$\pm$0.08} &  \\[1pt]
\hspace{0.5em}Latent prediction & 0.56 & 0.52 & 0.52 & 0.40 & 0.49 & 0.47 & 0.68 & 0.65 & +0.48 & $<$0.001 \\
 & {\fontsize{6.5}{7.5}\selectfont$\pm$0.14} & {\fontsize{6.5}{7.5}\selectfont$\pm$0.21} & {\fontsize{6.5}{7.5}\selectfont$\pm$0.22} & {\fontsize{6.5}{7.5}\selectfont$\pm$0.25} & {\fontsize{6.5}{7.5}\selectfont$\pm$0.20} & {\fontsize{6.5}{7.5}\selectfont$\pm$0.21} & {\fontsize{6.5}{7.5}\selectfont$\pm$0.07} & {\fontsize{6.5}{7.5}\selectfont$\pm$0.11} & {\fontsize{6.5}{7.5}\selectfont$\pm$0.08} &  \\[1pt]
\bottomrule
\end{tabular}
\endgroup

%% file: figures/fig_latent_structure/table.tex
\begingroup\fontsize{8}{9.5}\selectfont\setlength{\tabcolsep}{3pt}
\begin{tabular}{@{}l*{6}{c}@{}}
\toprule
 & \shortstack{Components\\for 95\%} & \shortstack{Participation\\ratio} & \shortstack{Turning-\\angle ratio} & \shortstack{Within / across\\video distance} & \shortstack{Spectral\\slope} & \shortstack{High-frequency\\spectral share} \\
\midrule
\multicolumn{7}{@{}l}{\textit{Equal reconstruction, three seeds}} \\
\hspace{0.5em}Reconstruction & \begin{tabular}[t]{@{}c@{}}388\\{\fontsize{6}{7}\selectfont$\pm$0.9}\end{tabular} & \begin{tabular}[t]{@{}c@{}}52.7\\{\fontsize{6}{7}\selectfont$\pm$0.03}\end{tabular} & \begin{tabular}[t]{@{}c@{}}1.100\\{\fontsize{6}{7}\selectfont$\pm$0.0001}\end{tabular} & \begin{tabular}[t]{@{}c@{}}0.423\\{\fontsize{6}{7}\selectfont$\pm$0.0004}\end{tabular} & \begin{tabular}[t]{@{}c@{}}$-$0.837\\{\fontsize{6}{7}\selectfont$\pm$0.0003}\end{tabular} & \begin{tabular}[t]{@{}c@{}}0.073\\{\fontsize{6}{7}\selectfont$\pm$0.00008}\end{tabular} \\
\hspace{0.5em}Latent prediction & \begin{tabular}[t]{@{}c@{}}177\\{\fontsize{6}{7}\selectfont$\pm$8}\end{tabular} & \begin{tabular}[t]{@{}c@{}}49.4\\{\fontsize{6}{7}\selectfont$\pm$1.1}\end{tabular} & \begin{tabular}[t]{@{}c@{}}1.019\\{\fontsize{6}{7}\selectfont$\pm$0.007}\end{tabular} & \begin{tabular}[t]{@{}c@{}}0.344\\{\fontsize{6}{7}\selectfont$\pm$0.005}\end{tabular} & \begin{tabular}[t]{@{}c@{}}$-$0.907\\{\fontsize{6}{7}\selectfont$\pm$0.010}\end{tabular} & \begin{tabular}[t]{@{}c@{}}0.061\\{\fontsize{6}{7}\selectfont$\pm$0.002}\end{tabular} \\
\hspace{0.5em}Pixel prediction & \begin{tabular}[t]{@{}c@{}}445\\{\fontsize{6}{7}\selectfont$\pm$0.3}\end{tabular} & \begin{tabular}[t]{@{}c@{}}64.4\\{\fontsize{6}{7}\selectfont$\pm$0.2}\end{tabular} & \begin{tabular}[t]{@{}c@{}}1.110\\{\fontsize{6}{7}\selectfont$\pm$0.0004}\end{tabular} & \begin{tabular}[t]{@{}c@{}}0.443\\{\fontsize{6}{7}\selectfont$\pm$0.001}\end{tabular} & \begin{tabular}[t]{@{}c@{}}$-$0.763\\{\fontsize{6}{7}\selectfont$\pm$0.0008}\end{tabular} & \begin{tabular}[t]{@{}c@{}}0.094\\{\fontsize{6}{7}\selectfont$\pm$0.0002}\end{tabular} \\
\hspace{0.5em}Contrastive prediction & \begin{tabular}[t]{@{}c@{}}293\\{\fontsize{6}{7}\selectfont$\pm$6}\end{tabular} & \begin{tabular}[t]{@{}c@{}}46.7\\{\fontsize{6}{7}\selectfont$\pm$0.7}\end{tabular} & \begin{tabular}[t]{@{}c@{}}1.086\\{\fontsize{6}{7}\selectfont$\pm$0.0009}\end{tabular} & \begin{tabular}[t]{@{}c@{}}0.384\\{\fontsize{6}{7}\selectfont$\pm$0.002}\end{tabular} & \begin{tabular}[t]{@{}c@{}}$-$0.933\\{\fontsize{6}{7}\selectfont$\pm$0.010}\end{tabular} & \begin{tabular}[t]{@{}c@{}}0.052\\{\fontsize{6}{7}\selectfont$\pm$0.002}\end{tabular} \\
\midrule
\multicolumn{7}{@{}l}{\textit{Equal budget, past-conditioned decoder, three seeds}} \\
\hspace{0.5em}Reconstruction & \begin{tabular}[t]{@{}c@{}}316\\{\fontsize{6}{7}\selectfont$\pm$10}\end{tabular} & \begin{tabular}[t]{@{}c@{}}47.7\\{\fontsize{6}{7}\selectfont$\pm$0.7}\end{tabular} & \begin{tabular}[t]{@{}c@{}}1.090\\{\fontsize{6}{7}\selectfont$\pm$0.002}\end{tabular} & \begin{tabular}[t]{@{}c@{}}0.402\\{\fontsize{6}{7}\selectfont$\pm$0.004}\end{tabular} & \begin{tabular}[t]{@{}c@{}}$-$0.886\\{\fontsize{6}{7}\selectfont$\pm$0.008}\end{tabular} & \begin{tabular}[t]{@{}c@{}}0.063\\{\fontsize{6}{7}\selectfont$\pm$0.002}\end{tabular} \\
\hspace{0.5em}Latent prediction & \begin{tabular}[t]{@{}c@{}}110\\{\fontsize{6}{7}\selectfont$\pm$1}\end{tabular} & \begin{tabular}[t]{@{}c@{}}36.0\\{\fontsize{6}{7}\selectfont$\pm$0.5}\end{tabular} & \begin{tabular}[t]{@{}c@{}}0.879\\{\fontsize{6}{7}\selectfont$\pm$0.002}\end{tabular} & \begin{tabular}[t]{@{}c@{}}0.323\\{\fontsize{6}{7}\selectfont$\pm$0.004}\end{tabular} & \begin{tabular}[t]{@{}c@{}}$-$0.948\\{\fontsize{6}{7}\selectfont$\pm$0.004}\end{tabular} & \begin{tabular}[t]{@{}c@{}}0.049\\{\fontsize{6}{7}\selectfont$\pm$0.001}\end{tabular} \\
\hspace{0.5em}Inverse dynamics & \begin{tabular}[t]{@{}c@{}}239\\{\fontsize{6}{7}\selectfont$\pm$4}\end{tabular} & \begin{tabular}[t]{@{}c@{}}44.9\\{\fontsize{6}{7}\selectfont$\pm$2.0}\end{tabular} & \begin{tabular}[t]{@{}c@{}}1.005\\{\fontsize{6}{7}\selectfont$\pm$0.001}\end{tabular} & \begin{tabular}[t]{@{}c@{}}0.464\\{\fontsize{6}{7}\selectfont$\pm$0.002}\end{tabular} & \begin{tabular}[t]{@{}c@{}}$-$1.020\\{\fontsize{6}{7}\selectfont$\pm$0.009}\end{tabular} & \begin{tabular}[t]{@{}c@{}}0.043\\{\fontsize{6}{7}\selectfont$\pm$0.002}\end{tabular} \\
\midrule
\multicolumn{7}{@{}l}{\textit{Equal budget, ordinary decoder (reconstruction: three seeds; pretrained: one)}} \\
\hspace{0.5em}Reconstruction & \begin{tabular}[t]{@{}c@{}}345\\{\fontsize{6}{7}\selectfont$\pm$12}\end{tabular} & \begin{tabular}[t]{@{}c@{}}49.8\\{\fontsize{6}{7}\selectfont$\pm$0.8}\end{tabular} & \begin{tabular}[t]{@{}c@{}}1.092\\{\fontsize{6}{7}\selectfont$\pm$0.002}\end{tabular} & \begin{tabular}[t]{@{}c@{}}0.407\\{\fontsize{6}{7}\selectfont$\pm$0.005}\end{tabular} & \begin{tabular}[t]{@{}c@{}}$-$0.864\\{\fontsize{6}{7}\selectfont$\pm$0.009}\end{tabular} & \begin{tabular}[t]{@{}c@{}}0.067\\{\fontsize{6}{7}\selectfont$\pm$0.002}\end{tabular} \\
\hspace{0.5em}Pretrained, no fine-tuning & 383 & 52.5 & 1.099 & 0.421 & $-$0.838 & 0.072 \\
\bottomrule
\end{tabular}
\endgroup

%% file: figures/tab_dino/table_full.tex
\begingroup\fontsize{8}{9.5}\selectfont\setlength{\tabcolsep}{3pt}
\begin{tabular}{@{}l*{6}{c}@{}}
\toprule
 & \multicolumn{3}{c}{ID} & \multicolumn{3}{c}{OOD} \\
\cmidrule(lr){2-4}\cmidrule(lr){5-7}
 & MSE $\downarrow$ & LPIPS $\downarrow$ & SSIM $\uparrow$ & MSE $\downarrow$ & LPIPS $\downarrow$ & SSIM $\uparrow$ \\
\midrule
Reconstruction & 10.78 & 0.158 & 0.797 & 35.11 & 0.278 & 0.698 \\
Latent prediction, $\lambda = 0.1$ & {\boldmath\bfseries 10.57} & {\boldmath\bfseries 0.150} & {\boldmath\bfseries 0.805} & 30.31 & {\boldmath\bfseries 0.257} & {\boldmath\bfseries 0.715} \\
Latent prediction, $\lambda = 0.3$ & 11.67 & 0.163 & 0.790 & 29.68 & 0.271 & 0.704 \\
Latent prediction, $\lambda = 0.7$ & 11.98 & 0.178 & 0.763 & {\boldmath\bfseries 28.21} & 0.299 & 0.673 \\
\midrule[\heavyrulewidth]
 & \multicolumn{3}{c}{2-step} & \multicolumn{3}{c}{Long-horizon} \\
\cmidrule(lr){2-4}\cmidrule(lr){5-7}
 & MSE $\downarrow$ & LPIPS $\downarrow$ & SSIM $\uparrow$ & MSE $\downarrow$ & LPIPS $\downarrow$ & SSIM $\uparrow$ \\
\midrule
Reconstruction & 11.01 & 0.160 & 0.800 & 30.94 & 0.289 & 0.687 \\
Latent prediction, $\lambda = 0.1$ & {\boldmath\bfseries 10.82} & {\boldmath\bfseries 0.151} & {\boldmath\bfseries 0.807} & {\boldmath\bfseries 26.75} & {\boldmath\bfseries 0.265} & {\boldmath\bfseries 0.716} \\
Latent prediction, $\lambda = 0.3$ & 12.19 & 0.165 & 0.792 & 32.39 & 0.292 & 0.685 \\
Latent prediction, $\lambda = 0.7$ & 13.10 & 0.185 & 0.761 & 33.15 & 0.298 & 0.667 \\
\midrule[\heavyrulewidth]
 & \multicolumn{2}{c}{Predictability} & \multicolumn{2}{c}{Distribution} \\
\cmidrule(lr){2-3}\cmidrule(lr){4-5}
 & ID latent MSE $\downarrow$ & OOD latent MSE $\downarrow$ & FID $\downarrow$ & FVD $\downarrow$ \\
\midrule
Reconstruction & 0.522 & 0.531 & 8.7 & 26.0 \\
Latent prediction, $\lambda = 0.1$ & 0.156 & 0.168 & {\boldmath\bfseries 7.5} & {\boldmath\bfseries 21.5} \\
Latent prediction, $\lambda = 0.3$ & 0.058 & 0.065 & 13.2 & 36.4 \\
Latent prediction, $\lambda = 0.7$ & {\boldmath\bfseries 0.026} & {\boldmath\bfseries 0.033} & 22.8 & 56.2 \\
\bottomrule
\end{tabular}
\endgroup

%% file: figures/tab_kl/table_full.tex
\begingroup\fontsize{8}{9.5}\selectfont\setlength{\tabcolsep}{3pt}
\begin{tabular}{@{}l*{6}{c}@{}}
\toprule
 & \multicolumn{3}{c}{ID} & \multicolumn{3}{c}{OOD} \\
\cmidrule(lr){2-4}\cmidrule(lr){5-7}
 & MSE $\downarrow$ & LPIPS $\downarrow$ & SSIM $\uparrow$ & MSE $\downarrow$ & LPIPS $\downarrow$ & SSIM $\uparrow$ \\
\midrule
\multicolumn{7}{@{}l}{\textit{No KL penalty}} \\
\hspace{0.5em}Reconstruction & 13.76 & 0.178 & 0.801 & 31.81 & 0.274 & 0.713 \\
\hspace{0.5em}Latent prediction & 14.16 & 0.170 & 0.808 & 26.58 & 0.260 & 0.738 \\
\multicolumn{7}{@{}l}{\textit{KL penalty, weight $10^{-6}$}} \\
\hspace{0.5em}Reconstruction & 13.71 & 0.169 & 0.810 & 36.35 & 0.258 & 0.725 \\
\hspace{0.5em}Latent prediction & {\boldmath\bfseries 13.55} & {\boldmath\bfseries 0.164} & {\boldmath\bfseries 0.814} & {\boldmath\bfseries 25.27} & {\boldmath\bfseries 0.241} & {\boldmath\bfseries 0.751} \\
\midrule[\heavyrulewidth]
 & \multicolumn{3}{c}{2-step} & \multicolumn{3}{c}{Long-horizon} \\
\cmidrule(lr){2-4}\cmidrule(lr){5-7}
 & MSE $\downarrow$ & LPIPS $\downarrow$ & SSIM $\uparrow$ & MSE $\downarrow$ & LPIPS $\downarrow$ & SSIM $\uparrow$ \\
\midrule
\multicolumn{7}{@{}l}{\textit{No KL penalty}} \\
\hspace{0.5em}Reconstruction & 17.87 & 0.203 & 0.782 & 45.27 & 0.373 & 0.620 \\
\hspace{0.5em}Latent prediction & 15.38 & 0.178 & 0.804 & {\boldmath\bfseries 40.98} & {\boldmath\bfseries 0.344} & {\boldmath\bfseries 0.647} \\
\multicolumn{7}{@{}l}{\textit{KL penalty, weight $10^{-6}$}} \\
\hspace{0.5em}Reconstruction & 16.67 & 0.187 & 0.796 & 43.57 & 0.364 & 0.627 \\
\hspace{0.5em}Latent prediction & {\boldmath\bfseries 14.23} & {\boldmath\bfseries 0.170} & {\boldmath\bfseries 0.812} & 41.84 & 0.348 & 0.640 \\
\midrule[\heavyrulewidth]
 & \multicolumn{2}{c}{Distribution} \\
\cmidrule(lr){2-3}
 & FID $\downarrow$ & FVD $\downarrow$ \\
\midrule
\multicolumn{7}{@{}l}{\textit{No KL penalty}} \\
\hspace{0.5em}Reconstruction & 18.5 & 36.5 \\
\hspace{0.5em}Latent prediction & 20.7 & 43.8 \\
\multicolumn{7}{@{}l}{\textit{KL penalty, weight $10^{-6}$}} \\
\hspace{0.5em}Reconstruction & 15.4 & {\boldmath\bfseries 35.3} \\
\hspace{0.5em}Latent prediction & {\boldmath\bfseries 14.8} & 38.2 \\
\bottomrule
\end{tabular}
\endgroup